\documentclass[11pt]{article}

\usepackage[final]{acl}

\usepackage{times}
\usepackage{latexsym}
\usepackage{booktabs}
\usepackage{longtable}
\usepackage{amsmath}
\usepackage[T1]{fontenc}

\usepackage[utf8]{inputenc}
\usepackage{multirow}
\usepackage{microtype}

\usepackage{inconsolata}

\usepackage{graphicx}
\usepackage{enumitem}
\usepackage{algorithm}
\usepackage{algpseudocode}
\usepackage{pifont}
\newcommand{\cmark}{\ding{51}}
\newcommand{\xmark}{\ding{55}}
\usepackage{amssymb}
\usepackage[dvipsnames]{xcolor}
\newcommand{\cAcc}{\textit{c-S2}}
\newcommand{\cIoU}{\textit{c-mIoU}}

\usepackage{colortbl}
\usepackage{xcolor}
\definecolor{c00}{HTML}{F5F8FA}
\definecolor{c20}{HTML}{D6E4EF}
\definecolor{c40}{HTML}{B3D1E5}
\definecolor{c60}{HTML}{7FAFCF}
\definecolor{c80}{HTML}{4A90C7}
\definecolor{c100}{HTML}{1F6FA8}
\newcommand{\hc}[2]{\cellcolor{#1}#2}
\newcommand{\ci}[3]{\shortstack{#1\\{\fontsize{6}{6.5}\selectfont[#2,\,#3]}}}

\title{Cultural Moment Benchmark: \\ Evaluating Video Cultural Reasoning and Grounding in Southeast Asia}

\author{
 \textbf{Burak Satar\textsuperscript{1}},
 \textbf{Zhixin Ma\textsuperscript{1}},
 \textbf{Cheng Yu-Tong\textsuperscript{1}},
\\
 \textbf{Huy Hoang Tran\textsuperscript{2}},
 \textbf{Phuong Anh Nguyen\textsuperscript{1}},
 \textbf{Chong-Wah Ngo\textsuperscript{1}},
\\
\texttt{\{buraks, cwngo\}@smu.edu.sg}
\\
 \textsuperscript{1}Singapore Management University,
 \textsuperscript{2}UIT, VNU-HCM
\\[2pt]
\href{https://culturalmoment-benchmark.github.io}{\includegraphics[height=6ex]{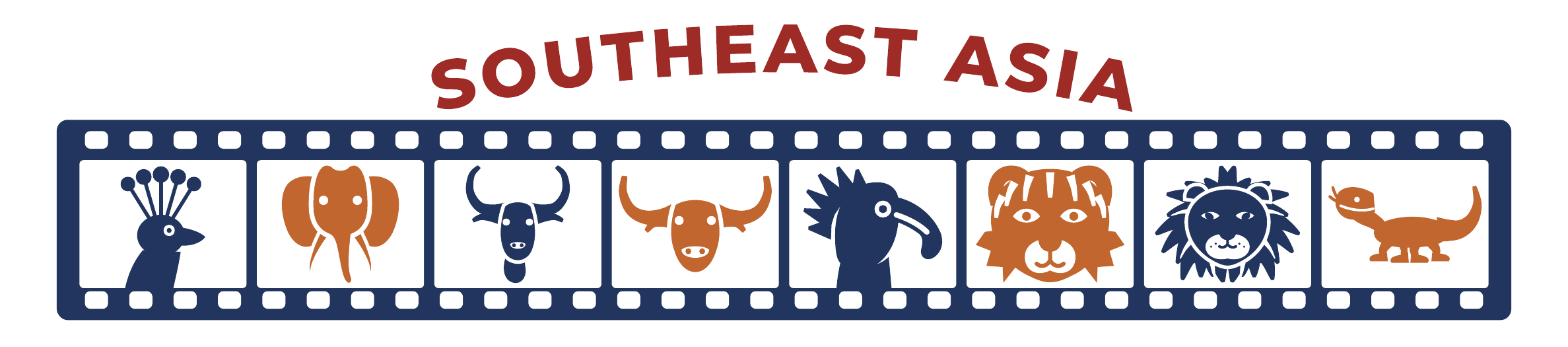}}\\[-1pt]
{\color{NavyBlue}\href{https://culturalmoment-benchmark.github.io}{\small https://culturalmoment-benchmark.github.io}}\\
}
\begin{document}
\maketitle

\begin{abstract}
Cultural understanding in video means more than recognizing what is visible; it requires grasping the symbolic and temporal significance of cultural concepts. We decompose this into three abilities: naming what a concept symbolizes, visually recognizing it on video, and locating its sub-events in time. Existing video-cultural benchmarks tend to test what is seen, collapsing these three abilities into a single score that hides the bottleneck. We introduce the \textbf{Cultural Moment Benchmark} (CMB): 306 expert-curated concepts from seven countries in Southeast Asia across five categories. We evaluate each concept through three stages, one per ability. Given a description, \textbf{Stage 1} (S1) selects from four candidate concept names, \textbf{Stage 2} (S2) selects from four candidate video moments, and \textbf{Stage 3} (S3) predicts the start and end times of the moment in a video. To keep each stage focused on a distinct ability, we use three design choices: semantic-similarity distractors (S1, S2), unlabeled video moments (S2), and free-form localization on a different example video (S3). Across six vision-language models, failure modes vary by ability and modality. i) Even the strongest closed-source models score below 30\% when all three stages must be correct; ii) The three abilities do not fully cascade: naming a concept correctly helps half the models recognize it on video, but recognizing it has little effect on locating the sub-event in time; iii) Audio is complementary, redundant, or distracting depending on the concept, more often distracting in non-Latin-script countries; removing both audio and subtitles hurts Games and Music the most. Our 14-rater human study shows that even \textit{Expert} raters score below chance on concepts from a neighboring country, indicating that CMB requires country-specific cultural knowledge. CMB acts as a diagnostic harness, attributing failures to a specific ability or modality.

\end{abstract}

\begin{figure}[!htb]
  \centering
  \includegraphics[width=0.89\linewidth]{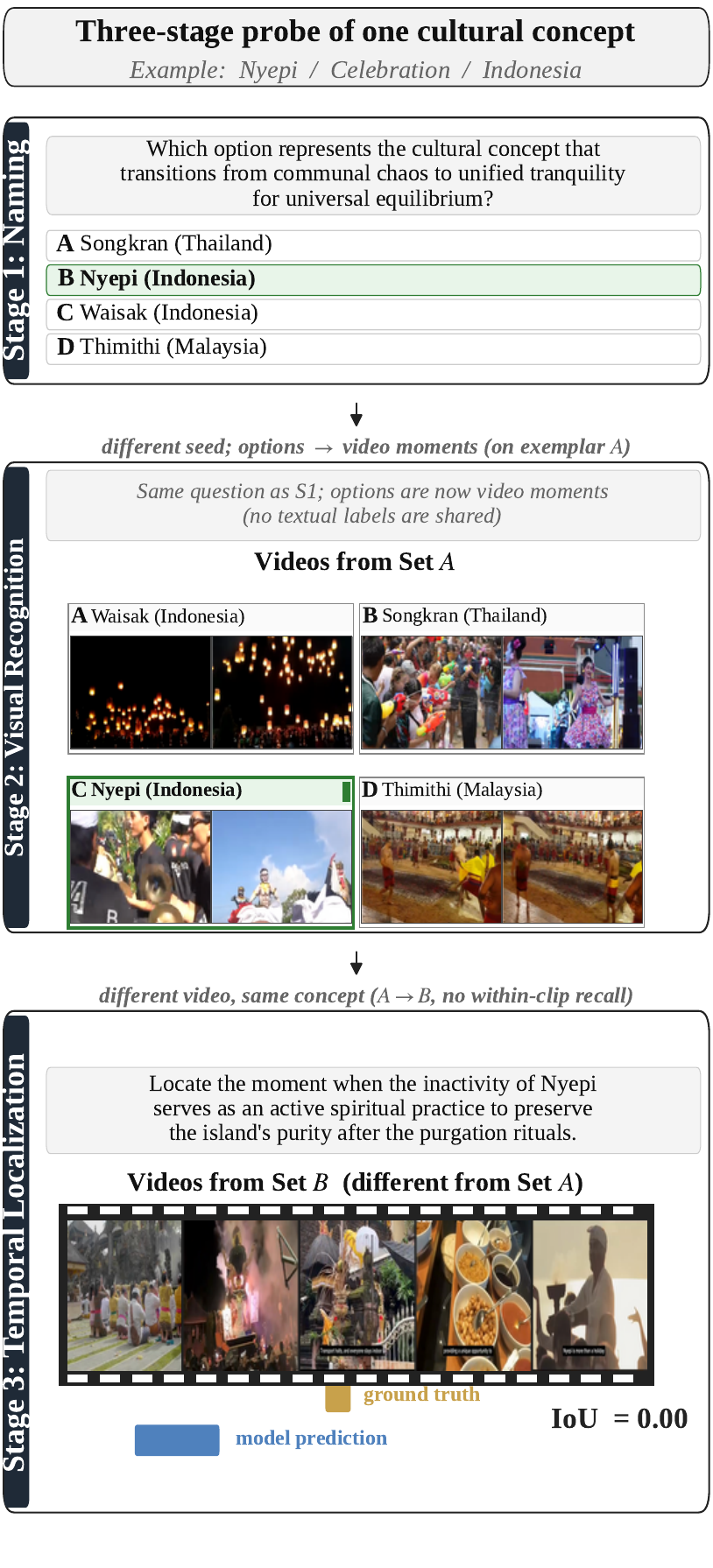}
  \caption{CMB three-stage probe. Stage 1 picks among four concept names. Stage 2 keeps the same stem but replaces the options with four video moments from example video $A$; it uses an independent random seed, so the option positions differ from S1. Stage 3 predicts a free-form $[t_{\mathrm{start}}, t_{\mathrm{end}}]$ span for a sub-event of the concept on a different example video $B$. The shift from $A$ to $B$ prevents within-clip memorization in Stage 3.}
  \label{fig:teaser}
\end{figure}

\section{Introduction}
\label{sec:intro}

Vision-language models (VLMs) have made substantial progress on video question answering, text-video retrieval \cite{satar2021semantic}, and temporal grounding \cite{Fu2024VideoMMETF, Li2024MVBenchAC}. Yet their ability to understand cultural content beyond what is visible remains underexplored. Existing video-cultural benchmarks \cite{shafique-etal-2025-culturally, chen-etal-2025-videovista, Singh2026MINERVACulturalAB} tend to ask questions like ``Which country is this clothing from?'' or ``Where is the dish popular?''. Such questions can be answered by identifying what is seen, without engaging what it symbolizes. The same problem extends to cultural reasoning questions: even when a question appears to require cultural reasoning, the model can often answer it by matching keywords. A question about ``a Burmese water-throwing celebration'' is solved by finding a water frame without needing to know what Thingyan symbolizes \cite{satar-etal-2025-seeing}. Our benchmark addresses this gap by asking what each concept symbolizes, such as ``the washing away of misfortunes'' for Thingyan, instead of just naming visible cues, as Figure~\ref{fig:teaser} illustrates.

A single accuracy number conflates naming, visual recognition, and temporal localization, hiding which ability is failing. To diagnose this, we treat these three abilities as distinct and score each separately, providing a per-model failure profile rather than a single aggregate score. We instantiate this approach in the Cultural Moment Benchmark (CMB), focused on Southeast Asia (SEA). SEA cultures are diverse and underrepresented in current video-cultural benchmarks. Neighboring traditions often share enough surface features to demand fine-grained discrimination. CMB contains 306 expert-curated concepts from seven SEA countries across five categories. Four use the Latin script: Indonesia, Malaysia, the Philippines, and Vietnam. Three use non-Latin scripts: Cambodia, Myanmar, and Thailand. 

We score each ability separately by routing every concept through three stages. \textbf{Stage 1} (S1) tests naming: given a description of what a concept symbolizes, the model picks its name from four candidates. \textbf{Stage 2} (S2) tests visual recognition: from the same description, it picks among four candidate video moments. \textbf{Stage 3} (S3) tests temporal localization: on a different example video, it predicts the start and end times of a target sub-event.
Three design choices keep each stage focused on its own ability: S1 and S2 distractors are semantically similar to the correct answer, so keyword matching does not solve either stage; S2 moments carry no titles, captions, or filenames, so the judgment must be visual; and S3 uses a fresh video with a free-form span, so cues from earlier stages cannot be reused.
Beyond ability isolation, we evaluate each stage under three modes that vary the prior-stage context: Reset (no prior context), Carry (the model's own previous answer), and Feedback (the previous answer plus the ground truth). Together, the three stages and three modes form a $3 \times 3$ evaluation framework.

With this setup, we evaluate six state-of-the-art VLMs on CMB. Across these six models, the failure mode varies by ability and modality, confirming that a single score hides the bottleneck. The following patterns stand out. i) Single-stage scores understate the gap; end-to-end success reveals it. Even Gemini 3.1 Pro and GPT-5.4, the two strongest closed-source models, get fewer than 30\% of concepts right across all three stages. The four open-source models score in single digits, and InternVL-14B is near zero. ii) Across stages, the three abilities do not fully cascade. Naming a concept correctly helps half the models recognize it on video (S1→S2). However, recognizing it has little effect on locating the sub-event in time on a different video (S2→S3). iii) Modality roles also vary: audio is not uniformly useful. Depending on the concept, it can be complementary, redundant, or distracting. The distracting role is more frequent in countries with non-Latin scripts. Removing audio and subtitles together hurts Games and Music the most, where commentary and lyrics align with the sub-event; in many traditional concepts, audio is ambient backing that holds throughout the clip. Beyond modality, cultural knowledge is country-specific, not regional. In a 14-rater human study, even \textit{Expert} raters score below chance on concepts from a neighboring country, indicating that CMB requires country-specific cultural knowledge.

\begin{table*}[t]
\centering
\small
\setlength{\tabcolsep}{5pt}
\renewcommand{\arraystretch}{1.15}
\resizebox{\textwidth}{!}{
\begin{tabular}{lcccccccc}
\toprule
\textbf{Benchmark} & \textbf{SEA} & \textbf{Multi-stage} & \textbf{Cond.} & \textbf{Free-form} & \textbf{Avg. dur.} & \textbf{Human} & \textbf{Cultures} \\
                   & \textbf{focus} & \textbf{cascade} & \textbf{metrics} & \textbf{temp. loc.} & \textbf{(min)} & \textbf{eval} & \textbf{covered} \\
\midrule
SCB$^{\text{image}}$ \citep{satar-etal-2025-seeing}             & \cmark    & \cmark    & \cmark    & \xmark    & n/a          & \xmark           & 7 \\
AVMeme Exam \citep{jiang2026avmeme}                            & \xmark    & \xmark    & \xmark    & \xmark    & $\leq$0.5    & \cmark\,(closed) & 5+ \\
ChineseVideoBench \citep{nie2025chinesevideobench}             & \xmark    & \xmark    & \xmark    & \xmark    & $\approx$1.0 & \xmark           & 1 \\
GIMMICK \citep{schneider2025gimmickgloballyinclusive}          & partial   & \xmark    & \xmark    & \xmark    & 0.17         & \xmark           & 139 \\
MINERVA-Cultural \citep{Singh2026MINERVACulturalAB}            & partial   & \xmark    & \cmark    & \xmark    & 12.62        & \cmark           & 18 \\
VideoNorms \citep{VideoNorms_2025}              & \xmark    & \xmark    & \xmark    & \xmark    & 0.25         & \xmark           & 2 \\
VideoVista-CulturalLingo \citep{chen-etal-2025-videovista}     & \xmark    & \xmark    & \xmark    & \xmark    & 4.05         & \xmark           & 3 \\
ViMUL-Bench \citep{shafique-etal-2025-culturally}              & \xmark    & \xmark    & \xmark    & \xmark    & 2.92         & partial          & 14 \\
\midrule
\textbf{CMB (ours)}                                            & \cmark    & \cmark    & \cmark    & \cmark    & 6.2*          & \cmark\,(closed) & 7 \\
\bottomrule
\end{tabular}
}
\caption{Comparison of CMB to recent multicultural video benchmarks. SCB$^{\text{image}}$ is the image-based precursor; all other rows are video benchmarks. \textbf{Multi-stage cascade}: distinct ability stages scored sequentially with conditional dependency. \textbf{Cond.\ metrics}: conditional metrics that report cross-stage success-given-success (e.g., $P(\text{S2} \mid \text{S1 correct})$). \textbf{Free-form temp. loc.}: model predicts an unconstrained $[\hat{t}_{\text{start}}, \hat{t}_{\text{end}}]$ span rather than selecting from candidate timestamps or orderings. \textbf{Human eval}: \cmark indicates humans recruited separately from annotators were evaluated on the same task; \textit{closed} adds an explicit prohibition on web search or external aids. *: S3 videos.}
\label{tab:related_comparison}
\end{table*}

CMB acts as a diagnostic harness, attributing failures to a specific ability or modality rather than aggregating them into a single score. We contribute: 
\begin{itemize}[noitemsep, topsep=0pt]
    \item CMB, an expert-curated video benchmark dedicated to SEA cultures, grounded in real-world traditions; 
    \item A 3-stage × 3-mode evaluation framework with per-stage scores, conditional metrics (c-S2, c-mIoU) for cascade analysis, and a Joint score for end-to-end success; 
    \item Failure profiles for six VLMs to pinpoint each model's bottleneck stage, suggesting model-specific intervention priorities.
\end{itemize}

\section{Related Work}
\label{sec:related_work}

\subsection{Multicultural Image Benchmarks}
\label{sec:related_image}

Most multicultural benchmarks for VLMs use a single image as input and evaluate cultural understanding via either textual or visual answer choices. The textual-choice family includes CVQA \citep{mogrovejo2024cvqa}, CulturalVQA \citep{nayak-etal-2024-benchmarking_CulturalVQA}, SEA-VQA \citep{2024_seavqa}, K-Viscuit \citep{corr_kviscuit}, CultureVerse \citep{liu2025culturevlm}, the image subset of GIMMICK \citep{schneider2025gimmickgloballyinclusive}, and ALM-Bench \citep{ALM-Bench_2025}. CultureVerse and parts of GIMMICK additionally test cultural reasoning. MaRVL \citep{liu-etal-2021-visually_MaRVL} checks truth-values of image captions across five languages. Region- or language-specific image benchmarks include Rice-VL \citep{Pranav2025RiceVLEV} and MMA-ASIA \citep{zheng2025mmaasia} for Asia, IndicVisionBench \citep{faraz2026indicvisionbench} for India, TaiwanVQA \citep{hsieh-etal-2025-taiwanvqa} for Taiwan, Afri-MCQA \citep{tonja2026afrimcq} for Africa, and BanglaProtha \citep{Fahim_2026_WACV} for Bengali culture. 
FoodieQA \citep{li-etal-2024-foodieqa} introduces visual options for Chinese cuisine. Beyond static accuracy, diagnostic studies report that VLMs falter when cultural cues are mixed or perturbed \citep{irawan2025confused, kim2025worldinaframe} and that their cultural knowledge is brittle under rephrasing and visual grounding \citep{tan2025blendvis}. ValueGround probes the complementary axis of culture-conditioned visual value grounding \citep{wang2026valueground}. On the text side, resources specific to Southeast Asia now range from embedding benchmarks \citep{ponwitayarat2026seabed} to culturally grounded safety filters \citep{tasawong2026seaguard}. Most relevant to our work, SCB \citep{satar-etal-2025-seeing} adapts the visual-option design to SEA cultural reasoning and pairs it with cultural-artifact segmentation at the image level. CMB carries this design into video, adding a naming stage on the same description (S1), video-moment options (S2), and free-form temporal localization on a different video (S3). Image-only setups, including SCB, do not isolate naming from visual recognition and cannot locate when a cultural sub-event unfolds within a video.

\subsection{Multicultural Video Benchmarks}
\label{sec:related_video}

Cultural understanding has recently been extended to video, where the challenge is harder because cultural concepts unfold across time, audio, and visual modalities. VideoVista-CulturalLingo \citep{chen-etal-2025-videovista} covers three cultural regions (Chinese, North American, European) in MCQ design across 1,389 videos. ViMUL-Bench \cite{shafique-etal-2025-culturally} is a multilingual, culturally diverse VideoQA benchmark spanning 14 languages and cultures via multimodal QA in 337 videos. MINERVA-Cultural \citep{Singh2026MINERVACulturalAB} emphasizes long-form video reasoning across 18 locales with human-authored reasoning traces. AVMeme Exam \citep{jiang2026avmeme} probes cultural and contextual knowledge in 1,032 audio-visual memes spanning speech, songs, music, and sound effects. VideoNorms \citep{VideoNorms_2025} targets cultural norms across 384 concepts in American and Chinese contexts. ChineseVideoBench \citep{nie2025chinesevideobench} is a mono-cultural Chinese video QA benchmark with 1{,}625 videos across eight task categories. The video subset of GIMMICK \citep{schneider2025gimmickgloballyinclusive} also includes short video moments from 139 cultures, accompanied by AI-generated open-ended questions. Adjacent video work includes MAVEN \citep{li2026maven} and CultureVidBench \citep{han2026culturevidbench} for multicultural text-to-video generation, culture-lens evaluation of visual generative models \citep{koksal2025benchmarking}, and MusiQAl \citep{Christodoulou-2025} for audio-video music question answering; these test generation and music-specific understanding rather than cultural video QA. 
Even within video, existing benchmarks share three design gaps that CMB closes. First, none decomposes performance by ability. Second, none uses video moments as answer options for cultural recognition. Third, none includes free-form temporal localization on a different example video. Table~\ref{tab:related_comparison} contrasts CMB against these benchmarks across these dimensions.

\begin{figure*}[!htb]
  \includegraphics[width=1\linewidth]{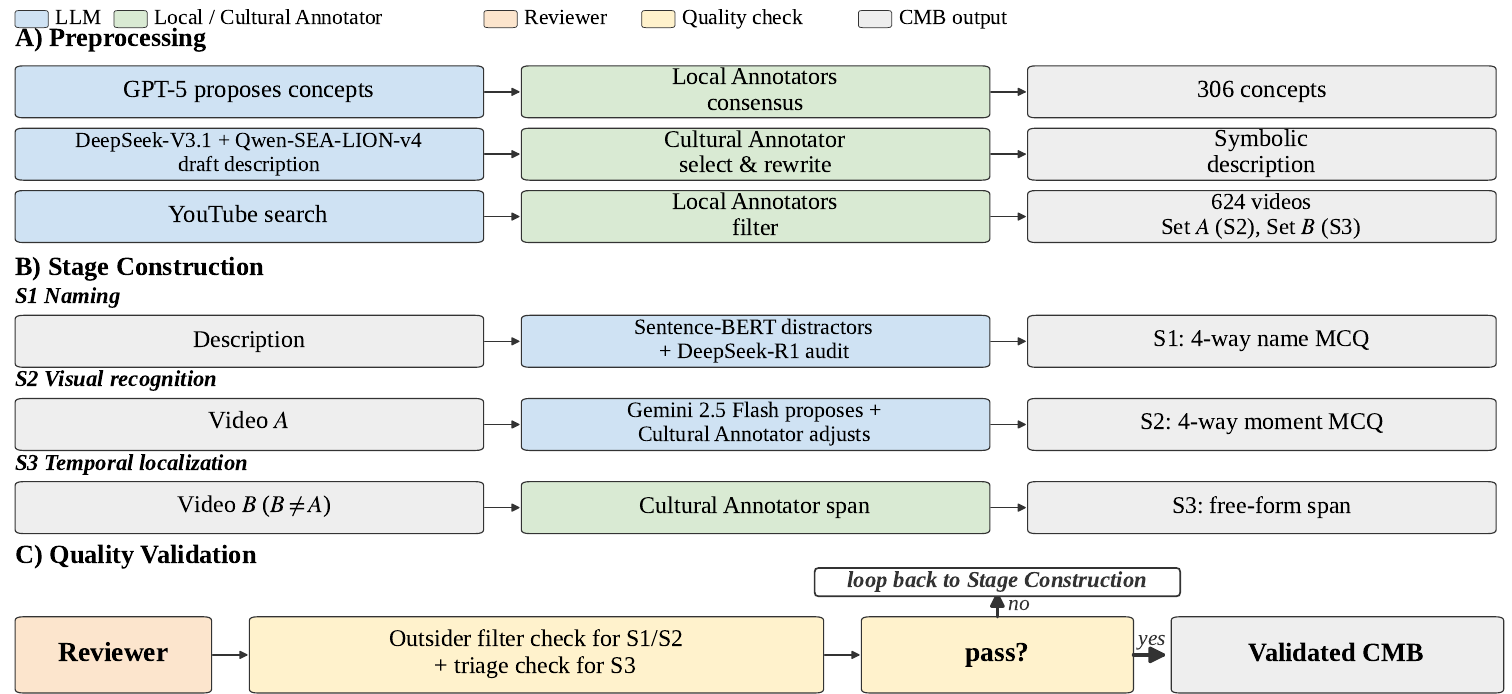}
   \caption{CMB construction pipeline. A) Preprocessing yields 306 concepts (proposed by GPT-5, validated by Local Annotators) and 624 source videos (Set $A$ for S2, Set $B$ for S3); descriptions are drafted by DeepSeek-V3.1 and Qwen-SEA-LION-v4 and rewritten by a Cultural Annotator. B) Each concept is materialized into the S1 and S2 questions (four-way MCQ with semantic-similarity distractors) and the S3 question (free-form span on a different video); flagged S3 sub-events are also audited by a Reviewer. C) A Reviewer runs an Outsider filter check on S1/S2 descriptions (does the Reviewer answer correctly from regional cues alone?) and a triage check on flagged S3 spans. Concepts that pass both enter the Validated CMB; failures loop back to Stage Construction.}  
   \label{fig:pipeline}
\end{figure*}

\section{Cultural Moment Benchmark}
\label{sec:cmb}

CMB is built on one design principle: humans produce all visual ground truth; LLMs scaffold only textual artifacts. The benchmark is constructed in three phases: i) concept curation and video collection, ii) stage construction with semantic-similarity distractors, and iii) quality validation through LLM auditing, an Outsider Filter, and S3 sub-event triage.

\paragraph{Annotators and Reviewers.}

We separate three distinct roles by qualifications and tasks. \textit{Local Annotators} are native speakers of the country's primary language; they validate the concepts proposed by GPT-5 and filter the candidate videos. \textit{Cultural Annotators} are also native speakers and additionally have research or co-curricular involvement in their country's cultural traditions; they curate the symbolic S1/S2 descriptions and annotate S2 moments and S3 spans. \textit{Reviewers} are non-local researchers with documented SEA cultural exposure, distinct from Annotators; they run the Outsider Filter on every concept and triage flagged S3 sub-events. Two Cultural Annotators are co-authors (Vietnam, Malaysia); the remaining five were recruited and compensated.

\subsection{Concepts and Videos}
\label{sec:concepts_videos}

CMB concepts are organized hierarchically as \texttt{country/category/concept}. The seven countries span the diversity of SEA, four with Latin script and three with non-Latin scripts. The five categories are Celebration, Dance, Game, Music, and Wedding.
For each \texttt{(country, category)} pair, GPT-5 \cite{2025gpt5} proposes candidate concepts. Each candidate is reviewed by two Local Annotators per country (three for the Philippines and Myanmar), all of whom are native speakers of the country's primary language. We retain only candidates on which all Annotators reach a unanimous consensus that the candidate names a genuine cultural tradition rather than a stereotype or an overly broad descriptor. 
This yields a final pool of 306 concepts. For each concept, we record four name variants along two axes: common (colloquial) vs. official (formal), and Latin transliteration vs. local script. See Appendix.
For each concept, we query YouTube and Google Video Search by concept and country, e.g. ``Thingyan festival, Myanmar'', collecting up to 20 candidates per concept and about 6{,}120 in total.
Local Annotators filter for real-world depiction, non-duplication, and visual clarity. The final video pool comprises 624 unique videos: Set A (one per concept, 306 videos, average 2.0 min) for S2, and Set B (318 videos, average 6.2 min) for S3, with multi-event traditions contributing more than one Set B video.

\subsection{Stage Construction}
\label{sec:stages}

\paragraph{Symbolic concept descriptions.}

Each S1 and S2 question begins with a textual description of the concept that captures its symbolic meaning, not its visual appearance. Two LLMs from different families, DeepSeek-V3.1~\citep{deepseekv3technicalreport_2025} and Qwen-SEA-LION-v4~\citep{sealion2025}, each draft an independent description (e.g., ``the washing away of misfortunes'' for Thingyan); the Cultural Annotator selects one and rewrites it if it lists only visible objects or actions. This enforces the symbolic constraint: a question that can be answered by generic object recognition (``find the water frame'') is rejected at this stage. 

\paragraph{Stage 1 (Naming).}

Given the description, the model selects the correct concept name from four candidates. The three distractors are drawn via a Semantic Similarity Matrix \cite{Reimers2019SentenceBERTSE} over description embeddings, as shown in Figure~\ref{fig:pipeline}. One distractor is from the same country as the correct answer, forcing fine-grained intra-country discrimination, and two are from different countries with semantically similar concepts, preventing reliance on broad regional priors. To prevent any single concept from dominating the distractor set across items, re-use of any single concept as a distractor is capped at roughly 10\% of the category's concepts.  The full sampling algorithm is given in the Appendix.

\paragraph{Stage 2 (Visual recognition).} 

The model sees the same description and four candidate video moments that correspond to the same concept names as S1. Each moment is a sub-event sourced from the corresponding concept's video in Set A. Gemini 2.5 Flash \citep{2025gemini} proposes a small set of candidate moments per video, highlighting cultural cues; a Cultural Annotator then retains, adjusts the boundary timestamps of, or rejects each candidate, and only Annotator-confirmed moments enter the S2 pool. 
Within each stage, the correct answer letter is balanced across A, B, C, and D so that no letter is correct more often than another across the item pool. S1 and S2 use independent random seeds, so the position of the correct answer is randomized independently across the two stages. 

\paragraph{Stage 3 (Temporal localization).} 

Given the concept name and a textual description of a target sub-event, the model predicts a free-form start and end time on a different \textit{example video B}. Using a separate video from S2 forces the model to apply cultural knowledge to a fresh visual context rather than memorize within-clip cues. Each S3 span is annotated by a Cultural Annotator, who also verifies that the target sub-event occurs exactly once in the source video; videos containing multiple matching sub-events are excluded.

\paragraph{Modality handling at construction and evaluation.}

Cultural Annotators viewed each video with the original audio and with subtitles, if any. Reviewers performing the Outsider Filter for S1 and S2 saw only the textual description and the four name options, with no access to video, audio, or subtitles; this limits the filter to country-specific symbolic knowledge. For S3 evaluation, models receive the full video with native audio, subtitle tracks, and on-screen text intact by default.

\subsection{Quality Validation}
\label{sec:quality}

\paragraph{LLM auditing.}

DeepSeek-R1 \cite{Guo_2025_deepseek-r1} audits the four S1 name options for mutual exclusivity and logical distinctness (``Are these options mutually exclusive and logically distinct?''), and rates each S3 sub-event description for well-definedness; low-rated items are surfaced to Reviewers for rewrite or discard. DeepSeek-R1 never sees video, audio, or frames; all S2 moments and S3 spans are produced and verified by human Annotators and Reviewers.

\paragraph{Outsider Filter.}

For every concept, a Reviewer answers the S1/S2 description from regional priors alone, without consulting country-specific sources. The Reviewer is called an \emph{Outsider} because they have general exposure to SEA but no familiarity with the concept's country of origin. If the Outsider picks the correct name, the description has leaked something a non-local could guess from regional priors: it is either rewritten to be more symbolic (less dependent on visible surface features) or, if no rewrite restores difficulty, the concept is discarded. Because S1 and S2 share the description as their stem, the filter covers both stages.

\paragraph{S3 Reviewer triage.}

For each S3 candidate, the textual rating produced by DeepSeek-R1 for the sub-event description flags items that may be underspecified or ambiguously locatable. A Reviewer inspects each flagged item, watches the candidate span on video~B, and either rewrites the sub-event description, adjusts the span boundaries, or discards the candidate. Unflagged items pass without Reviewer review. This triage focuses Reviewer effort on the highest-risk spans rather than re-auditing all 306 items end-to-end.

\paragraph{Release.} 

The benchmark records and a full evaluation harness, covering all six models and all three stages under all three modes, are released on HuggingFace and GitHub; the release schedule, licensing, and opt-out procedure are given in the Ethical Considerations section.

\begin{table*}[!htb]
\centering
\small
\setlength{\tabcolsep}{3pt}
\renewcommand{\arraystretch}{1.05}
\resizebox{0.85\textwidth}{!}{
\begin{tabular}{l c c c c c c c c c c}
\toprule
\multirow{2}{*}{\textbf{Model}} & \multirow{2}{*}{\textbf{Stage 1 (\%)}} & \multicolumn{3}{c}{\textbf{Stage 2 (\%)}} & \multicolumn{3}{c}{\textbf{Stage 3 (\%)}} & \multicolumn{3}{c}{\textbf{Joint (\%)}} \\
\cmidrule(lr){3-5} \cmidrule(lr){6-8} \cmidrule(lr){9-11}
 & & Reset & Carry & Feedback & Reset & Carry & Feedback & Reset & Carry & Feedback \\
\midrule
Gemini 3.1 Pro       & \hc{c80}{77.8} & \hc{c40}{62.7} & \hc{c60}{72.9} & \hc{c100}{85.9} & \hc{c20}{31.9} & \hc{c20}{33.4} & \hc{c20}{33.3} & \hc{c00}{20.6} & \hc{c20}{27.0} & \hc{c20}{27.8} \\
GPT-5.4              & \hc{c60}{70.9} & \hc{c20}{52.6} & \hc{c40}{68.3} & \hc{c80}{81.7} & \hc{c20}{29.5} & \hc{c20}{31.0} & \hc{c20}{30.3} & \hc{c00}{14.5} & \hc{c20}{21.7} & \hc{c20}{21.5} \\
\midrule
Qwen3-VL-32B         & \hc{c40}{51.0} & \hc{c20}{39.9} & \hc{c00}{31.4} & \hc{c00}{28.4} & \hc{c00}{22.0} & \hc{c00}{22.3} & \hc{c20}{30.0} & \hc{c00}{6.4}  & \hc{c00}{4.3}  & \hc{c00}{4.6}  \\
Qwen3.5-27B          & \hc{c40}{53.6} & \hc{c20}{39.9} & \hc{c20}{40.5} & \hc{c00}{39.9} & \hc{c00}{23.2} & \hc{c00}{24.1} & \hc{c20}{30.2} & \hc{c00}{6.7}  & \hc{c00}{8.1}  & \hc{c00}{7.7}  \\
\midrule
InternVL3.5-14B (n)  & \hc{c20}{46.7} & \hc{c00}{26.5} & \hc{c00}{24.5} & \hc{c00}{9.8}  & \hc{c00}{2.1}  & \hc{c00}{2.5}  & \hc{c00}{12.5} & \hc{c00}{0.4}  & \hc{c00}{0.4}  & \hc{c00}{0.7}  \\
InternVL3.5-14B (t)  & \hc{c20}{41.2} & \hc{c00}{26.5} & \hc{c00}{26.8} & \hc{c00}{16.7} & \hc{c00}{3.2}  & \hc{c00}{3.0}  & \hc{c00}{17.8} & \hc{c00}{0.8}  & \hc{c00}{0.6}  & \hc{c00}{0.8}  \\
\bottomrule
\end{tabular}
}
\caption{Overall results from the 3-stage $\times$ 3-mode analysis. S1 and S2 scores refer to accuracy, and S3 score refers to mIoU. Midrules separate closed-source models (top), 27--32B open-source models (middle), and the 14B InternVL configurations (bottom). S1 has a single column because there is no prior context to carry.}
\label{tab:overall}
\end{table*}

\section{Experiments}

\subsection{Implementation Details}
\label{sec:models}

We evaluate six vision-language models. The two closed-source models are Gemini 3.1 Pro and GPT-5.4. The other four are open-source: Qwen3-VL-32B-Instruct \cite{bai2025qwen3vl}, Qwen3.5-27B \cite{qwen3.5}, and InternVL3.5-14B \cite{wang2025internvl35} evaluated in two configurations, (t) with inference-time reasoning and (n) without. Gemini, GPT, and Qwen read video at one frame per second; InternVL uses its native twelve uniformly spaced frames.

Each model runs each stage three times across three modes, varying what it sees from earlier stages, as shown in Table~\ref{tab:overall}. Under \textit{Reset}, the model sees nothing from earlier stages. Under \textit{Carry}, the model's previous answer is carried over to the next stage. Under \textit{Feedback}, the model's previous answer is carried over along with the ground truth.

S1 and S2 are scored as four-option multiple-choice accuracy. S3 uses the mean Intersection over Union (mIoU) between the predicted span $[\hat{t}_{\text{start}}, \hat{t}_{\text{end}}]$ and the ground-truth span $[t_{\text{start}}, t_{\text{end}}]$, averaged over the $631$ Stage-3 pairs. 

\subsection{Overall Results}
\label{sec:overall}

\paragraph{End-to-end success is low for every model.}

We summarize end-to-end success with the Joint score, which credits a concept only when naming and recognition are both correct, weighted by localization quality: 
$\text{Joint} = \frac{1}{N}\sum_{i=1}^{N} \mathbf{1}\{\text{S1}_{i}\}\cdot \mathbf{1}\{\text{S2}_{i}\}\cdot \text{IoU}_{i}$, where $i$ indexes the $N{=}306$ concepts and $\mathbf{1}\{\cdot\}$ is the indicator function. Overall, Joint scores are low across all six models. Even Gemini and GPT, the two strongest models, achieve a Joint score below 30\%, as shown in Table~\ref{tab:overall}. The four open-source models do much worse: the two Qwen variants score in the single digits, and InternVL sits close to zero. Model scale does not explain the spread: Qwen3.5-27B beats InternVL on Joint by an order of magnitude, despite being only twice its size.

\paragraph{Reset / Carry / Feedback: effect of prior context.}

Table~\ref{tab:overall} shows that prior context helps the strongest models and hurts the weakest. Under \textit{Carry}, Gemini's S2 accuracy rises by about ten points over \textit{Reset}. Under \textit{Feedback}, it rises by another thirteen points. GPT follows the same pattern. The Qwen models show no consistent change across modes, and the InternVL configurations lose accuracy under \textit{Carry}. Prior context confirms the reasoning when the model can already perform the task; otherwise, it serves as a distraction.

\paragraph{The gap between closed and open models survives resampling; the ordering inside each
group does not.}

Table~\ref{tab:overall} reports point estimates. Cluster-bootstrap 95\% confidence intervals over concepts, per country and per category under \textit{Carry} with the base prompt, are in Appendix~\ref{sec:app_stats}. 
Because all six models see the same 306 concepts, we also test each
pairwise difference within each resample rather than asking whether two marginal intervals overlap in Table~\ref{tab:app_paired_bootstrap}. Every closed-source against open-source
comparison is significant on all three stages, and in 12 of the 12 country and category cells the
weaker closed-source model still scores above the stronger open-source one on S2 and mIoU. Within
each group the picture is different: Gemini leads GPT-5.4 on S1 by about seven points but not on
S2 or mIoU, and the two InternVL configurations are indistinguishable on every metric. The
reliable finding is the gap between model families, not the ranking inside one.

\paragraph{Reweighting by country or category leaves recognition unchanged and raises
localization.} 

Table~\ref{tab:app_macro_micro} places micro averages beside macro averages that
give each of the seven countries, and each of the five categories, equal weight. S1 and S2 agree
within 2.5 points for every model under both schemes and no model ordering changes, so the
recognition results are not an artifact of how concepts are distributed. mIoU behaves
differently: for the four models that localize above the floor it runs higher under macro, by
up to 4 points. The cause is Stage-3 pair mass rather than concept counts. Wedding supplies 3.6\% of
concepts but 23.5\% of the 631 Stage-3 pairs and is the hardest cell for every model, while Music
is the mirror image at 22.2\% of concepts and 6.3\% of pairs. Micro mIoU is therefore a
pair-weighted quantity that inherits Wedding's mass, and we report both weightings so that
neither carries a claim on its own. The same imbalance limits every per-country and per-category comparison in this section; see \emph{Long-tailed distribution} in Limitations for its three sources.

\subsection{Cascade Analysis}
\label{sec:cascade}

\begin{table}[t]
\centering
\small
\setlength{\tabcolsep}{4pt}
\resizebox{0.93\columnwidth}{!}{
\begin{tabular}{lclcc}
\toprule
& \multicolumn{2}{c}{\textbf{S1\,$\rightarrow$\,S2 (\%)}} & \multicolumn{2}{c}{\textbf{S2\,$\rightarrow$\,S3 (\%)}} \\
\cmidrule(lr){2-3}\cmidrule(lr){4-5}
\textbf{Model} & S2 & \cAcc & mIoU & \cIoU \\
\midrule
Gemini 3.1 Pro     & 72.9 & \textbf{89.5}$^{***}$ & 33.4 & 33.0 \\
GPT-5.4            & 68.3 & \textbf{87.6}$^{***}$ & 31.0 & 29.7 \\
Qwen3.5-27B        & 40.5 & \textbf{46.3}$^{*}$   & 24.1 & 23.4 \\
Qwen3-VL-32B       & 31.4 & 30.8                  & 22.3 & 23.1 \\
InternVL-14B\,(t)  & 26.8 & 22.2                  & 3.0  & \textbf{5.0}$^{*}$ \\
InternVL-14B\,(n)  & 24.5 & 21.7                  & 2.5  & 3.9 \\
\bottomrule
\end{tabular}
}
\caption{Cascade analysis. The S2 and mIoU columns report unconditional performance; c-S2 and c-mIoU are the conditional metrics. Bold marks conditional metrics that differ significantly from their unconditional counterparts.
$^{*}\,p{<}.05$, $^{**}\,p{<}.01$, $^{***}\,p{<}.001$.}
\label{tab:cascade}
\end{table}

We test whether correct performance on one stage improves performance on the next, using two conditional metrics. $\text{c-S2} = P(\text{S2} \mid \text{S1 correct})$ measures how correct naming relates to visual recognition, and $\text{c-mIoU} = \mathbb{E}[\text{IoU} \mid \text{S2 correct}]$ measures how correct visual recognition relates to temporal localization. Both are computed under \textit{Carry}, where the model sees its own previous answer. Table~\ref{tab:cascade} reports both, with significance markers against the unconditional S2 and mIoU baselines.

\paragraph{S1 to S2: correct naming helps the strongest models.}

For Gemini and GPT, once they correctly name a concept, their chance of picking the right video moment increases by about 18 points. The improvement shrinks across the open-source models: Qwen3.5-27B picks up about a third of it; Qwen3-VL-32B and both InternVL configurations show none. The smallest model, InternVL-14B (t), inverts the trend: c-S2 is below the unconditional S2 rate.

\paragraph{S2 to S3: correct recognition rarely improves localization.}

For five models, choosing the right S2 moment provides no measurable benefit when the model must predict a start-end span on a fresh clip. Temporal localization quality is essentially the same regardless of whether S2 was correct. The one exception is InternVL with inference-time reasoning, with c-mIoU two points above its baseline, but the baseline is so low that the effect is not practically meaningful.

\subsection{Modality Dependence Is Concept-Specific}
\label{sec:modality}

\begin{figure}[!htb]
  \includegraphics[width=1\columnwidth]{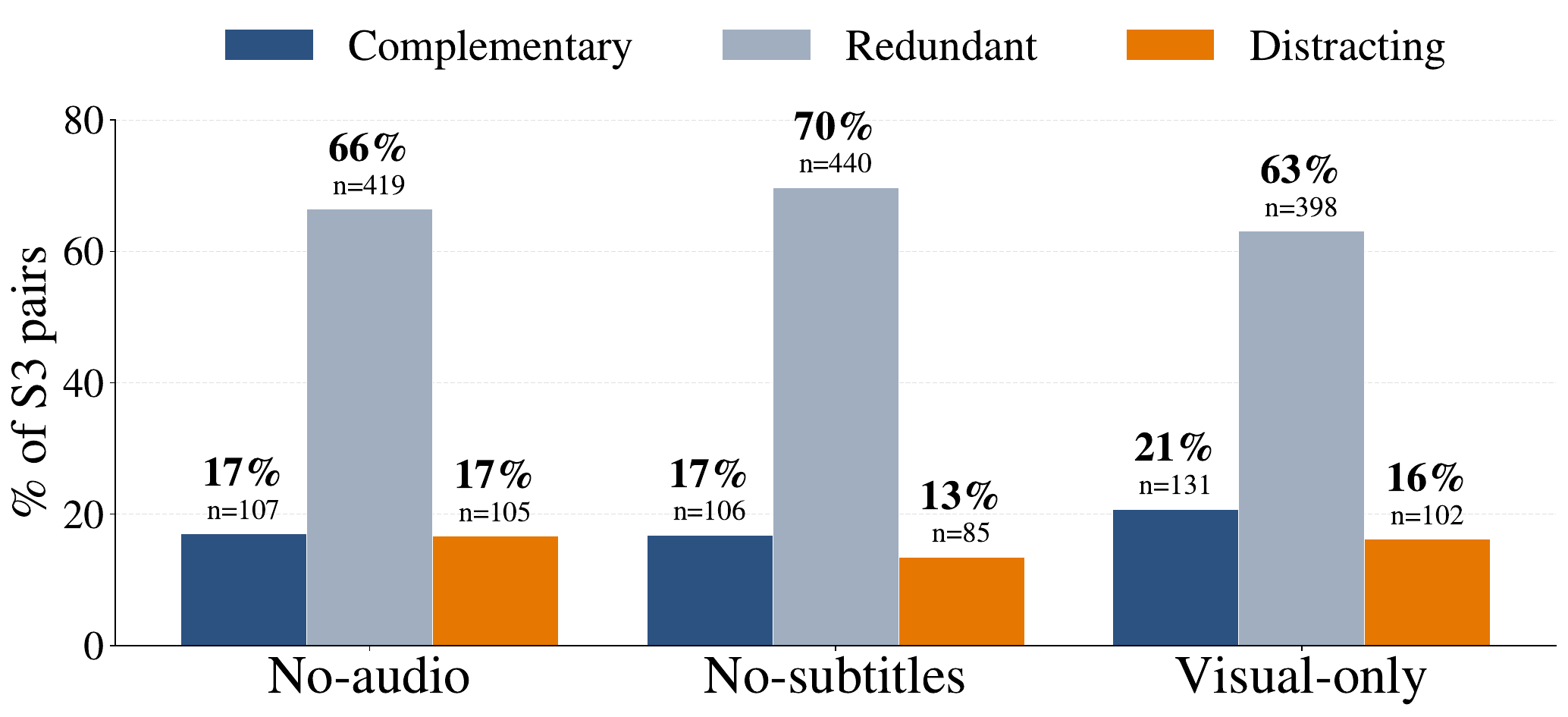}
  \caption{Share of S3 video pairs ($n{=}631$) in each modality role (Complementary, Redundant, Distracting; defined in \S\ref{sec:modality}) for the three ablations. The mIoU change $\Delta$ is averaged across three models before binning. Complementary and Distracting shares are roughly balanced and cancel in the mean mIoU.}
  \label{fig:modality_overview}
\end{figure}

For each video we compute $\Delta$, the change in mIoU in percentage points when a modality is removed, averaged over the three ablation-instrumented models. A modality is \textit{Complementary} for that video when removing it costs at least 5 points ($\Delta \leq -5$), \textit{Distracting} when removing it gains at least 5 points ($\Delta \geq +5$), and \textit{Redundant} otherwise ($|\Delta| < 5$). The 5-point cut is a reporting threshold, not a statistical test; Appendix~\ref{sec:app_threshold} repeats the classification at 3 and 7 points and both claims below hold at every cut. Figure~\ref{fig:modality_overview} shows the role shares when audio, subtitles, or both are removed. The audio and subtitle modalities are redundant for temporal localization in more than 60\% of videos, indicating that removing them has little or no effect on performance. However, these modalities are complementary in approximately 17\%–21\% of the videos, where their removal degrades localization performance. In contrast, for about 13\%–17\% of the videos, removing these modalities improves performance, suggesting that they introduce distracting or noisy information. Next, we provide further insights into modality dependency.

\paragraph{Removing both audio and subtitles is most damaging to Games and Music.}

\begin{figure}[!htb]
  \centering
  \includegraphics[width=1\columnwidth]{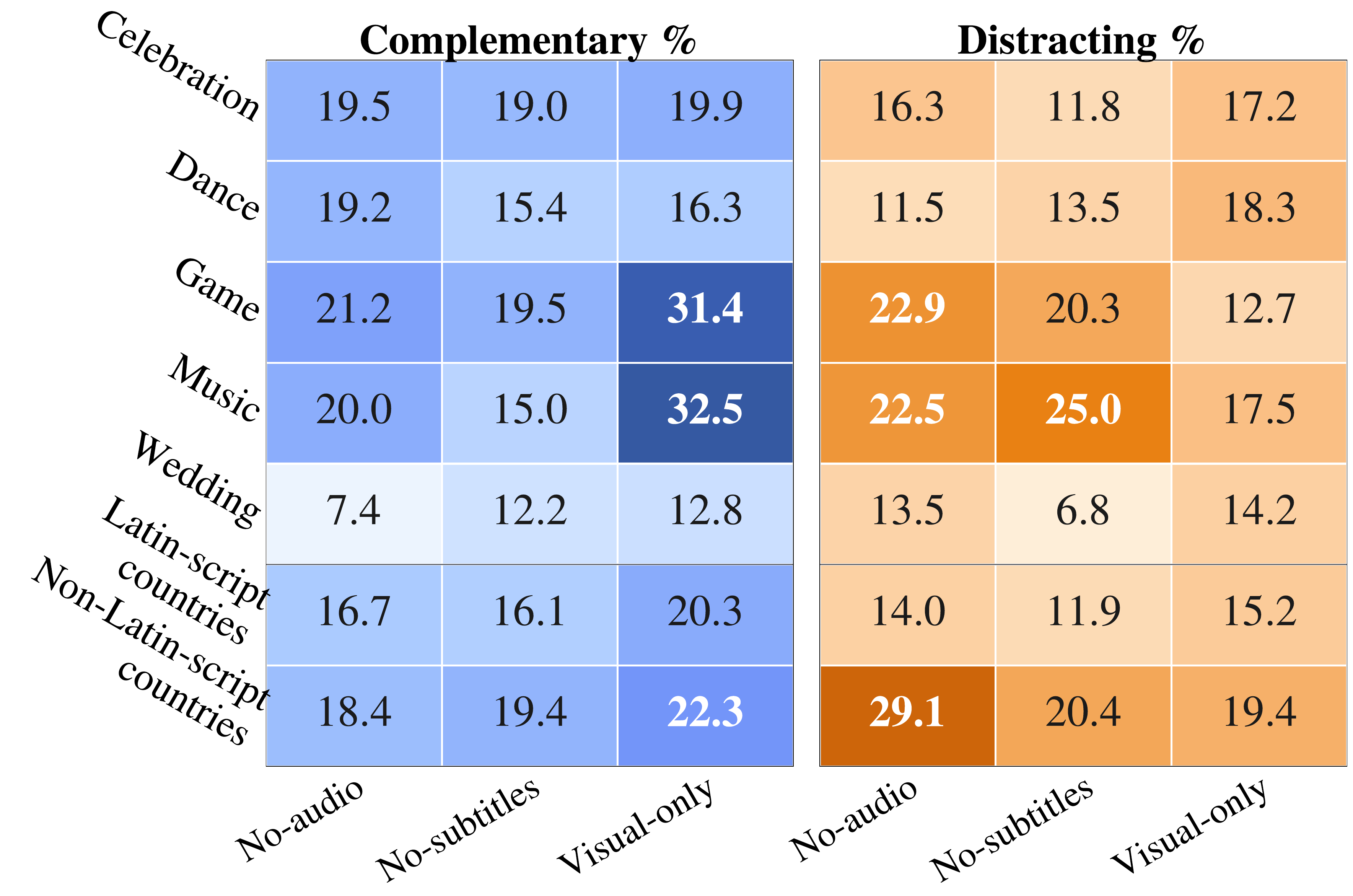}
  \caption{Modality roles across all S3 video pairs, segmented by category (top 5 rows) and country writing-system clusters (bottom 2 rows).}
\label{fig:modality_categcluster}
\end{figure}

Figure~\ref{fig:modality_categcluster} shows the per-category proportions of videos in which these modalities play complementary or distracting roles, respectively. As shown in the top five rows, incorporating both modalities improves performance for slightly more than 30\% of the videos in the Game and Music categories. In the Game category, subtitles and audio cues such as commentary, callouts, and on-screen score updates provide strong temporal signals that help localize relevant moments more precisely. Similarly, in the Music category, the alignment between audio content (e.g., songs or background music) and on-screen textual cues such as lyrics, captions, and titles substantially improves temporal localization accuracy. In these cases, removing either modality often leads to noticeable performance degradation, indicating that the modalities provide complementary information. In contrast, for other video categories, visual modality alone is distinctive enough for most videos. In some cases, audio and subtitle modalities are only weakly correlated with the target moments. For example, videos in the Wedding and Celebration categories are often accompanied by ambient background noise, music, or unrelated conversations that provide limited temporal cues and may even negatively affect localization performance. Similarly, in the Dance category, the accompanying music often persists throughout the entire video and provides limited cues for temporal localization.


\paragraph{Non-Latin-script countries show a higher Distracting share for audio.}

29\% of the videos from Cambodia, Myanmar, and Thailand have audio rated as \textit{Distracting}, compared with 14\% for Indonesia, Malaysia, the Philippines, and Vietnam, as shown in Figure~\ref{fig:modality_categcluster}. Subtitles follow the same pattern with a weaker effect. The \textit{Distracting} cases are dominated by clips featuring traditional solo instruments that play continuously across the sub-event. Thus, the audio lacks a reliable temporal cue for the moment the question asks about, acting more as noise than as signal.

\subsection{Human Evaluation}
\label{sec:human}

\begin{table}[!htb]
\centering
\small
\setlength{\tabcolsep}{2pt}
\resizebox{0.9\columnwidth}{!}{%
\begin{tabular}{lcccc}
\toprule
                 & \multicolumn{2}{c}{S1 (\%)} & \multicolumn{2}{c}{S2 (\%)} \\
\cmidrule(lr){2-3} \cmidrule(lr){4-5}
Rater            & Native & Non-native & Native & Non-native \\
\midrule
Expert           & 58.4   & 21.3       & 51.9   & 28.7 \\
Non-expert       & 39.2   & 27.9       & 45.2   & 31.5 \\
\bottomrule
\end{tabular}
}
\caption{Human evaluation accuracy (\%) on S1 and S2, by rater group (Expert vs Non-expert) and role (Native vs Non-native). Each cell averages over the seven country-level mean scores. More in Appendix~\ref{sec:app_humaneval}.}
\label{tab:human_eval}
\end{table}

We test whether CMB requires country-specific cultural knowledge, not generic regional cues, with a 14-rater human study (two raters per country). All raters are native speakers of their country's primary language and fluent in English at university level. They are categorized into two groups: \textit{Expert}, who are actively involved in cultural studies through research or co-curricular work, and \textit{Non-expert}, who are not. Each rater answered S1 and S2 in two roles. As \textit{Native}, they evaluated concepts from their own country. As \textit{Non-native}, they evaluated concepts from a neighboring country. The seven countries form a cyclic order. One rater per country received concepts from the next country; the other, from the country after that. About 20\% of items were double-rated for inter-rater reliability. The study covers all 306 concepts. We focus on two contrasts that we report below.

\paragraph{Native vs Non-native: cultural knowledge is country-specific.}

Both rater groups perform far worse on a neighboring country's concepts than on their own, as shown in Table~\ref{tab:human_eval}: \textit{Expert} raters drop by 37 points on S1 and 23 on S2, while \textit{Non-expert} raters drop by 11 and 14, respectively, from a lower starting point. The sharpest case is that \textit{Expert} raters on a neighboring country score below the 25\% chance level on S1, consistent with the semantic-similarity distractors misleading even \textit{Expert} raters. This indicates that CMB tests for country-specific knowledge rather than regional cues.

\paragraph{Expert vs Non-expert: the expertise effect is country-conditional.} 

On their own country's concepts, \textit{Expert} raters outperform \textit{Non-expert} raters by 19 points on S1 and 7 points on S2, as shown in Table~\ref{tab:human_eval}. However, the ordering flips for a neighboring country's concepts: \textit{Expert} raters score slightly lower than \textit{Non-expert} raters on both S1 and S2. The cultural knowledge that helps \textit{Expert} raters at home misleads them abroad. We also report the model mean across the six VLMs as a human-ceiling reference. On S1, the model mean is 57\%, matching \textit{Expert} raters and beating \textit{Non-expert} raters by 18 points. On S2 (Carry mode), it drops to 44\%, level with \textit{Non-expert} raters and 8 points below \textit{Expert} raters. The model's average performance is therefore stage-specific: Expert-level at naming, Non-expert-level at visual recognition, consistent with the partial S1$\to$S2 cascade.

\section{Conclusion}
\label{sec:conclusion}

Cultural video understanding is not a single capability but three, and they do not compose in current VLMs. CMB's cascade shows that correctly naming a concept helps only some models recognize it on video, and correctly recognizing it rarely helps any locate its sub-events in time. Temporal localization is the one floor every model shares; above it, which link in the cascade breaks depends on the model. A 14-rater human study confirms that the test measures country-specific cultural knowledge rather than regional cues, with model performance reaching Expert level at naming but only Non-expert level at visual recognition. CMB therefore acts as a diagnostic harness, pointing to model-specific priorities: cultural knowledge for the 14B InternVL configurations, visual recognition for the Qwen variants, and temporal localization throughout.

\section*{Limitations}

\paragraph{Cultural scope.} 

CMB covers seven SEA countries and five categories (Celebration, Dance, Game, Music, Wedding); other SEA countries (Brunei, Laos, Timor-Leste, Singapore) and traditions outside these categories are out of scope, reflecting the availability of qualified Local Annotators and of high-quality source videos. We hope the CMB framework serves as a template for broadening cultural assessment to other underrepresented regions (e.g., Africa, South America, Oceania).

\paragraph{Scale vs. depth.}

CMB prioritizes expert curation (306 concepts, 624 source videos) over scale, making it suitable for benchmark evaluation but not for VLM training or fine-tuning. Further scaling would require semi-automated pipelines or expanded annotator capacity.

\paragraph{Long-tailed distribution.} 
\label{lim:longtail}

Three distributional imbalances should temper per-category and per-country comparisons. i) Concept counts are uneven across categories (Celebration: 91 vs. Wedding: 11) and across countries (Vietnam: 55 vs. Cambodia: 28), reflecting the natural breadth of each category and the volume of online video content in each country. ii) The Set~$B$ S3 pool is shaped differently from the concept pool: concepts with many ritual sub-events (Wedding, $\sim$13 spans per concept) generate many S3 questions, while object-like concepts (Music instruments, $<$1 span per concept) generate few. iii) Sub-national diversity (minority languages, regional variants, diaspora communities) is also under-represented; per-category aggregate scores should therefore be read as averages over uneven concept and span counts, not as uniform coverage of the underlying tradition space.

\paragraph{Modality ablation scope.}

The modality ablation reruns three of the six evaluated models (Gemini 3.1 Pro, Qwen3-VL-32B, Qwen3.5-27B) due to API and compute-budget constraints; the Complementary / Redundant / Distracting role labels are derived from these three models and may differ for GPT-5.4 or InternVL.

\paragraph{Static-bias diagnostic.}

CMB does not include a single-frame baseline that would isolate how much S2 accuracy depends on a static cue versus the temporal motion in the moment. The low S2$\to$S3 transfer (\S\ref{sec:cascade}) provides indirect evidence that models rely on cues beyond concept identity, but a direct single-frame comparison is left for future work.

\paragraph{Source-video reproducibility.} 

CMB references source videos by YouTube URL. If an uploader later removes their video, it becomes inaccessible to downstream users. We maintain versioned releases of CMB so each evaluation can be tied to a specific snapshot, but the underlying video itself cannot be recovered once it is removed from YouTube. See Ethical Considerations for the full release and opt-out protocol.

\section*{Ethical Considerations}

\paragraph{Annotator compensation.} 

The two co-author Cultural Annotators (Vietnam and Malaysia) contributed without compensation. The remaining five Cultural Annotators were recruited through the authors' academic networks at universities in Southeast Asia and compensated fairly, with rates exceeding the local minimum wage and living wage standards in their respective countries, paid in each annotator's local currency. Local Annotators received a flat per-concept honorarium for the consensus and video-filtering passes, calibrated to the same minimum-wage and living-wage standards. Reviewers are co-authors and received no separate compensation for the review pass. The 14 human-study raters were each paid a fixed honorarium calibrated to the same minimum-wage and living-wage standards as the Cultural Annotators. They gave informed consent before participating and could withdraw at any time; no personally identifying information was collected, and rater identifiers in the released metadata are pseudonymous.

\paragraph{Source videos, release format, and opt-out.} 

All 624 source videos (Set A for S2 and Set B for S3) were publicly hosted on YouTube at the time of annotation. To respect the platform's terms of service and the uploader's copyright, CMB will be released on HuggingFace as a JSON file containing YouTube video IDs, S2 moment timestamps for video A, and S3 ground-truth spans for video B, accompanied by the full evaluation suite; we do not redistribute raw video bytes. The suite is released in full on publication and covers inference wrappers for all six evaluated models, scoring for all three stages under all three modes including the conditional metrics \cAcc{} and \cIoU{} and the Joint score, the cluster-bootstrap script behind every confidence interval reported here, and the analysis code for the human study. Alongside it we release the scored per-pair records for all 306 concepts, giving the per-stage outcome and IoU of every model on every item without the question text, options or gold spans, ensuring reproducibility. The benchmark items themselves are released in stages: a stratified public sample covering all seven countries and five categories is available immediately for development, and the remaining items are held back as a hidden test set so that a planned shared task can measure generalization rather than exposure; they are released once it concludes. Data is hosted on HuggingFace and code on GitHub. The CMB annotations and the evaluation suite are released under CC BY-NC-SA 4.0 for non-commercial research use. 
CMB contains no personally identifiable information: the released metadata covers only cultural concepts, timestamps, and annotation records. Users obtain the underlying videos directly from YouTube under YouTube's terms. Uploaders who wish to have a video removed from CMB may contact the authors; we will remove the corresponding entries within 7 days and reissue a versioned release.

\paragraph{Cultural sensitivity and integrity.}
The primary ethical concern in this work is the potential for misrepresentation or commodification of cultural heritage. To mitigate this, we use a human-in-the-loop curation process involving local experts from each of the seven countries in Southeast Asia. These experts verified that the questions and annotations accurately reflect the cultural semantics of the artifacts and rituals depicted, minimizing the risk of imposing external or Western-centric interpretations on the data. We caution users that culture is fluid and multifaceted: our annotations represent a valid interpretation, not necessarily the only interpretation.

\paragraph{Writing.}

In this project, we used GitHub Copilot for implementation and Claude to help revise our human-authored paper drafts.

\section*{Acknowledgements} 
We thank our Annotators and Reviewers for their careful annotation and validation work: Palakorn Achananuparp, Yochi Okta Andrawina, Rin Rath, Marfito Bryan Jimenez, Winshwesin Oo, Yibin Lai, Rimmon Saloman Bhosale, Sompassorn Ruksomboonde, Thura Aung, Jimson Paulo Layacan, Nimol Thuon, Thieu Khang Nguyen, Kian Yu Gan, Azra Randjani, Lady Pawitra, Rachana, Shaira Lee Pabalan, Sydney Khemtonglang, and a few more who prefer not to be disclosed.

This research is supported by the Ministry of Education, Singapore, under its Academic Research Fund Tier 2 (Proposal ID: T2EP20222-0047) and Academic Research Fund (AcRF) Tier 1 grant (Proposal ID: 25-SIS-SMU-004). Any opinions, findings, and conclusions or recommendations expressed in this material are those of the authors and do not reflect the views of the Ministry of Education, Singapore.

\bibliography{custom}

@inproceedings{satar-etal-2025-seeing,
    title = "Seeing Culture: A Benchmark for Visual Reasoning and Grounding",
    author = "Satar, Burak  and
      Ma, Zhixin  and
      Irawan, Patrick Amadeus  and
      Mulyawan, Wilfried Ariel  and
      Jiang, Jing  and
      Lim, Ee-Peng  and
      Ngo, Chong-Wah",
    editor = "Christodoulopoulos, Christos  and
      Chakraborty, Tanmoy  and
      Rose, Carolyn  and
      Peng, Violet",
    booktitle = "Proceedings of the 2025 Conference on Empirical Methods in Natural Language Processing",
    month = nov,
    year = "2025",
    address = "Suzhou, China",
    publisher = "Association for Computational Linguistics",
    url = "https://aclanthology.org/2025.emnlp-main.1131/",
    doi = "10.18653/v1/2025.emnlp-main.1131",
    pages = "22238--22254",
    ISBN = "979-8-89176-332-6"
}

@inproceedings{chen-etal-2025-videovista,
    title = "{V}ideo{V}ista-{C}ultural{L}ingo: 360{\textdegree} Horizons-Bridging Cultures, Languages, and Domains in Video Comprehension",
    author = "Chen, Xinyu  and
      Li, Yunxin  and
      Shi, Haoyuan  and
      Hu, Baotian  and
      Luo, Wenhan  and
      Wang, Yaowei  and
      Zhang, Min",
    editor = "Che, Wanxiang  and
      Nabende, Joyce  and
      Shutova, Ekaterina  and
      Pilehvar, Mohammad Taher",
    booktitle = "Proceedings of the 63rd Annual Meeting of the Association for Computational Linguistics (Volume 1: Long Papers)",
    month = jul,
    year = "2025",
    address = "Vienna, Austria",
    publisher = "Association for Computational Linguistics",
    url = "https://aclanthology.org/2025.acl-long.1315/",
    doi = "10.18653/v1/2025.acl-long.1315",
    pages = "27102--27128",
    ISBN = "979-8-89176-251-0",
}

@inproceedings{shafique-etal-2025-culturally,
    title = "A Culturally-diverse Multilingual Multimodal Video Benchmark {\&} Model",
    author = "Shafique, Bhuiyan Sanjid  and
      Vayani, Ashmal  and
      Maaz, Muhammad  and
      Rasheed, Hanoona Abdul  and
      Dissanayake, Dinura  and
      Kurpath, Mohammed Irfan  and
      Hmaiti, Yahya  and
      Inoue, Go  and
      Lahoud, Jean  and
      Rashid, Md. Safirur  and
      Quasem, Shadid Intisar  and
      Fatima, Maheen  and
      Vidal, Franco  and
      Maslych, Mykola  and
      More, Ketan Pravin  and
      Baliah, Sanoojan  and
      Watawana, Hasindri  and
      Li, Yuhao  and
      Farestam, Fabian  and
      Schaller, Leon  and
      Tymtsiv, Roman  and
      Weber, Simon  and
      Cholakkal, Hisham  and
      Laptev, Ivan  and
      Satoh, Shin{'}ichi  and
      Felsberg, Michael  and
      Shah, Mubarak  and
      Khan, Salman  and
      Khan, Fahad Shahbaz",
    editor = "Christodoulopoulos, Christos  and
      Chakraborty, Tanmoy  and
      Rose, Carolyn  and
      Peng, Violet",
    booktitle = "Proceedings of the 2025 Conference on Empirical Methods in Natural Language Processing",
    month = nov,
    year = "2025",
    address = "Suzhou, China",
    publisher = "Association for Computational Linguistics",
    url = "https://aclanthology.org/2025.emnlp-main.1012/",
    doi = "10.18653/v1/2025.emnlp-main.1012",
    pages = "20009--20033",
    ISBN = "979-8-89176-332-6"
}

@misc{liu2025culturevlm,
      title={{CultureVLM}: Characterizing and Improving Cultural Understanding of Vision-Language Models for over 100 Countries}, 
      author={Shudong Liu and Yiqiao Jin and Cheng Li and Derek F. Wong and Qingsong Wen and Lichao Sun and Haipeng Chen and Xing Xie and Jindong Wang},
      year={2025},
      eprint={2501.01282},
      archivePrefix={arXiv},
      primaryClass={cs.AI},
      url={https://arxiv.org/abs/2501.01282}, 
}

@misc{schneider2025gimmickgloballyinclusive,
      title={{GIMMICK} -- Globally Inclusive Multimodal Multitask Cultural Knowledge Benchmarking}, 
      author={Florian Schneider and Carolin Holtermann and Chris Biemann and Anne Lauscher},
      year={2025},
      eprint={2502.13766},
      archivePrefix={arXiv},
      primaryClass={cs.CL},
      url={https://arxiv.org/abs/2502.13766}, 
}

@inproceedings{
mogrovejo2024cvqa,
title={{CVQA}: Culturally-diverse Multilingual Visual Question Answering Benchmark},
author={David Orlando Romero Mogrovejo and Chenyang Lyu and Haryo Akbarianto Wibowo and Santiago G{\'o}ngora and Aishik Mandal and Sukannya Purkayastha and Jesus-German Ortiz-Barajas and Emilio Villa Cueva and Jinheon Baek and Soyeong Jeong and Injy Hamed and Zheng Xin Yong and Zheng Wei Lim and Paula M{\'o}nica Silva and Jocelyn Dunstan and M{\'e}lanie Jouitteau and David LE MEUR and Joan Nwatu and Ganzorig Batnasan and Munkh-Erdene Otgonbold and Munkhjargal Gochoo and Guido Ivetta and Luciana Benotti and Laura Alonso Alemany and Hern{\'a}n Maina and Jiahui Geng and Tiago Timponi Torrent and Frederico Belcavello and Marcelo Viridiano and Jan Christian Blaise Cruz and Dan John Velasco and Oana Ignat and Zara Burzo and Chenxi Whitehouse and Artem Abzaliev and Teresa Clifford and Gr{\'a}inne Caulfield and Teresa Lynn and Christian Salamea-Palacios and Vladimir Araujo and Yova Kementchedjhieva and Mihail Minkov Mihaylov and Israel Abebe Azime and Henok Biadglign Ademtew and Bontu Fufa Balcha and Naome A Etori and David Ifeoluwa Adelani and Rada Mihalcea and Atnafu Lambebo Tonja and Maria Camila Buitrago Cabrera and Gisela Vallejo and Holy Lovenia and Ruochen Zhang and Marcos Estecha-Garitagoitia and Mario Rodr{\'\i}guez-Cantelar and Toqeer Ehsan and Rendi Chevi and Muhammad Farid Adilazuarda and Ryandito Diandaru and Samuel Cahyawijaya and Fajri Koto and Tatsuki Kuribayashi and Haiyue Song and Aditya Nanda Kishore Khandavally and Thanmay Jayakumar and Raj Dabre and Mohamed Fazli Mohamed Imam and Kumaranage Ravindu Yasas Nagasinghe and Alina Dragonetti and Luis Fernando D'Haro and Olivier NIYOMUGISHA and Jay Gala and Pranjal A Chitale and Fauzan Farooqui and Thamar Solorio and Alham Fikri Aji},
booktitle={The Thirty-eight Conference on Neural Information Processing Systems Datasets and Benchmarks Track},
year={2024},
url={https://openreview.net/forum?id=E18kRXTGmV}
}

@inproceedings{nayak-etal-2024-benchmarking_CulturalVQA,
    title = "Benchmarking Vision Language Models for Cultural Understanding",
    author = "Nayak, Shravan  and
      Jain, Kanishk  and
      Awal, Rabiul  and
      Reddy, Siva  and
      Steenkiste, Sjoerd Van  and
      Hendricks, Lisa Anne  and
      Stanczak, Karolina  and
      Agrawal, Aishwarya",
    editor = "Al-Onaizan, Yaser  and
      Bansal, Mohit  and
      Chen, Yun-Nung",
    booktitle = "Proceedings of the 2024 Conference on Empirical Methods in Natural Language Processing",
    month = nov,
    year = "2024",
    address = "Miami, Florida, USA",
    publisher = "Association for Computational Linguistics",
    url = "https://aclanthology.org/2024.emnlp-main.329/",
    doi = "10.18653/v1/2024.emnlp-main.329",
    pages = "5769--5790"
}

@inproceedings{2024_seavqa,
    title = "{SEA}-{VQA}: {S}outheast {A}sian Cultural Context Dataset For Visual Question Answering",
    author = "Urailertprasert, Norawit  and
      Limkonchotiwat, Peerat  and
      Suwajanakorn, Supasorn  and
      Nutanong, Sarana",
    editor = "Gu, Jing  and
      Fu, Tsu-Jui (Ray)  and
      Hudson, Drew  and
      Celikyilmaz, Asli  and
      Wang, William",
    booktitle = "Proceedings of the 3rd Workshop on Advances in Language and Vision Research (ALVR)",
    month = aug,
    year = "2024",
    address = "Bangkok, Thailand",
    publisher = "Association for Computational Linguistics",
    url = "https://aclanthology.org/2024.alvr-1.15/",
    doi = "10.18653/v1/2024.alvr-1.15",
    pages = "173--185"
}

@misc{nie2025chinesevideobench,
      title={{ChineseVideoBench}: Benchmarking Multi-modal Large Models for Chinese Video Question Answering}, 
      author={Yuxiang Nie and Han Wang and Yongjie Ye and Haiyang Yu and Weitao Jia and Tao Zeng and Hao Feng and Xiang Fei and Yang Li and Xiaohui Lv and Guozhi Tang and Jingqun Tang and Jinghui Lu and Zehui Dai and Jiacong Wang and Dingkang Yang and An-Lan Wang and Can Huang},
      year={2025},
      eprint={2511.18399},
      archivePrefix={arXiv},
      primaryClass={cs.CV},
      url={https://arxiv.org/abs/2511.18399}, 
}

@inproceedings{liu-etal-2021-visually_MaRVL,
    title = "Visually Grounded Reasoning across Languages and Cultures",
    author = "Liu, Fangyu  and
      Bugliarello, Emanuele  and
      Ponti, Edoardo Maria  and
      Reddy, Siva  and
      Collier, Nigel  and
      Elliott, Desmond",
    booktitle = "Proceedings of the 2021 Conference on Empirical Methods in Natural Language Processing",
    month = nov,
    year = "2021",
    address = "Online and Punta Cana, Dominican Republic",
    publisher = "Association for Computational Linguistics",
    url = "https://aclanthology.org/2021.emnlp-main.818",
    pages = "10467--10485"
}

@misc{2025gpt5,
      title={{OpenAI GPT-5} System Card}, 
      author={Aaditya Singh and Adam Fry and Adam Perelman and Adam Tart and Adi Ganesh and Ahmed El-Kishky and Aidan McLaughlin and Aiden Low and AJ Ostrow and Akhila Ananthram and Akshay Nathan and Alan Luo and Alec Helyar and Aleksander Madry and Aleksandr Efremov and Aleksandra Spyra and Alex Baker-Whitcomb and Alex Beutel and Alex Karpenko and Alex Makelov and Alex Neitz and Alex Wei and Alexandra Barr and Alexandre Kirchmeyer and Alexey Ivanov and Alexi Christakis and Alistair Gillespie and Allison Tam and Ally Bennett and Alvin Wan and Alyssa Huang and Amy McDonald Sandjideh and Amy Yang and Ananya Kumar and Andre Saraiva and Andrea Vallone and Andrei Gheorghe and Andres Garcia Garcia and Andrew Braunstein and Andrew Liu and Andrew Schmidt and Andrey Mereskin and Andrey Mishchenko and Andy Applebaum and Andy Rogerson and Ann Rajan and Annie Wei and Anoop Kotha and Anubha Srivastava and Anushree Agrawal and Arun Vijayvergiya and Ashley Tyra and Ashvin Nair and Avi Nayak and Ben Eggers and Bessie Ji and Beth Hoover and Bill Chen and Blair Chen and Boaz Barak and Borys Minaiev and Botao Hao and Bowen Baker and Brad Lightcap and Brandon McKinzie and Brandon Wang and Brendan Quinn and Brian Fioca and Brian Hsu and Brian Yang and Brian Yu and Brian Zhang and Brittany Brenner and Callie Riggins Zetino and Cameron Raymond and Camillo Lugaresi and Carolina Paz and Cary Hudson and Cedric Whitney and Chak Li and Charles Chen and Charlotte Cole and Chelsea Voss and Chen Ding and Chen Shen and Chengdu Huang and Chris Colby and Chris Hallacy and Chris Koch and Chris Lu and Christina Kaplan and Christina Kim and CJ Minott-Henriques and Cliff Frey and Cody Yu and Coley Czarnecki and Colin Reid and Colin Wei and Cory Decareaux and Cristina Scheau and Cyril Zhang and Cyrus Forbes and Da Tang and Dakota Goldberg and Dan Roberts and Dana Palmie and Daniel Kappler and Daniel Levine and Daniel Wright and Dave Leo and David Lin and David Robinson and Declan Grabb and Derek Chen and Derek Lim and Derek Salama and Dibya Bhattacharjee and Dimitris Tsipras and Dinghua Li and Dingli Yu and DJ Strouse and Drew Williams and Dylan Hunn and Ed Bayes and Edwin Arbus and Ekin Akyurek and Elaine Ya Le and Elana Widmann and Eli Yani and Elizabeth Proehl and Enis Sert and Enoch Cheung and Eri Schwartz and Eric Han and Eric Jiang and Eric Mitchell and Eric Sigler and Eric Wallace and Erik Ritter and Erin Kavanaugh and Evan Mays and Evgenii Nikishin and Fangyuan Li and Felipe Petroski Such and Filipe de Avila Belbute Peres and Filippo Raso and Florent Bekerman and Foivos Tsimpourlas and Fotis Chantzis and Francis Song and Francis Zhang and Gaby Raila and Garrett McGrath and Gary Briggs and Gary Yang and Giambattista Parascandolo and Gildas Chabot and Grace Kim and Grace Zhao and Gregory Valiant and Guillaume Leclerc and Hadi Salman and Hanson Wang and Hao Sheng and Haoming Jiang and Haoyu Wang and Haozhun Jin and Harshit Sikchi and Heather Schmidt and Henry Aspegren and Honglin Chen and Huida Qiu and Hunter Lightman and Ian Covert and Ian Kivlichan and Ian Silber and Ian Sohl and Ibrahim Hammoud and Ignasi Clavera and Ikai Lan and Ilge Akkaya and Ilya Kostrikov and Irina Kofman and Isak Etinger and Ishaan Singal and Jackie Hehir and Jacob Huh and Jacqueline Pan and Jake Wilczynski and Jakub Pachocki and James Lee and James Quinn and Jamie Kiros and Janvi Kalra and Jasmyn Samaroo and Jason Wang and Jason Wolfe and Jay Chen and Jay Wang and Jean Harb and Jeffrey Han and Jeffrey Wang and Jennifer Zhao and Jeremy Chen and Jerene Yang and Jerry Tworek and Jesse Chand and Jessica Landon and Jessica Liang and Ji Lin and Jiancheng Liu and Jianfeng Wang and Jie Tang and Jihan Yin and Joanne Jang and Joel Morris and Joey Flynn and Johannes Ferstad and Johannes Heidecke and John Fishbein and John Hallman and Jonah Grant and Jonathan Chien and Jonathan Gordon and Jongsoo Park and Jordan Liss and Jos Kraaijeveld and Joseph Guay and Joseph Mo and Josh Lawson and Josh McGrath and Joshua Vendrow and Joy Jiao and Julian Lee and Julie Steele and Julie Wang and Junhua Mao and Kai Chen and Kai Hayashi and Kai Xiao and Kamyar Salahi and Kan Wu and Karan Sekhri and Karan Sharma and Karan Singhal and Karen Li and Kenny Nguyen and Keren Gu-Lemberg and Kevin King and Kevin Liu and Kevin Stone and Kevin Yu and Kristen Ying and Kristian Georgiev and Kristie Lim and Kushal Tirumala and Kyle Miller and Lama Ahmad and Larry Lv and Laura Clare and Laurance Fauconnet and Lauren Itow and Lauren Yang and Laurentia Romaniuk and Leah Anise and Lee Byron and Leher Pathak and Leon Maksin and Leyan Lo and Leyton Ho and Li Jing and Liang Wu and Liang Xiong and Lien Mamitsuka and Lin Yang and Lindsay McCallum and Lindsey Held and Liz Bourgeois and Logan Engstrom and Lorenz Kuhn and Louis Feuvrier and Lu Zhang and Lucas Switzer and Lukas Kondraciuk and Lukasz Kaiser and Manas Joglekar and Mandeep Singh and Mandip Shah and Manuka Stratta and Marcus Williams and Mark Chen and Mark Sun and Marselus Cayton and Martin Li and Marvin Zhang and Marwan Aljubeh and Matt Nichols and Matthew Haines and Max Schwarzer and Mayank Gupta and Meghan Shah and Melody Y. Guan and Melody Huang and Meng Dong and Mengqing Wang and Mia Glaese and Micah Carroll and Michael Lampe and Michael Malek and Michael Sharman and Michael Zhang and Michele Wang and Michelle Pokrass and Mihai Florian and Mikhail Pavlov and Miles Wang and Ming Chen and Mingxuan Wang and Minnia Feng and Mo Bavarian and Molly Lin and Moose Abdool and Mostafa Rohaninejad and Nacho Soto and Natalie Staudacher and Natan LaFontaine and Nathan Marwell and Nelson Liu and Nick Preston and Nick Turley and Nicklas Ansman and Nicole Blades and Nikil Pancha and Nikita Mikhaylin and Niko Felix and Nikunj Handa and Nishant Rai and Nitish Keskar and Noam Brown and Ofir Nachum and Oleg Boiko and Oleg Murk and Olivia Watkins and Oona Gleeson and Pamela Mishkin and Patryk Lesiewicz and Paul Baltescu and Pavel Belov and Peter Zhokhov and Philip Pronin and Phillip Guo and Phoebe Thacker and Qi Liu and Qiming Yuan and Qinghua Liu and Rachel Dias and Rachel Puckett and Rahul Arora and Ravi Teja Mullapudi and Raz Gaon and Reah Miyara and Rennie Song and Rishabh Aggarwal and RJ Marsan and Robel Yemiru and Robert Xiong and Rohan Kshirsagar and Rohan Nuttall and Roman Tsiupa and Ronen Eldan and Rose Wang and Roshan James and Roy Ziv and Rui Shu and Ruslan Nigmatullin and Saachi Jain and Saam Talaie and Sam Altman and Sam Arnesen and Sam Toizer and Sam Toyer and Samuel Miserendino and Sandhini Agarwal and Sarah Yoo and Savannah Heon and Scott Ethersmith and Sean Grove and Sean Taylor and Sebastien Bubeck and Sever Banesiu and Shaokyi Amdo and Shengjia Zhao and Sherwin Wu and Shibani Santurkar and Shiyu Zhao and Shraman Ray Chaudhuri and Shreyas Krishnaswamy and Shuaiqi and Xia and Shuyang Cheng and Shyamal Anadkat and Simón Posada Fishman and Simon Tobin and Siyuan Fu and Somay Jain and Song Mei and Sonya Egoian and Spencer Kim and Spug Golden and SQ Mah and Steph Lin and Stephen Imm and Steve Sharpe and Steve Yadlowsky and Sulman Choudhry and Sungwon Eum and Suvansh Sanjeev and Tabarak Khan and Tal Stramer and Tao Wang and Tao Xin and Tarun Gogineni and Taya Christianson and Ted Sanders and Tejal Patwardhan and Thomas Degry and Thomas Shadwell and Tianfu Fu and Tianshi Gao and Timur Garipov and Tina Sriskandarajah and Toki Sherbakov and Tomek Korbak and Tomer Kaftan and Tomo Hiratsuka and Tongzhou Wang and Tony Song and Tony Zhao and Troy Peterson and Val Kharitonov and Victoria Chernova and Vineet Kosaraju and Vishal Kuo and Vitchyr Pong and Vivek Verma and Vlad Petrov and Wanning Jiang and Weixing Zhang and Wenda Zhou and Wenlei Xie and Wenting Zhan and Wes McCabe and Will DePue and Will Ellsworth and Wulfie Bain and Wyatt Thompson and Xiangning Chen and Xiangyu Qi and Xin Xiang and Xinwei Shi and Yann Dubois and Yaodong Yu and Yara Khakbaz and Yifan Wu and Yilei Qian and Yin Tat Lee and Yinbo Chen and Yizhen Zhang and Yizhong Xiong and Yonglong Tian and Young Cha and Yu Bai and Yu Yang and Yuan Yuan and Yuanzhi Li and Yufeng Zhang and Yuguang Yang and Yujia Jin and Yun Jiang and Yunyun Wang and Yushi Wang and Yutian Liu and Zach Stubenvoll and Zehao Dou and Zheng Wu and Zhigang Wang},
      year={2026},
      eprint={2601.03267},
      archivePrefix={arXiv},
      primaryClass={cs.CL},
      url={https://arxiv.org/abs/2601.03267}, 
}

@misc{corr_kviscuit,
      title={Evaluating Visual and Cultural Interpretation: The {K-V}iscuit Benchmark with Human-{VLM} Collaboration}, 
      author={Yujin Baek and ChaeHun Park and Jaeseok Kim and Yu-Jung Heo and Du-Seong Chang and Jaegul Choo},
      year={2024},
      eprint={2406.16469},
      archivePrefix={arXiv},
      primaryClass={cs.CL},
      url={https://arxiv.org/abs/2406.16469}, 
}

@misc{VideoNorms_2025,
      title={{VideoNorms}: Benchmarking Cultural Awareness of Video Language Models}, 
      author={Nikhil Reddy Varimalla and Yunfei Xu and Meng Fan Wang and Arkadiy Saakyan and Smaranda Muresan},
      year={2026},
      eprint={2510.08543},
      archivePrefix={arXiv},
      primaryClass={cs.CV},
      url={https://arxiv.org/abs/2510.08543}, 
}

@article{Fu2024VideoMMETF,
  title={{Video-MME}: The First-Ever Comprehensive Evaluation Benchmark of {Multi-modal LLMs} in Video Analysis},
  author={Chaoyou Fu and Yuhan Dai and Yondong Luo and Lei Li and Shuhuai Ren and Renrui Zhang and Zihan Wang and Chenyu Zhou and Yunhang Shen and Mengdan Zhang and Peixian Chen and Yanwei Li and Shaohui Lin and Sirui Zhao and Ke Li and Tong Xu and Xiawu Zheng and Enhong Chen and Rongrong Ji and Xing Sun},
  journal={2025 IEEE/CVF Conference on Computer Vision and Pattern Recognition (CVPR)},
  year={2025},
  pages={24108-24118},
  url={https://api.semanticscholar.org/CorpusID:270199408}
}

@InProceedings{Li2024MVBenchAC,
    author    = {Li, Kunchang and Wang, Yali and He, Yinan and Li, Yizhuo and Wang, Yi and Liu, Yi and Wang, Zun and Xu, Jilan and Chen, Guo and Luo, Ping and Wang, Limin and Qiao, Yu},
    title     = {{MVBench}: A Comprehensive Multi-modal Video Understanding Benchmark},
    booktitle = {Proceedings of the IEEE/CVF Conference on Computer Vision and Pattern Recognition (CVPR)},
    month     = {June},
    year      = {2024},
    pages     = {22195-22206}
}

@misc{deepseekv3technicalreport_2025,
      title={{DeepSeek-V3} Technical Report}, 
      author={DeepSeek-AI and Aixin Liu and Bei Feng and Bing Xue and Bingxuan Wang and Bochao Wu and Chengda Lu and Chenggang Zhao and Chengqi Deng and Chenyu Zhang and Chong Ruan and Damai Dai and Daya Guo and Dejian Yang and Deli Chen and Dongjie Ji and Erhang Li and Fangyun Lin and Fucong Dai and Fuli Luo and Guangbo Hao and Guanting Chen and Guowei Li and H. Zhang and Han Bao and Hanwei Xu and Haocheng Wang and Haowei Zhang and Honghui Ding and Huajian Xin and Huazuo Gao and Hui Li and Hui Qu and J. L. Cai and Jian Liang and Jianzhong Guo and Jiaqi Ni and Jiashi Li and Jiawei Wang and Jin Chen and Jingchang Chen and Jingyang Yuan and Junjie Qiu and Junlong Li and Junxiao Song and Kai Dong and Kai Hu and Kaige Gao and Kang Guan and Kexin Huang and Kuai Yu and Lean Wang and Lecong Zhang and Lei Xu and Leyi Xia and Liang Zhao and Litong Wang and Liyue Zhang and Meng Li and Miaojun Wang and Mingchuan Zhang and Minghua Zhang and Minghui Tang and Mingming Li and Ning Tian and Panpan Huang and Peiyi Wang and Peng Zhang and Qiancheng Wang and Qihao Zhu and Qinyu Chen and Qiushi Du and R. J. Chen and R. L. Jin and Ruiqi Ge and Ruisong Zhang and Ruizhe Pan and Runji Wang and Runxin Xu and Ruoyu Zhang and Ruyi Chen and S. S. Li and Shanghao Lu and Shangyan Zhou and Shanhuang Chen and Shaoqing Wu and Shengfeng Ye and Shengfeng Ye and Shirong Ma and Shiyu Wang and Shuang Zhou and Shuiping Yu and Shunfeng Zhou and Shuting Pan and T. Wang and Tao Yun and Tian Pei and Tianyu Sun and W. L. Xiao and Wangding Zeng and Wanjia Zhao and Wei An and Wen Liu and Wenfeng Liang and Wenjun Gao and Wenqin Yu and Wentao Zhang and X. Q. Li and Xiangyue Jin and Xianzu Wang and Xiao Bi and Xiaodong Liu and Xiaohan Wang and Xiaojin Shen and Xiaokang Chen and Xiaokang Zhang and Xiaosha Chen and Xiaotao Nie and Xiaowen Sun and Xiaoxiang Wang and Xin Cheng and Xin Liu and Xin Xie and Xingchao Liu and Xingkai Yu and Xinnan Song and Xinxia Shan and Xinyi Zhou and Xinyu Yang and Xinyuan Li and Xuecheng Su and Xuheng Lin and Y. K. Li and Y. Q. Wang and Y. X. Wei and Y. X. Zhu and Yang Zhang and Yanhong Xu and Yanhong Xu and Yanping Huang and Yao Li and Yao Zhao and Yaofeng Sun and Yaohui Li and Yaohui Wang and Yi Yu and Yi Zheng and Yichao Zhang and Yifan Shi and Yiliang Xiong and Ying He and Ying Tang and Yishi Piao and Yisong Wang and Yixuan Tan and Yiyang Ma and Yiyuan Liu and Yongqiang Guo and Yu Wu and Yuan Ou and Yuchen Zhu and Yuduan Wang and Yue Gong and Yuheng Zou and Yujia He and Yukun Zha and Yunfan Xiong and Yunxian Ma and Yuting Yan and Yuxiang Luo and Yuxiang You and Yuxuan Liu and Yuyang Zhou and Z. F. Wu and Z. Z. Ren and Zehui Ren and Zhangli Sha and Zhe Fu and Zhean Xu and Zhen Huang and Zhen Zhang and Zhenda Xie and Zhengyan Zhang and Zhewen Hao and Zhibin Gou and Zhicheng Ma and Zhigang Yan and Zhihong Shao and Zhipeng Xu and Zhiyu Wu and Zhongyu Zhang and Zhuoshu Li and Zihui Gu and Zijia Zhu and Zijun Liu and Zilin Li and Ziwei Xie and Ziyang Song and Ziyi Gao and Zizheng Pan},
      year={2025},
      eprint={2412.19437},
      archivePrefix={arXiv},
      primaryClass={cs.CL},
      url={https://arxiv.org/abs/2412.19437}, 
}

@article{Guo_2025_deepseek-r1,
   title={{DeepSeek-R1} incentivizes reasoning in {LLMs} through reinforcement learning},
   volume={645},
   ISSN={1476-4687},
   url={http://dx.doi.org/10.1038/s41586-025-09422-z},
   DOI={10.1038/s41586-025-09422-z},
   number={8081},
   journal={Nature},
   publisher={Springer Science and Business Media LLC},
   author={Guo, Daya and Yang, Dejian and Zhang, Haowei and Song, Junxiao and Wang, Peiyi and Zhu, Qihao and Xu, Runxin and Zhang, Ruoyu and Ma, Shirong and Bi, Xiao and Zhang, Xiaokang and Yu, Xingkai and Wu, Yu and Wu, Z. F. and Gou, Zhibin and Shao, Zhihong and Li, Zhuoshu and Gao, Ziyi and Liu, Aixin and Xue, Bing and Wang, Bingxuan and Wu, Bochao and Feng, Bei and Lu, Chengda and Zhao, Chenggang and Deng, Chengqi and Ruan, Chong and Dai, Damai and Chen, Deli and Ji, Dongjie and Li, Erhang and Lin, Fangyun and Dai, Fucong and Luo, Fuli and Hao, Guangbo and Chen, Guanting and Li, Guowei and Zhang, H. and Xu, Hanwei and Ding, Honghui and Gao, Huazuo and Qu, Hui and Li, Hui and Guo, Jianzhong and Li, Jiashi and Chen, Jingchang and Yuan, Jingyang and Tu, Jinhao and Qiu, Junjie and Li, Junlong and Cai, J. L. and Ni, Jiaqi and Liang, Jian and Chen, Jin and Dong, Kai and Hu, Kai and You, Kaichao and Gao, Kaige and Guan, Kang and Huang, Kexin and Yu, Kuai and Wang, Lean and Zhang, Lecong and Zhao, Liang and Wang, Litong and Zhang, Liyue and Xu, Lei and Xia, Leyi and Zhang, Mingchuan and Zhang, Minghua and Tang, Minghui and Zhou, Mingxu and Li, Meng and Wang, Miaojun and Li, Mingming and Tian, Ning and Huang, Panpan and Zhang, Peng and Wang, Qiancheng and Chen, Qinyu and Du, Qiushi and Ge, Ruiqi and Zhang, Ruisong and Pan, Ruizhe and Wang, Runji and Chen, R. J. and Jin, R. L. and Chen, Ruyi and Lu, Shanghao and Zhou, Shangyan and Chen, Shanhuang and Ye, Shengfeng and Wang, Shiyu and Yu, Shuiping and Zhou, Shunfeng and Pan, Shuting and Li, S. S. and Zhou, Shuang and Wu, Shaoqing and Yun, Tao and Pei, Tian and Sun, Tianyu and Wang, T. and Zeng, Wangding and Liu, Wen and Liang, Wenfeng and Gao, Wenjun and Yu, Wenqin and Zhang, Wentao and Xiao, W. L. and An, Wei and Liu, Xiaodong and Wang, Xiaohan and Chen, Xiaokang and Nie, Xiaotao and Cheng, Xin and Liu, Xin and Xie, Xin and Liu, Xingchao and Yang, Xinyu and Li, Xinyuan and Su, Xuecheng and Lin, Xuheng and Li, X. Q. and Jin, Xiangyue and Shen, Xiaojin and Chen, Xiaosha and Sun, Xiaowen and Wang, Xiaoxiang and Song, Xinnan and Zhou, Xinyi and Wang, Xianzu and Shan, Xinxia and Li, Y. K. and Wang, Y. Q. and Wei, Y. X. and Zhang, Yang and Xu, Yanhong and Li, Yao and Zhao, Yao and Sun, Yaofeng and Wang, Yaohui and Yu, Yi and Zhang, Yichao and Shi, Yifan and Xiong, Yiliang and He, Ying and Piao, Yishi and Wang, Yisong and Tan, Yixuan and Ma, Yiyang and Liu, Yiyuan and Guo, Yongqiang and Ou, Yuan and Wang, Yuduan and Gong, Yue and Zou, Yuheng and He, Yujia and Xiong, Yunfan and Luo, Yuxiang and You, Yuxiang and Liu, Yuxuan and Zhou, Yuyang and Zhu, Y. X. and Huang, Yanping and Li, Yaohui and Zheng, Yi and Zhu, Yuchen and Ma, Yunxian and Tang, Ying and Zha, Yukun and Yan, Yuting and Ren, Z. Z. and Ren, Zehui and Sha, Zhangli and Fu, Zhe and Xu, Zhean and Xie, Zhenda and Zhang, Zhengyan and Hao, Zhewen and Ma, Zhicheng and Yan, Zhigang and Wu, Zhiyu and Gu, Zihui and Zhu, Zijia and Liu, Zijun and Li, Zilin and Xie, Ziwei and Song, Ziyang and Pan, Zizheng and Huang, Zhen and Xu, Zhipeng and Zhang, Zhongyu and Zhang, Zhen},
   year={2025},
   month=Sept, pages={633–638} }

@misc{bai2025qwen3vl,
      title={{Qwen3-VL} Technical Report}, 
      author={Shuai Bai and Yuxuan Cai and Ruizhe Chen and Keqin Chen and Xionghui Chen and Zesen Cheng and Lianghao Deng and Wei Ding and Chang Gao and Chunjiang Ge and Wenbin Ge and Zhifang Guo and Qidong Huang and Jie Huang and Fei Huang and Binyuan Hui and Shutong Jiang and Zhaohai Li and Mingsheng Li and Mei Li and Kaixin Li and Zicheng Lin and Junyang Lin and Xuejing Liu and Jiawei Liu and Chenglong Liu and Yang Liu and Dayiheng Liu and Shixuan Liu and Dunjie Lu and Ruilin Luo and Chenxu Lv and Rui Men and Lingchen Meng and Xuancheng Ren and Xingzhang Ren and Sibo Song and Yuchong Sun and Jun Tang and Jianhong Tu and Jianqiang Wan and Peng Wang and Pengfei Wang and Qiuyue Wang and Yuxuan Wang and Tianbao Xie and Yiheng Xu and Haiyang Xu and Jin Xu and Zhibo Yang and Mingkun Yang and Jianxin Yang and An Yang and Bowen Yu and Fei Zhang and Hang Zhang and Xi Zhang and Bo Zheng and Humen Zhong and Jingren Zhou and Fan Zhou and Jing Zhou and Yuanzhi Zhu and Ke Zhu},
      year={2025},
      eprint={2511.21631},
      archivePrefix={arXiv},
      primaryClass={cs.CV},
      url={https://arxiv.org/abs/2511.21631}, 
}

@misc{qwen3.5,
    title  = {{Qwen3.5}: Towards Native Multimodal Agents},
    author = {{Qwen Team}},
    month  = {February},
    year   = {2026},
    note   = {\url{https://qwen.ai/blog?id=qwen3.5}}
}

@inproceedings{Reimers2019SentenceBERTSE,
    title = "Sentence-{BERT}: Sentence Embeddings using {S}iamese {BERT}-Networks",
    author = "Reimers, Nils  and
      Gurevych, Iryna",
    editor = "Inui, Kentaro  and
      Jiang, Jing  and
      Ng, Vincent  and
      Wan, Xiaojun",
    booktitle = "Proceedings of the 2019 Conference on Empirical Methods in Natural Language Processing and the 9th International Joint Conference on Natural Language Processing (EMNLP-IJCNLP)",
    month = nov,
    year = "2019",
    address = "Hong Kong, China",
    publisher = "Association for Computational Linguistics",
    url = "https://aclanthology.org/D19-1410/",
    doi = "10.18653/v1/D19-1410",
    pages = "3982--3992"
}

@misc{2025gemini,
      title={Gemini 2.5: Pushing the Frontier with Advanced Reasoning, Multimodality, Long Context, and Next Generation Agentic Capabilities}, 
      author={Gheorghe Comanici and Eric Bieber and Mike Schaekermann and Ice Pasupat and Noveen Sachdeva and Inderjit Dhillon and Marcel Blistein and Ori Ram and Dan Zhang and Evan Rosen and Luke Marris and Sam Petulla and Colin Gaffney and Asaf Aharoni and Nathan Lintz and Tiago Cardal Pais and Henrik Jacobsson and Idan Szpektor and Nan-Jiang Jiang and Krishna Haridasan and Ahmed Omran and Nikunj Saunshi and Dara Bahri and Gaurav Mishra and Eric Chu and Toby Boyd and Brad Hekman and Aaron Parisi and Chaoyi Zhang and Kornraphop Kawintiranon and Tania Bedrax-Weiss and Oliver Wang and Ya Xu and Ollie Purkiss and Uri Mendlovic and Ilaï Deutel and Nam Nguyen and Adam Langley and Flip Korn and Lucia Rossazza and Alexandre Ramé and Sagar Waghmare and Helen Miller and Nathan Byrd and Ashrith Sheshan and Raia Hadsell and Sangnie Bhardwaj and Pawel Janus and Tero Rissa and Dan Horgan and Alvin Abdagic and Lior Belenki and James Allingham and Anima Singh and Theo Guidroz and Srivatsan Srinivasan and Herman Schmit and Kristen Chiafullo and Andre Elisseeff and Nilpa Jha and Prateek Kolhar and Leonard Berrada and Frank Ding and Xiance Si and Shrestha Basu Mallick and Franz Och and Sofia Erell and Eric Ni and Tejasi Latkar and Sherry Yang and Petar Sirkovic and Ziqiang Feng and Robert Leland and Rachel Hornung and Gang Wu and Charles Blundell and Hamidreza Alvari and Po-Sen Huang and Cathy Yip and Sanja Deur and Li Liu and Gabriela Surita and Pablo Duque and Dima Damen and Johnson Jia and Arthur Guez and Markus Mircea and Animesh Sinha and Alberto Magni and Paweł Stradomski and Tal Marian and Vlado Galić and Wenhu Chen and Hisham Husain and Achintya Singhal and Dominik Grewe and François-Xavier Aubet and Shuang Song and Lorenzo Blanco and Leland Rechis and Lewis Ho and Rich Munoz and Kelvin Zheng and Jessica Hamrick and Kevin Mather and Hagai Taitelbaum and Eliza Rutherford and Yun Lei and Kuangyuan Chen and Anand Shukla and Erica Moreira and Eric Doi and Berivan Isik and Nir Shabat and Dominika Rogozińska and Kashyap Kolipaka and Jason Chang and Eugen Vušak and Srinivasan Venkatachary and Shadi Noghabi and Tarun Bharti and Younghoon Jun and Aleksandr Zaks and Simon Green and Jeshwanth Challagundla and William Wong and Muqthar Mohammad and Dean Hirsch and Yong Cheng and Iftekhar Naim and Lev Proleev and Damien Vincent and Aayush Singh and Maxim Krikun and Dilip Krishnan and Zoubin Ghahramani and Aviel Atias and Rajeev Aggarwal and Christo Kirov and Dimitrios Vytiniotis and Christy Koh and Alexandra Chronopoulou and Pawan Dogra and Vlad-Doru Ion and Gladys Tyen and Jason Lee and Felix Weissenberger and Trevor Strohman and Ashwin Balakrishna and Jack Rae and Marko Velic and Raoul de Liedekerke and Oded Elyada and Wentao Yuan and Canoee Liu and Lior Shani and Sergey Kishchenko and Bea Alessio and Yandong Li and Richard Song and Sam Kwei and Orion Jankowski and Aneesh Pappu and Youhei Namiki and Yenai Ma and Nilesh Tripuraneni and Colin Cherry and Marissa Ikonomidis and Yu-Cheng Ling and Colin Ji and Beka Westberg and Auriel Wright and Da Yu and David Parkinson and Swaroop Ramaswamy and Jerome Connor and Soheil Hassas Yeganeh and Snchit Grover and George Kenwright and Lubo Litchev and Chris Apps and Alex Tomala and Felix Halim and Alex Castro-Ros and Zefei Li and Anudhyan Boral and Pauline Sho and Michal Yarom and Eric Malmi and David Klinghoffer and Rebecca Lin and Alan Ansell and Pradeep Kumar S and Shubin Zhao and Siqi Zuo and Adam Santoro and Heng-Tze Cheng and Solomon Demmessie and Yuchi Liu and Nicole Brichtova and Allie Culp and Nathaniel Braun and Dan Graur and Will Ng and Nikhil Mehta and Aaron Phillips and Patrik Sundberg and Varun Godbole and Fangyu Liu and Yash Katariya and David Rim and Mojtaba Seyedhosseini and Sean Ammirati and Jonas Valfridsson and Mahan Malihi and Timothy Knight and Andeep Toor and Thomas Lampe and Abe Ittycheriah and Lewis Chiang and Chak Yeung and Alexandre Fréchette and Jinmeng Rao and Huisheng Wang and Himanshu Srivastava and Richard Zhang and Rocky Rhodes and Ariel Brand and Dean Weesner and Ilya Figotin and Felix Gimeno and Rachana Fellinger and Pierre Marcenac and José Leal and Eyal Marcus and Victor Cotruta and Rodrigo Cabrera and Sheryl Luo and Dan Garrette and Vera Axelrod and Sorin Baltateanu and David Barker and Dongkai Chen and Horia Toma and Ben Ingram and Jason Riesa and Chinmay Kulkarni and Yujing Zhang and Hongbin Liu and Chao Wang and Martin Polacek and Will Wu and Kai Hui and Adrian N Reyes and Yi Su and Megan Barnes and Ishaan Malhi and Anfal Siddiqui and Qixuan Feng and Mihai Damaschin and Daniele Pighin and Andreas Steiner and Samuel Yang and Ramya Sree Boppana and Simeon Ivanov and Arun Kandoor and Aditya Shah and Asier Mujika and Da Huang and Christopher A. Choquette-Choo and Mohak Patel and Tianhe Yu and Toni Creswell and Jerry and Liu and Catarina Barros and Yasaman Razeghi and Aurko Roy and Phil Culliton and Binbin Xiong and Jiaqi Pan and Thomas Strohmann and Tolly Powell and Babi Seal and Doug DeCarlo and Pranav Shyam and Kaan Katircioglu and Xuezhi Wang and Cassidy Hardin and Immanuel Odisho and Josef Broder and Oscar Chang and Arun Nair and Artem Shtefan and Maura O'Brien and Manu Agarwal and Sahitya Potluri and Siddharth Goyal and Amit Jhindal and Saksham Thakur and Yury Stuken and James Lyon and Kristina Toutanova and Fangxiaoyu Feng and Austin Wu and Ben Horn and Alek Wang and Alex Cullum and Gabe Taubman and Disha Shrivastava and Chongyang Shi and Hamish Tomlinson and Roma Patel and Tao Tu and Ada Maksutaj Oflazer and Francesco Pongetti and Mingyao Yang and Adrien Ali Taïga and Vincent Perot and Nuo Wang Pierse and Feng Han and Yoel Drori and Iñaki Iturrate and Ayan Chakrabarti and Legg Yeung and Dave Dopson and Yi-ting Chen and Apoorv Kulshreshtha and Tongfei Guo and Philip Pham and Tal Schuster and Junquan Chen and Alex Polozov and Jinwei Xing and Huanjie Zhou and Praneeth Kacham and Doron Kukliansky and Antoine Miech and Sergey Yaroshenko and Ed Chi and Sholto Douglas and Hongliang Fei and Mathieu Blondel and Preethi Myla and Lior Madmoni and Xing Wu and Daniel Keysers and Kristian Kjems and Isabela Albuquerque and Lijun Yu and Joel D'sa and Michelle Plantan and Vlad Ionescu and Jaume Sanchez Elias and Abhirut Gupta and Manish Reddy Vuyyuru and Fred Alcober and Tong Zhou and Kaiyang Ji and Florian Hartmann and Subha Puttagunta and Hugo Song and Ehsan Amid and Anca Stefanoiu and Andrew Lee and Paul Pucciarelli and Emma Wang and Amit Raul and Slav Petrov and Isaac Tian and Valentin Anklin and Nana Nti and Victor Gomes and Max Schumacher and Grace Vesom and Alex Panagopoulos and Konstantinos Bousmalis and Daniel Andor and Josh Jacob and Yuan Zhang and Bill Rosgen and Matija Kecman and Matthew Tung and Alexandra Belias and Noah Goodman and Paul Covington and Brian Wieder and Nikita Saxena and Elnaz Davoodi and Muhuan Huang and Sharath Maddineni and Vincent Roulet and Folawiyo Campbell-Ajala and Pier Giuseppe Sessa and Xintian and Wu and Guangda Lai and Paul Collins and Alex Haig and Vytenis Sakenas and Xiaowei Xu and Marissa Giustina and Laurent El Shafey and Pichi Charoenpanit and Shefali Garg and Joshua Ainslie and Boone Severson and Montse Gonzalez Arenas and Shreya Pathak and Sujee Rajayogam and Jie Feng and Michiel Bakker and Sheng Li and Nevan Wichers and Jamie Rogers and Xinyang Geng and Yeqing Li and Rolf Jagerman and Chao Jia and Nadav Olmert and David Sharon and Matthew Mauger and Sandeep Mariserla and Hongxu Ma and Megha Mohabey and Kyuyeun Kim and Alek Andreev and Scott Pollom and Juliette Love and Vihan Jain and Priyanka Agrawal and Yannick Schroecker and Alisa Fortin and Manfred Warmuth and Ji Liu and Andrew Leach and Irina Blok and Ganesh Poomal Girirajan and Roee Aharoni and Benigno Uria and Andrei Sozanschi and Dan Goldberg and Lucian Ionita and Marco Tulio Ribeiro and Martin Zlocha and Vighnesh Birodkar and Sami Lachgar and Liangzhe Yuan and Himadri Choudhury and Matt Ginsberg and Fei Zheng and Gregory Dibb and Emily Graves and Swachhand Lokhande and Gabriel Rasskin and George-Cristian Muraru and Corbin Quick and Sandeep Tata and Pierre Sermanet and Aditya Chawla and Itay Karo and Yan Wang and Susan Zhang and Orgad Keller and Anca Dragan and Guolong Su and Ian Chou and Xi Liu and Yiqing Tao and Shruthi Prabhakara and Marc Wilson and Ruibo Liu and Shibo Wang and Georgie Evans and David Du and Alfonso Castaño and Gautam Prasad and Mona El Mahdy and Sebastian Gerlach and Machel Reid and Jarrod Kahn and Amir Zait and Thanumalayan Sankaranarayana Pillai and Thatcher Ulrich and Guanyu Wang and Jan Wassenberg and Efrat Farkash and Kiran Yalasangi and Congchao Wang and Maria Bauza and Simon Bucher and Ting Liu and Jun Yan and Gary Leung and Vikas Sindhwani and Parker Barnes and Avi Singh and Ivan Jurin and Jichuan Chang and Niket Kumar Bhumihar and Sivan Eiger and Gui Citovsky and Ben Withbroe and Zhang Li and Siyang Xue and Niccolò Dal Santo and Georgi Stoyanov and Yves Raimond and Steven Zheng and Yilin Gao and Vít Listík and Sławek Kwasiborski and Rachel Saputro and Adnan Ozturel and Ganesh Mallya and Kushal Majmundar and Ross West and Paul Caron and Jinliang Wei and Lluis Castrejon and Sharad Vikram and Deepak Ramachandran and Nikhil Dhawan and Jiho Park and Sara Smoot and George van den Driessche and Yochai Blau and Chase Malik and Wei Liang and Roy Hirsch and Cicero Nogueira dos Santos and Eugene Weinstein and Aäron van den Oord and Sid Lall and Nicholas FitzGerald and Zixuan Jiang and Xuan Yang and Dale Webster and Ali Elqursh and Aedan Pope and Georges Rotival and David Raposo and Wanzheng Zhu and Jeff Dean and Sami Alabed and Dustin Tran and Arushi Gupta and Zach Gleicher and Jessica Austin and Edouard Rosseel and Megh Umekar and Dipanjan Das and Yinghao Sun and Kai Chen and Karolis Misiunas and Xiang Zhou and Yixian Di and Alyssa Loo and Josh Newlan and Bo Li and Vinay Ramasesh and Ying Xu and Alex Chen and Sudeep Gandhe and Radu Soricut and Nikita Gupta and Shuguang Hu and Seliem El-Sayed and Xavier Garcia and Idan Brusilovsky and Pu-Chin Chen and Andrew Bolt and Lu Huang and Alex Gurney and Zhiying Zhang and Alexander Pritzel and Jarek Wilkiewicz and Bryan Seybold and Bhargav Kanagal Shamanna and Felix Fischer and Josef Dean and Karan Gill and Ross Mcilroy and Abhishek Bhowmick and Jeremy Selier and Antoine Yang and Derek Cheng and Vladimir Magay and Jie Tan and Dhriti Varma and Christian Walder and Tomas Kocisky and Ryo Nakashima and Paul Natsev and Mike Kwong and Ionel Gog and Chiyuan Zhang and Sander Dieleman and Thomas Jimma and Andrey Ryabtsev and Siddhartha Brahma and David Steiner and Dayou Du and Ante Žužul and Mislav Žanić and Mukund Raghavachari and Willi Gierke and Zeyu Zheng and Dessie Petrova and Yann Dauphin and Yuchuan Liu and Ido Kessler and Steven Hand and Chris Duvarney and Seokhwan Kim and Hyo Lee and Léonard Hussenot and Jeffrey Hui and Josh Smith and Deepali Jain and Jiawei Xia and Gaurav Singh Tomar and Keyvan Amiri and Du Phan and Fabian Fuchs and Tobias Weyand and Nenad Tomasev and Alexandra Cordell and Xin Liu and Jonathan Mallinson and Pankaj Joshi and Andy Crawford and Arun Suggala and Steve Chien and Nick Fernando and Mariella Sanchez-Vargas and Duncan Williams and Phil Crone and Xiyang Luo and Igor Karpov and Jyn Shan and Terry Thurk and Robin Strudel and Paul Voigtlaender and Piyush Patil and Tim Dozat and Ali Khodaei and Sahil Singla and Piotr Ambroszczyk and Qiyin Wu and Yifan Chang and Brian Roark and Chaitra Hegde and Tianli Ding and Angelos Filos and Zhongru Wu and André Susano Pinto and Shuang Liu and Saarthak Khanna and Aditya Pandey and Siobhan Mcloughlin and Qiujia Li and Sam Haves and Allan Zhou and Elena Buchatskaya and Isabel Leal and Peter de Boursac and Nami Akazawa and Nina Anderson and Terry Chen and Krishna Somandepalli and Chen Liang and Sheela Goenka and Stephanie Winkler and Alexander Grushetsky and Yifan Ding and Jamie Smith and Fan Ye and Jordi Pont-Tuset and Eric Li and Ruichao Li and Tomer Golany and Dawid Wegner and Tao Jiang and Omer Barak and Yuan Shangguan and Eszter Vértes and Renee Wong and Jörg Bornschein and Alex Tudor and Michele Bevilacqua and Tom Schaul and Ankit Singh Rawat and Yang Zhao and Kyriakos Axiotis and Lei Meng and Cory McLean and Jonathan Lai and Jennifer Beattie and Nate Kushman and Yaxin Liu and Blair Kutzman and Fiona Lang and Jingchen Ye and Praneeth Netrapalli and Pushkar Mishra and Myriam Khan and Megha Goel and Rob Willoughby and David Tian and Honglei Zhuang and JD Chen and Zak Tsai and Tasos Kementsietsidis and Arjun Khare and James Keeling and Keyang Xu and Nathan Waters and Florent Altché and Ashok Popat and Bhavishya Mittal and David Saxton and Dalia El Badawy and Michael Mathieu and Zheng Zheng and Hao Zhou and Nishant Ranka and Richard Shin and Qingnan Duan and Tim Salimans and Ioana Mihailescu and Uri Shaham and Ming-Wei Chang and Yannis Assael and Nishanth Dikkala and Martin Izzard and Vincent Cohen-Addad and Cat Graves and Vlad Feinberg and Grace Chung and DJ Strouse and Danny Karmon and Sahand Sharifzadeh and Zoe Ashwood and Khiem Pham and Jon Blanton and Alex Vasiloff and Jarred Barber and Mark Geller and Aurick Zhou and Fedir Zubach and Tzu-Kuo Huang and Lei Zhang and Himanshu Gupta and Matt Young and Julia Proskurnia and Ronny Votel and Valentin Gabeur and Gabriel Barcik and Aditya Tripathi and Hongkun Yu and Geng Yan and Beer Changpinyo and Filip Pavetić and Amy Coyle and Yasuhisa Fujii and Jorge Gonzalez Mendez and Tianhao Zhou and Harish Rajamani and Blake Hechtman and Eddie Cao and Da-Cheng Juan and Yi-Xuan Tan and Valentin Dalibard and Yilun Du and Natalie Clay and Kaisheng Yao and Wenhao Jia and Dimple Vijaykumar and Yuxiang Zhou and Xinyi Bai and Wei-Chih Hung and Steven Pecht and Georgi Todorov and Nikhil Khadke and Pramod Gupta and Preethi Lahoti and Arnaud Autef and Karthik Duddu and James Lee-Thorp and Alexander Bykovsky and Tautvydas Misiunas and Sebastian Flennerhag and Santhosh Thangaraj and Jed McGiffin and Zack Nado and Markus Kunesch and Andreas Noever and Amir Hertz and Marco Liang and Victor Stone and Evan Palmer and Samira Daruki and Arijit Pramanik and Siim Põder and Austin Kyker and Mina Khan and Evgeny Sluzhaev and Marvin Ritter and Avraham Ruderman and Wenlei Zhou and Chirag Nagpal and Kiran Vodrahalli and George Necula and Paul Barham and Ellie Pavlick and Jay Hartford and Izhak Shafran and Long Zhao and Maciej Mikuła and Tom Eccles and Hidetoshi Shimokawa and Kanav Garg and Luke Vilnis and Hanwen Chen and Ilia Shumailov and Kuang-Huei Lee and Abdelrahman Abdelhamed and Meiyan Xie and Vered Cohen and Ester Hlavnova and Dan Malkin and Chawin Sitawarin and James Lottes and Pauline Coquinot and Tianli Yu and Sandeep Kumar and Jingwei Zhang and Aroma Mahendru and Zafarali Ahmed and James Martens and Tao Chen and Aviel Boag and Daiyi Peng and Coline Devin and Arseniy Klimovskiy and Mary Phuong and Danny Vainstein and Jin Xie and Bhuvana Ramabhadran and Nathan Howard and Xinxin Yu and Gitartha Goswami and Jingyu Cui and Sam Shleifer and Mario Pinto and Chih-Kuan Yeh and Ming-Hsuan Yang and Sara Javanmardi and Dan Ethier and Chace Lee and Jordi Orbay and Suyog Kotecha and Carla Bromberg and Pete Shaw and James Thornton and Adi Gerzi Rosenthal and Shane Gu and Matt Thomas and Ian Gemp and Aditya Ayyar and Asahi Ushio and Aarush Selvan and Joel Wee and Chenxi Liu and Maryam Majzoubi and Weiren Yu and Jake Abernethy and Tyler Liechty and Renke Pan and Hoang Nguyen and Qiong and Hu and Sarah Perrin and Abhinav Arora and Emily Pitler and Weiyi Wang and Kaushik Shivakumar and Flavien Prost and Ben Limonchik and Jing Wang and Yi Gao and Timothee Cour and Shyamal Buch and Huan Gui and Maria Ivanova and Philipp Neubeck and Kelvin Chan and Lucy Kim and Huizhong Chen and Naman Goyal and Da-Woon Chung and Lu Liu and Yao Su and Anastasia Petrushkina and Jiajun Shen and Armand Joulin and Yuanzhong Xu and Stein Xudong Lin and Yana Kulizhskaya and Ciprian Chelba and Shobha Vasudevan and Eli Collins and Vasilisa Bashlovkina and Tony Lu and Doug Fritz and Jongbin Park and Yanqi Zhou and Chen Su and Richard Tanburn and Mikhail Sushkov and Mitchelle Rasquinha and Jinning Li and Jennifer Prendki and Yiming Li and Pallavi LV and Shriya Sharma and Hen Fitoussi and Hui Huang and Andrew Dai and Phuong Dao and Mike Burrows and Henry Prior and Danfeng Qin and Golan Pundak and Lars Lowe Sjoesund and Art Khurshudov and Zhenkai Zhu and Albert Webson and Elizabeth Kemp and Tat Tan and Saurabh Agrawal and Susie Sargsyan and Liqun Cheng and Jim Stephan and Tom Kwiatkowski and David Reid and Arunkumar Byravan and Assaf Hurwitz Michaely and Nicolas Heess and Luowei Zhou and Sonam Goenka and Viral Carpenter and Anselm Levskaya and Bo Wang and Reed Roberts and Rémi Leblond and Sharat Chikkerur and Stav Ginzburg and Max Chang and Robert Riachi and Chuqiao and Xu and Zalán Borsos and Michael Pliskin and Julia Pawar and Morgane Lustman and Hannah Kirkwood and Ankit Anand and Aditi Chaudhary and Norbert Kalb and Kieran Milan and Sean Augenstein and Anna Goldie and Laurel Prince and Karthik Raman and Yanhua Sun and Vivian Xia and Aaron Cohen and Zhouyuan Huo and Josh Camp and Seher Ellis and Lukas Zilka and David Vilar Torres and Lisa Patel and Sho Arora and Betty Chan and Jonas Adler and Kareem Ayoub and Jacky Liang and Fayaz Jamil and Jiepu Jiang and Simon Baumgartner and Haitian Sun and Yael Karov and Yaroslav Akulov and Hui Zheng and Irene Cai and Claudio Fantacci and James Rubin and Alex Rav Acha and Mengchao Wang and Nina D'Souza and Rohit Sathyanarayana and Shengyang Dai and Simon Rowe and Andrey Simanovsky and Omer Goldman and Yuheng Kuang and Xiaoyue Pan and Andrew Rosenberg and Tania Rojas-Esponda and Praneet Dutta and Amy Zeng and Irina Jurenka and Greg Farquhar and Yamini Bansal and Shariq Iqbal and Becca Roelofs and Ga-Young Joung and Parker Beak and Changwan Ryu and Ryan Poplin and Yan Wu and Jean-Baptiste Alayrac and Senaka Buthpitiya and Olaf Ronneberger and Caleb Habtegebriel and Wei Li and Paul Cavallaro and Aurora Wei and Guy Bensky and Timo Denk and Harish Ganapathy and Jeff Stanway and Pratik Joshi and Francesco Bertolini and Jessica Lo and Olivia Ma and Zachary Charles and Geta Sampemane and Himanshu Sahni and Xu Chen and Harry Askham and David Gaddy and Peter Young and Jiewen Tan and Matan Eyal and Arthur Bražinskas and Li Zhong and Zhichun Wu and Mark Epstein and Kai Bailey and Andrew Hard and Kamyu Lee and Sasha Goldshtein and Alex Ruiz and Mohammed Badawi and Matthias Lochbrunner and JK Kearns and Ashley Brown and Fabio Pardo and Theophane Weber and Haichuan Yang and Pan-Pan Jiang and Berkin Akin and Zhao Fu and Marcus Wainwright and Chi Zou and Meenu Gaba and Pierre-Antoine Manzagol and Wendy Kan and Yang Song and Karina Zainullina and Rui Lin and Jeongwoo Ko and Salil Deshmukh and Apoorv Jindal and James Svensson and Divya Tyam and Heri Zhao and Christine Kaeser-Chen and Scott Baird and Pooya Moradi and Jamie Hall and Qiuchen Guo and Vincent Tsang and Bowen Liang and Fernando Pereira and Suhas Ganesh and Ivan Korotkov and Jakub Adamek and Sridhar Thiagarajan and Vinh Tran and Charles Chen and Chris Tar and Sanil Jain and Ishita Dasgupta and Taylan Bilal and David Reitter and Kai Zhao and Giulia Vezzani and Yasmin Gehman and Pulkit Mehta and Lauren Beltrone and Xerxes Dotiwalla and Sergio Guadarrama and Zaheer Abbas and Stefani Karp and Petko Georgiev and Chun-Sung Ferng and Marc Brockschmidt and Liqian Peng and Christoph Hirnschall and Vikas Verma and Yingying Bi and Ying Xiao and Avigail Dabush and Kelvin Xu and Phil Wallis and Randall Parker and Qifei Wang and Yang Xu and Ilkin Safarli and Dinesh Tewari and Yin Zhang and Seungyeon Kim and Andrea Gesmundo and Mackenzie Thomas and Sergey Levi and Ahmed Chowdhury and Kanishka Rao and Peter Garst and Sam Conway-Rahman and Helen Ran and Kay McKinney and Zhisheng Xiao and Wenhao Yu and Rohan Agrawal and Axel Stjerngren and Catalin Ionescu and Jingjing Chen and Vivek Sharma and Justin Chiu and Fei Liu and Ken Franko and Clayton Sanford and Xingyu Cai and Paul Michel and Sanjay Ganapathy and Jane Labanowski and Zachary Garrett and Ben Vargas and Sean Sun and Bryan Gale and Thomas Buschmann and Guillaume Desjardins and Nimesh Ghelani and Palak Jain and Mudit Verma and Chulayuth Asawaroengchai and Julian Eisenschlos and Jitendra Harlalka and Hideto Kazawa and Don Metzler and Joshua Howland and Ying Jian and Jake Ades and Viral Shah and Tynan Gangwani and Seungji Lee and Roman Ring and Steven M. Hernandez and Dean Reich and Amer Sinha and Ashutosh Sathe and Joe Kovac and Ashleah Gill and Ajay Kannan and Andrea D'olimpio and Martin Sevenich and Jay Whang and Been Kim and Khe Chai Sim and Jilin Chen and Jiageng Zhang and Shuba Lall and Yossi Matias and Bill Jia and Abe Friesen and Sara Nasso and Ashish Thapliyal and Bryan Perozzi and Ting Yu and Anna Shekhawat and Safeen Huda and Peter Grabowski and Eric Wang and Ashwin Sreevatsa and Hilal Dib and Mehadi Hassen and Parker Schuh and Vedrana Milutinovic and Chris Welty and Michael Quinn and Ali Shah and Bangju Wang and Gabe Barth-Maron and Justin Frye and Natalie Axelsson and Tao Zhu and Yukun Ma and Irene Giannoumis and Hanie Sedghi and Chang Ye and Yi Luan and Kevin Aydin and Bilva Chandra and Vivek Sampathkumar and Ronny Huang and Victor Lavrenko and Ahmed Eleryan and Zhi Hong and Steven Hansen and Sara Mc Carthy and Bidisha Samanta and Domagoj Ćevid and Xin Wang and Fangtao Li and Michael Voznesensky and Matt Hoffman and Andreas Terzis and Vikash Sehwag and Gil Fidel and Luheng He and Mu Cai and Yanzhang He and Alex Feng and Martin Nikoltchev and Samrat Phatale and Jason Chase and Rory Lawton and Ming Zhang and Tom Ouyang and Manuel Tragut and Mehdi Hafezi Manshadi and Arjun Narayanan and Jiaming Shen and Xu Gao and Tolga Bolukbasi and Nick Roy and Xin Li and Daniel Golovin and Liviu Panait and Zhen Qin and Guangxing Han and Thomas Anthony and Sneha Kudugunta and Viorica Patraucean and Aniket Ray and Xinyun Chen and Xiaochen Yang and Tanuj Bhatia and Pranav Talluri and Alex Morris and Andrija Ražnatović and Bethanie Brownfield and James An and Sheng Peng and Patrick Kane and Ce Zheng and Nico Duduta and Joshua Kessinger and James Noraky and Siqi Liu and Keran Rong and Petar Veličković and Keith Rush and Alex Goldin and Fanny Wei and Shiva Mohan Reddy Garlapati and Caroline Pantofaru and Okwan Kwon and Jianmo Ni and Eric Noland and Julia Di Trapani and Françoise Beaufays and Abhijit Guha Roy and Yinlam Chow and Aybuke Turker and Geoffrey Cideron and Lantao Mei and Jon Clark and Qingyun Dou and Matko Bošnjak and Ralph Leith and Yuqing Du and Amir Yazdanbakhsh and Milad Nasr and Chester Kwak and Suraj Satishkumar Sheth and Alex Kaskasoli and Ankesh Anand and Balaji Lakshminarayanan and Sammy Jerome and David Bieber and Chun-Te Chu and Alexandre Senges and Tianxiao Shen and Mukund Sridhar and Ndaba Ndebele and Benjamin Beyret and Shakir Mohamed and Mia Chen and Markus Freitag and Jiaxian Guo and Luyang Liu and Paul Roit and Heng Chen and Shen Yan and Tom Stone and JD Co-Reyes and Jeremy Cole and Salvatore Scellato and Shekoofeh Azizi and Hadi Hashemi and Alicia Jin and Anand Iyer and Marcella Valentine and András György and Arun Ahuja and Daniel Hernandez Diaz and Chen-Yu Lee and Nathan Clement and Weize Kong and Drew Garmon and Ishaan Watts and Kush Bhatia and Khyatti Gupta and Matt Miecnikowski and Hugo Vallet and Ankur Taly and Edward Loper and Saket Joshi and James Atwood and Jo Chick and Mark Collier and Fotis Iliopoulos and Ryan Trostle and Beliz Gunel and Ramiro Leal-Cavazos and Arnar Mar Hrafnkelsson and Michael Guzman and Xiaoen Ju and Andy Forbes and Jesse Emond and Kushal Chauhan and Ben Caine and Li Xiao and Wenjun Zeng and Alexandre Moufarek and Daniel Murphy and Maya Meng and Nitish Gupta and Felix Riedel and Anil Das and Elijah Lawal and Shashi Narayan and Tiberiu Sosea and James Swirhun and Linda Friso and Behnam Neyshabur and Jing Lu and Sertan Girgin and Michael Wunder and Edouard Yvinec and Aroonalok Pyne and Victor Carbune and Shruti Rijhwani and Yang Guo and Tulsee Doshi and Anton Briukhov and Max Bain and Ayal Hitron and Xuanhui Wang and Ashish Gupta and Ke Chen and Cosmo Du and Weiyang Zhang and Dhruv Shah and Arjun Akula and Max Dylla and Ashyana Kachra and Weicheng Kuo and Tingting Zou and Lily Wang and Luyao Xu and Jifan Zhu and Justin Snyder and Sachit Menon and Orhan Firat and Igor Mordatch and Yuan Yuan and Natalia Ponomareva and Rory Blevins and Lawrence Moore and Weijun Wang and Phil Chen and Martin Scholz and Artur Dwornik and Jason Lin and Sicheng Li and Diego Antognini and Te I and Xiaodan Song and Matt Miller and Uday Kalra and Adam Raveret and Oscar Akerlund and Felix Wu and Andrew Nystrom and Namrata Godbole and Tianqi Liu and Hannah DeBalsi and Jewel Zhao and Buhuang Liu and Avi Caciularu and Lauren Lax and Urvashi Khandelwal and Victoria Langston and Eric Bailey and Silvio Lattanzi and Yufei Wang and Neel Kovelamudi and Sneha Mondal and Guru Guruganesh and Nan Hua and Ofir Roval and Paweł Wesołowski and Rishikesh Ingale and Jonathan Halcrow and Tim Sohn and Christof Angermueller and Bahram Raad and Eli Stickgold and Eva Lu and Alec Kosik and Jing Xie and Timothy Lillicrap and Austin Huang and Lydia Lihui Zhang and Dominik Paulus and Clement Farabet and Alex Wertheim and Bing Wang and Rishabh Joshi and Chu-ling Ko and Yonghui Wu and Shubham Agrawal and Lily Lin and XiangHai Sheng and Peter Sung and Tyler Breland-King and Christina Butterfield and Swapnil Gawde and Sumeet Singh and Qiao Zhang and Raj Apte and Shilpa Shetty and Adrian Hutter and Tao Li and Elizabeth Salesky and Federico Lebron and Jonni Kanerva and Michela Paganini and Arthur Nguyen and Rohith Vallu and Jan-Thorsten Peter and Sarmishta Velury and David Kao and Jay Hoover and Anna Bortsova and Colton Bishop and Shoshana Jakobovits and Alessandro Agostini and Alekh Agarwal and Chang Liu and Charles Kwong and Sasan Tavakkol and Ioana Bica and Alex Greve and Anirudh GP and Jake Marcus and Le Hou and Tom Duerig and Rivka Moroshko and Dave Lacey and Andy Davis and Julien Amelot and Guohui Wang and Frank Kim and Theofilos Strinopoulos and Hui Wan and Charline Le Lan and Shankar Krishnan and Haotian Tang and Peter Humphreys and Junwen Bai and Idan Heimlich Shtacher and Diego Machado and Chenxi Pang and Ken Burke and Dangyi Liu and Renga Aravamudhan and Yue Song and Ed Hirst and Abhimanyu Singh and Brendan Jou and Liang Bai and Francesco Piccinno and Chuyuan Kelly Fu and Robin Alazard and Barak Meiri and Daniel Winter and Charlie Chen and Mingda Zhang and Jens Heitkaemper and John Lambert and Jinhyuk Lee and Alexander Frömmgen and Sergey Rogulenko and Pranav Nair and Paul Niemczyk and Anton Bulyenov and Bibo Xu and Hadar Shemtov and Morteza Zadimoghaddam and Serge Toropov and Mateo Wirth and Hanjun Dai and Sreenivas Gollapudi and Daniel Zheng and Alex Kurakin and Chansoo Lee and Kalesha Bullard and Nicolas Serrano and Ivana Balazevic and Yang Li and Johan Schalkwyk and Mark Murphy and Mingyang Zhang and Kevin Sequeira and Romina Datta and Nishant Agrawal and Charles Sutton and Nithya Attaluri and Mencher Chiang and Wael Farhan and Gregory Thornton and Kate Lin and Travis Choma and Hung Nguyen and Kingshuk Dasgupta and Dirk Robinson and Iulia Comşa and Michael Riley and Arjun Pillai and Basil Mustafa and Ben Golan and Amir Zandieh and Jean-Baptiste Lespiau and Billy Porter and David Ross and Sujeevan Rajayogam and Mohit Agarwal and Subhashini Venugopalan and Bobak Shahriari and Qiqi Yan and Hao Xu and Taylor Tobin and Pavel Dubov and Hongzhi Shi and Adrià Recasens and Anton Kovsharov and Sebastian Borgeaud and Lucio Dery and Shanthal Vasanth and Elena Gribovskaya and Linhai Qiu and Mahdis Mahdieh and Wojtek Skut and Elizabeth Nielsen and CJ Zheng and Adams Yu and Carrie Grimes Bostock and Shaleen Gupta and Aaron Archer and Chris Rawles and Elinor Davies and Alexey Svyatkovskiy and Tomy Tsai and Yoni Halpern and Christian Reisswig and Bartek Wydrowski and Bo Chang and Joan Puigcerver and Mor Hazan Taege and Jian Li and Eva Schnider and Xinjian Li and Dragos Dena and Yunhan Xu and Umesh Telang and Tianze Shi and Heiga Zen and Kyle Kastner and Yeongil Ko and Neesha Subramaniam and Aviral Kumar and Pete Blois and Zhuyun Dai and John Wieting and Yifeng Lu and Yoel Zeldes and Tian Xie and Anja Hauth and Alexandru Ţifrea and Yuqi Li and Sam El-Husseini and Dan Abolafia and Howard Zhou and Wen Ding and Sahra Ghalebikesabi and Carlos Guía and Andrii Maksai and Ágoston Weisz and Sercan Arik and Nick Sukhanov and Aga Świetlik and Xuhui Jia and Luo Yu and Weiyue Wang and Mark Brand and Dawn Bloxwich and Sean Kirmani and Zhe Chen and Alec Go and Pablo Sprechmann and Nithish Kannen and Alen Carin and Paramjit Sandhu and Isabel Edkins and Leslie Nooteboom and Jai Gupta and Loren Maggiore and Javad Azizi and Yael Pritch and Pengcheng Yin and Mansi Gupta and Danny Tarlow and Duncan Smith and Desi Ivanov and Mohammad Babaeizadeh and Ankita Goel and Satish Kambala and Grace Chu and Matej Kastelic and Michelle Liu and Hagen Soltau and Austin Stone and Shivani Agrawal and Min Kim and Kedar Soparkar and Srinivas Tadepalli and Oskar Bunyan and Rachel Soh and Arvind Kannan and DY Kim and Blake JianHang Chen and Afief Halumi and Sudeshna Roy and Yulong Wang and Olcan Sercinoglu and Gena Gibson and Sijal Bhatnagar and Motoki Sano and Daniel von Dincklage and Qingchun Ren and Blagoj Mitrevski and Mirek Olšák and Jennifer She and Carl Doersch and Jilei and Wang and Bingyuan Liu and Qijun Tan and Tamar Yakar and Tris Warkentin and Alex Ramirez and Carl Lebsack and Josh Dillon and Rajiv Mathews and Tom Cobley and Zelin Wu and Zhuoyuan Chen and Jon Simon and Swaroop Nath and Tara Sainath and Alexei Bendebury and Ryan Julian and Bharath Mankalale and Daria Ćurko and Paulo Zacchello and Adam R. Brown and Kiranbir Sodhia and Heidi Howard and Sergi Caelles and Abhinav Gupta and Gareth Evans and Anna Bulanova and Lesley Katzen and Roman Goldenberg and Anton Tsitsulin and Joe Stanton and Benoit Schillings and Vitaly Kovalev and Corey Fry and Rushin Shah and Kuo Lin and Shyam Upadhyay and Cheng Li and Soroush Radpour and Marcello Maggioni and Jing Xiong and Lukas Haas and Jenny Brennan and Aishwarya Kamath and Nikolay Savinov and Arsha Nagrani and Trevor Yacovone and Ryan Kappedal and Kostas Andriopoulos and Li Lao and YaGuang Li and Grigory Rozhdestvenskiy and Kazuma Hashimoto and Andrew Audibert and Sophia Austin and Daniel Rodriguez and Anian Ruoss and Garrett Honke and Deep Karkhanis and Xi Xiong and Qing Wei and James Huang and Zhaoqi Leng and Vittal Premachandran and Stan Bileschi and Georgios Evangelopoulos and Thomas Mensink and Jay Pavagadhi and Denis Teplyashin and Paul Chang and Linting Xue and Garrett Tanzer and Sally Goldman and Kaushal Patel and Shixin Li and Jeremy Wiesner and Ivy Zheng and Ian Stewart-Binks and Jie Han and Zhi Li and Liangchen Luo and Karel Lenc and Mario Lučić and Fuzhao Xue and Ryan Mullins and Alexey Guseynov and Chung-Ching Chang and Isaac Galatzer-Levy and Adam Zhang and Garrett Bingham and Grace Hu and Ale Hartman and Yue Ma and Jordan Griffith and Alex Irpan and Carey Radebaugh and Summer Yue and Lijie Fan and Victor Ungureanu and Christina Sorokin and Hannah Teufel and Peiran Li and Rohan Anil and Dimitris Paparas and Todd Wang and Chu-Cheng Lin and Hui Peng and Megan Shum and Goran Petrovic and Demetra Brady and Richard Nguyen and Klaus Macherey and Zhihao Li and Harman Singh and Madhavi Yenugula and Mariko Iinuma and Xinyi Chen and Kavya Kopparapu and Alexey Stern and Shachi Dave and Chandu Thekkath and Florence Perot and Anurag Kumar and Fangda Li and Yang Xiao and Matthew Bilotti and Mohammad Hossein Bateni and Isaac Noble and Lisa Lee and Amelio Vázquez-Reina and Julian Salazar and Xiaomeng Yang and Boyu Wang and Ela Gruzewska and Anand Rao and Sindhu Raghuram and Zheng Xu and Eyal Ben-David and Jieru Mei and Sid Dalmia and Zhaoyi Zhang and Yuchen Liu and Gagan Bansal and Helena Pankov and Steven Schwarcz and Andrea Burns and Christine Chan and Sumit Sanghai and Ricky Liang and Ethan Liang and Antoine He and Amy Stuart and Arun Narayanan and Yukun Zhu and Christian Frank and Bahar Fatemi and Amit Sabne and Oran Lang and Indro Bhattacharya and Shane Settle and Maria Wang and Brendan McMahan and Andrea Tacchetti and Livio Baldini Soares and Majid Hadian and Serkan Cabi and Timothy Chung and Nikita Putikhin and Gang Li and Jeremy Chen and Austin Tarango and Henryk Michalewski and Mehran Kazemi and Hussain Masoom and Hila Sheftel and Rakesh Shivanna and Archita Vadali and Ramona Comanescu and Doug Reid and Joss Moore and Arvind Neelakantan and Michaël Sander and Jonathan Herzig and Aviv Rosenberg and Mostafa Dehghani and JD Choi and Michael Fink and Reid Hayes and Eric Ge and Shitao Weng and Chia-Hua Ho and John Karro and Kalpesh Krishna and Lam Nguyen Thiet and Amy Skerry-Ryan and Daniel Eppens and Marco Andreetto and Navin Sarma and Silvano Bonacina and Burcu Karagol Ayan and Megha Nawhal and Zhihao Shan and Mike Dusenberry and Shantanu Thakoor and Sagar Gubbi and Duc Dung Nguyen and Reut Tsarfaty and Samuel Albanie and Jovana Mitrović and Meet Gandhi and Bo-Juen Chen and Alessandro Epasto and Georgi Stephanov and Ye Jin and Samuel Gehman and Aida Amini and Jack Weber and Feryal Behbahani and Shawn Xu and Miltos Allamanis and Xi Chen and Myle Ott and Claire Sha and Michal Jastrzebski and Hang Qi and David Greene and Xinyi Wu and Abodunrinwa Toki and Daniel Vlasic and Jane Shapiro and Ragha Kotikalapudi and Zhe Shen and Takaaki Saeki and Sirui Xie and Albin Cassirer and Shikhar Bharadwaj and Tatsuya Kiyono and Srinadh Bhojanapalli and Elan Rosenfeld and Sam Ritter and Jieming Mao and João Gabriel Oliveira and Zoltan Egyed and Bernd Bandemer and Emilio Parisotto and Keisuke Kinoshita and Juliette Pluto and Petros Maniatis and Steve Li and Yaohui Guo and Golnaz Ghiasi and Jean Tarbouriech and Srimon Chatterjee and Julie Jin and Katrina and Xu and Jennimaria Palomaki and Séb Arnold and Madhavi Sewak and Federico Piccinini and Mohit Sharma and Ben Albrecht and Sean Purser-haskell and Ashwin Vaswani and Chongyan Chen and Matheus Wisniewski and Qin Cao and John Aslanides and Nguyet Minh Phu and Maximilian Sieb and Lauren Agubuzu and Anne Zheng and Daniel Sohn and Marco Selvi and Anders Andreassen and Krishan Subudhi and Prem Eruvbetine and Oliver Woodman and Tomas Mery and Sebastian Krause and Xiaoqi Ren and Xiao Ma and Jincheng Luo and Dawn Chen and Wei Fan and Henry Griffiths and Christian Schuler and Alice Li and Shujian Zhang and Jean-Michel Sarr and Shixin Luo and Riccardo Patana and Matthew Watson and Dani Naboulsi and Michael Collins and Sailesh Sidhwani and Emiel Hoogeboom and Sharon Silver and Emily Caveness and Xiaokai Zhao and Mikel Rodriguez and Maxine Deines and Libin Bai and Patrick Griffin and Marco Tagliasacchi and Emily Xue and Spandana Raj Babbula and Bo Pang and Nan Ding and Gloria Shen and Elijah Peake and Remi Crocker and Shubha Srinivas Raghvendra and Danny Swisher and Woohyun Han and Richa Singh and Ling Wu and Vladimir Pchelin and Tsendsuren Munkhdalai and Dana Alon and Geoff Bacon and Efren Robles and Jannis Bulian and Melvin Johnson and George Powell and Felipe Tiengo Ferreira and Yaoyiran Li and Frederik Benzing and Mihajlo Velimirović and Hubert Soyer and William Kong and Tony and Nguyên and Zhen Yang and Jeremiah Liu and Joost van Amersfoort and Daniel Gillick and Baochen Sun and Nathalie Rauschmayr and Katie Zhang and Serena Zhan and Tao Zhou and Alexey Frolov and Chengrun Yang and Denis Vnukov and Louis Rouillard and Hongji Li and Amol Mandhane and Nova Fallen and Rajesh Venkataraman and Clara Huiyi Hu and Jennifer Brennan and Jenny Lee and Jerry Chang and Martin Sundermeyer and Zhufeng Pan and Rosemary Ke and Simon Tong and Alex Fabrikant and William Bono and Jindong Gu and Ryan Foley and Yiran Mao and Manolis Delakis and Dhruva Bhaswar and Roy Frostig and Nick Li and Avital Zipori and Cath Hope and Olga Kozlova and Swaroop Mishra and Josip Djolonga and Craig Schiff and Majd Al Merey and Eleftheria Briakou and Peter Morgan and Andy Wan and Avinatan Hassidim and RJ Skerry-Ryan and Kuntal Sengupta and Mary Jasarevic and Praveen Kallakuri and Paige Kunkle and Hannah Brennan and Tom Lieber and Hassan Mansoor and Julian Walker and Bing Zhang and Annie Xie and Goran Žužić and Adaeze Chukwuka and Alex Druinsky and Donghyun Cho and Rui Yao and Ferjad Naeem and Shiraz Butt and Eunyoung Kim and Zhipeng Jia and Mandy Jordan and Adam Lelkes and Mark Kurzeja and Sophie Wang and James Zhao and Andrew Over and Abhishek Chakladar and Marcel Prasetya and Neha Jha and Sriram Ganapathy and Yale Cong and Prakash Shroff and Carl Saroufim and Sobhan Miryoosefi and Mohamed Hammad and Tajwar Nasir and Weijuan Xi and Yang Gao and Young Maeng and Ben Hora and Chin-Yi Cheng and Parisa Haghani and Yoad Lewenberg and Caden Lu and Martin Matysiak and Naina Raisinghani and Huiyu Wang and Lexi Baugher and Rahul Sukthankar and Minh Giang and John Schultz and Noah Fiedel and Minmin Chen and Cheng-Chun Lee and Tapomay Dey and Hao Zheng and Shachi Paul and Celine Smith and Andy Ly and Yicheng Wang and Rishabh Bansal and Bartek Perz and Susanna Ricco and Stasha Blank and Vaishakh Keshava and Deepak Sharma and Marvin Chow and Kunal Lad and Komal Jalan and Simon Osindero and Craig Swanson and Jacob Scott and Anastasija Ilić and Xiaowei Li and Siddhartha Reddy Jonnalagadda and Afzal Shama Soudagar and Yan Xiong and Bat-Orgil Batsaikhan and Daniel Jarrett and Naveen Kumar and Maulik Shah and Matt Lawlor and Austin Waters and Mark Graham and Rhys May and Sabela Ramos and Sandra Lefdal and Zeynep Cankara and Nacho Cano and Brendan O'Donoghue and Jed Borovik and Frederick Liu and Jordan Grimstad and Mahmoud Alnahlawi and Katerina Tsihlas and Tom Hudson and Nikolai Grigorev and Yiling Jia and Terry Huang and Tobenna Peter Igwe and Sergei Lebedev and Xiaodan Tang and Igor Krivokon and Frankie Garcia and Melissa Tan and Eric Jia and Peter Stys and Shikhar Vashishth and Yu Liang and Balaji Venkatraman and Chenjie Gu and Anastasios Kementsietsidis and Chen Zhu and Junehyuk Jung and Yunfei Bai and Mohammad Javad Hosseini and Faruk Ahmed and Aditya Gupta and Xin Yuan and Shereen Ashraf and Shitij Nigam and Gautam Vasudevan and Pranjal Awasthi and Adi Mayrav Gilady and Zelda Mariet and Ramy Eskander and Haiguang Li and Hexiang Hu and Guillermo Garrido and Philippe Schlattner and George Zhang and Rohun Saxena and Petar Dević and Kritika Muralidharan and Ashwin Murthy and Yiqian Zhou and Min Choi and Arissa Wongpanich and Zhengdong Wang and Premal Shah and Yuntao Xu and Yiling Huang and Stephen Spencer and Alice Chen and James Cohan and Junjie Wang and Jonathan Tompson and Junru Wu and Ruba Haroun and Haiqiong Li and Blanca Huergo and Fan Yang and Tongxin Yin and James Wendt and Michael Bendersky and Rahma Chaabouni and Javier Snaider and Johan Ferret and Abhishek Jindal and Tara Thompson and Andrew Xue and Will Bishop and Shubham Milind Phal and Archit Sharma and Yunhsuan Sung and Prabakar Radhakrishnan and Mo Shomrat and Reeve Ingle and Roopali Vij and Justin Gilmer and Mihai Dorin Istin and Sam Sobell and Yang Lu and Emily Nottage and Dorsa Sadigh and Jeremiah Willcock and Tingnan Zhang and Steve Xu and Sasha Brown and Katherine Lee and Gary Wang and Yun Zhu and Yi Tay and Cheolmin Kim and Audrey Gutierrez and Abhanshu Sharma and Yongqin Xian and Sungyong Seo and Claire Cui and Elena Pochernina and Cip Baetu and Krzysztof Jastrzębski and Mimi Ly and Mohamed Elhawaty and Dan Suh and Eren Sezener and Pidong Wang and Nancy Yuen and George Tucker and Jiahao Cai and Zuguang Yang and Cindy Wang and Alex Muzio and Hai Qian and Jae Yoo and Derek Lockhart and Kevin R. McKee and Mandy Guo and Malika Mehrotra and Artur Mendonça and Sanket Vaibhav Mehta and Sherry Ben and Chetan Tekur and Jiaqi Mu and Muye Zhu and Victoria Krakovna and Hongrae Lee and AJ Maschinot and Sébastien Cevey and HyunJeong Choe and Aijun Bai and Hansa Srinivasan and Derek Gasaway and Nick Young and Patrick Siegler and Dan Holtmann-Rice and Vihari Piratla and Kate Baumli and Roey Yogev and Alex Hofer and Hado van Hasselt and Svetlana Grant and Yuri Chervonyi and David Silver and Andrew Hogue and Ayushi Agarwal and Kathie Wang and Preeti Singh and Four Flynn and Josh Lipschultz and Robert David and Lizzetth Bellot and Yao-Yuan Yang and Long Le and Filippo Graziano and Kate Olszewska and Kevin Hui and Akanksha Maurya and Nikos Parotsidis and Weijie Chen and Tayo Oguntebi and Joe Kelley and Anirudh Baddepudi and Johannes Mauerer and Gregory Shaw and Alex Siegman and Lin Yang and Shravya Shetty and Subhrajit Roy and Yunting Song and Wojciech Stokowiec and Ryan Burnell and Omkar Savant and Robert Busa-Fekete and Jin Miao and Samrat Ghosh and Liam MacDermed and Phillip Lippe and Mikhail Dektiarev and Zach Behrman and Fabian Mentzer and Kelvin Nguyen and Meng Wei and Siddharth Verma and Chris Knutsen and Sudeep Dasari and Zhipeng Yan and Petr Mitrichev and Xingyu Wang and Virat Shejwalkar and Jacob Austin and Srinivas Sunkara and Navneet Potti and Yan Virin and Christian Wright and Gaël Liu and Oriana Riva and Etienne Pot and Greg Kochanski and Quoc Le and Gargi Balasubramaniam and Arka Dhar and Yuguo Liao and Adam Bloniarz and Divyansh Shukla and Elizabeth Cole and Jong Lee and Sheng Zhang and Sushant Kafle and Siddharth Vashishtha and Parsa Mahmoudieh and Grace Chen and Raphael Hoffmann and Pranesh Srinivasan and Agustin Dal Lago and Yoav Ben Shalom and Zi Wang and Michael Elabd and Anuj Sharma and Junhyuk Oh and Suraj Kothawade and Maigo Le and Marianne Monteiro and Shentao Yang and Kaiz Alarakyia and Robert Geirhos and Diana Mincu and Håvard Garnes and Hayato Kobayashi and Soroosh Mariooryad and Kacper Krasowiak and Zhixin and Lai and Shibl Mourad and Mingqiu Wang and Fan Bu and Ophir Aharoni and Guanjie Chen and Abhimanyu Goyal and Vadim Zubov and Ankur Bapna and Elahe Dabir and Nisarg Kothari and Kay Lamerigts and Nicola De Cao and Jeremy Shar and Christopher Yew and Nitish Kulkarni and Dre Mahaarachchi and Mandar Joshi and Zhenhai Zhu and Jared Lichtarge and Yichao Zhou and Hannah Muckenhirn and Vittorio Selo and Oriol Vinyals and Peter Chen and Anthony Brohan and Vaibhav Mehta and Sarah Cogan and Ruth Wang and Ty Geri and Wei-Jen Ko and Wei Chen and Fabio Viola and Keshav Shivam and Lisa Wang and Madeleine Clare Elish and Raluca Ada Popa and Sébastien Pereira and Jianqiao Liu and Raphael Koster and Donnie Kim and Gufeng Zhang and Sayna Ebrahimi and Partha Talukdar and Yanyan Zheng and Petra Poklukar and Ales Mikhalap and Dale Johnson and Anitha Vijayakumar and Mark Omernick and Matt Dibb and Ayush Dubey and Qiong Hu and Apurv Suman and Vaibhav Aggarwal and Ilya Kornakov and Fei Xia and Wing Lowe and Alexey Kolganov and Ted Xiao and Vitaly Nikolaev and Steven Hemingray and Bonnie Li and Joana Iljazi and Mikołaj Rybiński and Ballie Sandhu and Peggy Lu and Thang Luong and Rodolphe Jenatton and Vineetha Govindaraj and Hui and Li and Gabriel Dulac-Arnold and Wonpyo Park and Henry Wang and Abhinit Modi and Jean Pouget-Abadie and Kristina Greller and Rahul Gupta and Robert Berry and Prajit Ramachandran and Jinyu Xie and Liam McCafferty and Jianling Wang and Kilol Gupta and Hyeontaek Lim and Blaž Bratanič and Andy Brock and Ilia Akolzin and Jim Sproch and Dan Karliner and Duhyeon Kim and Adrian Goedeckemeyer and Noam Shazeer and Cordelia Schmid and Daniele Calandriello and Parul Bhatia and Krzysztof Choromanski and Ceslee Montgomery and Dheeru Dua and Ana Ramalho and Helen King and Yue Gao and Lynn Nguyen and David Lindner and Divya Pitta and Oleaser Johnson and Khalid Salama and Diego Ardila and Michael Han and Erin Farnese and Seth Odoom and Ziyue Wang and Xiangzhuo Ding and Norman Rink and Ray Smith and Harshal Tushar Lehri and Eden Cohen and Neera Vats and Tong He and Parthasarathy Gopavarapu and Adam Paszke and Miteyan Patel and Wouter Van Gansbeke and Lucia Loher and Luis Castro and Maria Voitovich and Tamara von Glehn and Nelson George and Simon Niklaus and Zach Eaton-Rosen and Nemanja Rakićević and Erik Jue and Sagi Perel and Carrie Zhang and Yuval Bahat and Angéline Pouget and Zhi Xing and Fantine Huot and Ashish Shenoy and Taylor Bos and Vincent Coriou and Bryan Richter and Natasha Noy and Yaqing Wang and Santiago Ontanon and Siyang Qin and Gleb Makarchuk and Demis Hassabis and Zhuowan Li and Mandar Sharma and Kumaran Venkatesan and Iurii Kemaev and Roxanne Daniel and Shiyu Huang and Saloni Shah and Octavio Ponce and Warren and Chen and Manaal Faruqui and Jialin Wu and Slavica Andačić and Szabolcs Payrits and Daniel McDuff and Tom Hume and Yuan Cao and MH Tessler and Qingze Wang and Yinan Wang and Ivor Rendulic and Eirikur Agustsson and Matthew Johnson and Tanya Lando and Andrew Howard and Sri Gayatri Sundara Padmanabhan and Mayank Daswani and Andrea Banino and Michael Kilgore and Jonathan Heek and Ziwei Ji and Alvaro Caceres and Conglong Li and Nora Kassner and Alexey Vlaskin and Zeyu Liu and Alex Grills and Yanhan Hou and Roykrong Sukkerd and Gowoon Cheon and Nishita Shetty and Larisa Markeeva and Piotr Stanczyk and Tejas Iyer and Yuan Gong and Shawn Gao and Keerthana Gopalakrishnan and Tim Blyth and Malcolm Reynolds and Avishkar Bhoopchand and Misha Bilenko and Dero Gharibian and Vicky Zayats and Aleksandra Faust and Abhinav Singh and Min Ma and Hongyang Jiao and Sudheendra Vijayanarasimhan and Lora Aroyo and Vikas Yadav and Sarah Chakera and Ashwin Kakarla and Vilobh Meshram and Karol Gregor and Gabriela Botea and Evan Senter and Dawei Jia and Geza Kovacs and Neha Sharma and Sebastien Baur and Kai Kang and Yifan He and Lin Zhuo and Marija Kostelac and Itay Laish and Songyou Peng and Louis O'Bryan and Daniel Kasenberg and Girish Ramchandra Rao and Edouard Leurent and Biao Zhang and Sage Stevens and Ana Salazar and Ye Zhang and Ivan Lobov and Jake Walker and Allen Porter and Morgan Redshaw and Han Ke and Abhishek Rao and Alex Lee and Hoi Lam and Michael Moffitt and Jaeyoun Kim and Siyuan Qiao and Terry Koo and Robert Dadashi and Xinying Song and Mukund Sundararajan and Peng Xu and Chizu Kawamoto and Yan Zhong and Clara Barbu and Apoorv Reddy and Mauro Verzetti and Leon Li and George Papamakarios and Hanna Klimczak-Plucińska and Mary Cassin and Koray Kavukcuoglu and Rigel Swavely and Alain Vaucher and Jeffrey Zhao and Ross Hemsley and Michael Tschannen and Heming Ge and Gaurav Menghani and Yang Yu and Natalie Ha and Wei He and Xiao Wu and Maggie Song and Rachel Sterneck and Stefan Zinke and Dan A. Calian and Annie Marsden and Alejandro Cruzado Ruiz and Matteo Hessel and Almog Gueta and Benjamin Lee and Brian Farris and Manish Gupta and Yunjie Li and Mohammad Saleh and Vedant Misra and Kefan Xiao and Piermaria Mendolicchio and Gavin Buttimore and Varvara Krayvanova and Nigamaa Nayakanti and Matthew Wiethoff and Yash Pande and Azalia Mirhoseini and Ni Lao and Jasmine Liu and Yiqing Hua and Angie Chen and Yury Malkov and Dmitry Kalashnikov and Shubham Gupta and Kartik Audhkhasi and Yuexiang Zhai and Sudhindra Kopalle and Prateek Jain and Eran Ofek and Clemens Meyer and Khuslen Baatarsukh and Hana Strejček and Jun Qian and James Freedman and Ricardo Figueira and Michal Sokolik and Olivier Bachem and Raymond Lin and Dia Kharrat and Chris Hidey and Pingmei Xu and Dennis Duan and Yin Li and Muge Ersoy and Richard Everett and Kevin Cen and Rebeca Santamaria-Fernandez and Amir Taubenfeld and Ian Mackinnon and Linda Deng and Polina Zablotskaia and Shashank Viswanadha and Shivanker Goel and Damion Yates and Yunxiao Deng and Peter Choy and Mingqing Chen and Abhishek Sinha and Alex Mossin and Yiming Wang and Arthur Szlam and Susan Hao and Paul Kishan Rubenstein and Metin Toksoz-Exley and Miranda Aperghis and Yin Zhong and Junwhan Ahn and Michael Isard and Olivier Lacombe and Florian Luisier and Chrysovalantis Anastasiou and Yogesh Kalley and Utsav Prabhu and Emma Dunleavy and Shaan Bijwadia and Justin Mao-Jones and Kelly Chen and Rama Pasumarthi and Emily Wood and Adil Dostmohamed and Nate Hurley and Jiri Simsa and Alicia Parrish and Mantas Pajarskas and Matt Harvey and Ondrej Skopek and Yony Kochinski and Javier Rey and Verena Rieser and Denny Zhou and Sun Jae Lee and Trilok Acharya and Guowang Li and Joe Jiang and Xiaofan Zhang and Bryant Gipson and Ethan Mahintorabi and Marco Gelmi and Nima Khajehnouri and Angel Yeh and Kayi Lee and Loic Matthey and Leslie Baker and Trang Pham and Han Fu and Alex Pak and Prakhar Gupta and Cristina Vasconcelos and Adam Sadovsky and Brian Walker and Sissie Hsiao and Patrik Zochbauer and Andreea Marzoca and Noam Velan and Junhao Zeng and Gilles Baechler and Danny Driess and Divya Jain and Yanping Huang and Lizzie Tao and John Maggs and Nir Levine and Jon Schneider and Erika Gemzer and Samuel Petit and Shan Han and Zach Fisher and Dustin Zelle and Courtney Biles and Eugene Ie and Asya Fadeeva and Casper Liu and Juliana Vicente Franco and Adrian Collister and Hao Zhang and Renshen Wang and Ruizhe Zhao and Leandro Kieliger and Kurt Shuster and Rui Zhu and Boqing Gong and Lawrence Chan and Ruoxi Sun and Sujoy Basu and Roland Zimmermann and Jamie Hayes and Abhishek Bapna and Jasper Snoek and Weel Yang and Puranjay Datta and Jad Al Abdallah and Kevin Kilgour and Lu Li and SQ Mah and Yennie Jun and Morgane Rivière and Abhijit Karmarkar and Tammo Spalink and Tao Huang and Lucas Gonzalez and Duc-Hieu Tran and Averi Nowak and John Palowitch and Martin Chadwick and Ellie Talius and Harsh Mehta and Thibault Sellam and Philipp Fränken and Massimo Nicosia and Kyle He and Aditya Kini and David Amos and Sugato Basu and Harrison Jobe and Eleni Shaw and Qiantong Xu and Colin Evans and Daisuke Ikeda and Chaochao Yan and Larry Jin and Lun Wang and Sachin Yadav and Ilia Labzovsky and Ramesh Sampath and Ada Ma and Candice Schumann and Aditya Siddhant and Rohin Shah and John Youssef and Rishabh Agarwal and Natalie Dabney and Alessio Tonioni and Moran Ambar and Jing Li and Isabelle Guyon and Benny Li and David Soergel and Boya Fang and Georgi Karadzhov and Cristian Udrescu and Trieu Trinh and Vikas Raunak and Seb Noury and Dee Guo and Sonal Gupta and Mara Finkelstein and Denis Petek and Lihao Liang and Greg Billock and Pei Sun and David Wood and Yiwen Song and Xiaobin Yu and Tatiana Matejovicova and Regev Cohen and Kalyan Andra and David D'Ambrosio and Zhiwei Deng and Vincent Nallatamby and Ebrahim Songhori and Rumen Dangovski and Andrew Lampinen and Pankil Botadra and Adam Hillier and Jiawei Cao and Nagabhushan Baddi and Adhi Kuncoro and Toshihiro Yoshino and Ankit Bhagatwala and Marcáurelio Ranzato and Rylan Schaeffer and Tianlin Liu and Shuai Ye and Obaid Sarvana and John Nham and Chenkai Kuang and Isabel Gao and Jinoo Baek and Shubham Mittal and Ayzaan Wahid and Anita Gergely and Bin Ni and Josh Feldman and Carrie Muir and Pascal Lamblin and Wolfgang Macherey and Ethan Dyer and Logan Kilpatrick and Víctor Campos and Mukul Bhutani and Stanislav Fort and Yanif Ahmad and Aliaksei Severyn and Kleopatra Chatziprimou and Oleksandr Ferludin and Mason Dimarco and Aditya Kusupati and Joe Heyward and Dan Bahir and Kevin Villela and Katie Millican and Dror Marcus and Sanaz Bahargam and Caglar Unlu and Nicholas Roth and Zichuan Wei and Siddharth Gopal and Deepanway Ghoshal and Edward Lee and Sharon Lin and Jennie Lees and Dayeong Lee and Anahita Hosseini and Connie Fan and Seth Neel and Marcus Wu and Yasemin Altun and Honglong Cai and Enrique Piqueras and Josh Woodward and Alessandro Bissacco and Salem Haykal and Mahyar Bordbar and Prasha Sundaram and Sarah Hodkinson and Daniel Toyama and George Polovets and Austin Myers and Anu Sinha and Tomer Levinboim and Kashyap Krishnakumar and Rachita Chhaparia and Tatiana Sholokhova and Nitesh Bharadwaj Gundavarapu and Ganesh Jawahar and Haroon Qureshi and Jieru Hu and Nikola Momchev and Matthew Rahtz and Renjie Wu and Aishwarya P S and Kedar Dhamdhere and Meiqi Guo and Umang Gupta and Ali Eslami and Mariano Schain and Michiel Blokzijl and David Welling and Dave Orr and Levent Bolelli and Nicolas Perez-Nieves and Mikhail Sirotenko and Aman Prasad and Arjun Kar and Borja De Balle Pigem and Tayfun Terzi and Gellért Weisz and Dipankar Ghosh and Aditi Mavalankar and Dhruv Madeka and Kaspar Daugaard and Hartwig Adam and Viraj Shah and Dana Berman and Maggie Tran and Steven Baker and Ewa Andrejczuk and Grishma Chole and Ganna Raboshchuk and Mahdi Mirzazadeh and Thais Kagohara and Shimu Wu and Christian Schallhart and Bernett Orlando and Chen Wang and Alban Rrustemi and Hao Xiong and Hao Liu and Arpi Vezer and Nolan Ramsden and Shuo-yiin Chang and Sidharth Mudgal and Yan Li and Nino Vieillard and Yedid Hoshen and Farooq Ahmad and Ambrose Slone and Amy Hua and Natan Potikha and Mirko Rossini and Jon Stritar and Sushant Prakash and Zifeng Wang and Xuanyi Dong and Alireza Nazari and Efrat Nehoran and Kaan Tekelioglu and Yinxiao Li and Kartikeya Badola and Tom Funkhouser and Yuanzhen Li and Varun Yerram and Ramya Ganeshan and Daniel Formoso and Karol Langner and Tian Shi and Huijian Li and Yumeya Yamamori and Amayika Panda and Alaa Saade and Angelo Scorza Scarpati and Chris Breaux and CJ Carey and Zongwei Zhou and Cho-Jui Hsieh and Sophie Bridgers and Alena Butryna and Nishesh Gupta and Vaibhav Tulsyan and Sanghyun Woo and Evgenii Eltyshev and Will Grathwohl and Chanel Parks and Seth Benjamin and Rina Panigrahy and Shenil Dodhia and Daniel De Freitas and Chris Sauer and Will Song and Ferran Alet and Jackson Tolins and Cosmin Paduraru and Xingyi Zhou and Brian Albert and Zizhao Zhang and Lei Shu and Mudit Bansal and Sarah Nguyen and Amir Globerson and Owen Xiao and James Manyika and Tom Hennigan and Rong Rong and Josip Matak and Anton Bakalov and Ankur Sharma and Danila Sinopalnikov and Andrew Pierson and Stephen Roller and Geoff Brown and Mingcen Gao and Toshiyuki Fukuzawa and Amin Ghafouri and Kenny Vassigh and Iain Barr and Zhicheng Wang and Anna Korsun and Rajesh Jayaram and Lijie Ren and Tim Zaman and Samira Khan and Yana Lunts and Dan Deutsch and Dave Uthus and Nitzan Katz and Masha Samsikova and Amr Khalifa and Nikhil Sethi and Jiao Sun and Luming Tang and Uri Alon and Xianghong Luo and Dian Yu and Abhishek Nayyar and Bryce Petrini and Will Truong and Vincent Hellendoorn and Nikolai Chinaev and Chris Alberti and Wei Wang and Jingcao Hu and Vahab Mirrokni and Ananth Balashankar and Avia Aharon and Aahil Mehta and Ahmet Iscen and Joseph Kready and Lucas Manning and Anhad Mohananey and Yuankai Chen and Anshuman Tripathi and Allen Wu and Igor Petrovski and Dawsen Hwang and Martin Baeuml and Shreyas Chandrakaladharan and Yuan Liu and Rey Coaguila and Maxwell Chen and Sally Ma and Pouya Tafti and Susheel Tatineni and Terry Spitz and Jiayu Ye and Paul Vicol and Mihaela Rosca and Adrià Puigdomènech and Zohar Yahav and Sanjay Ghemawat and Hanzhao Lin and Phoebe Kirk and Zaid Nabulsi and Sergey Brin and Bernd Bohnet and Ken Caluwaerts and Aditya Srikanth Veerubhotla and Dan Zheng and Zihang Dai and Petre Petrov and Yichong Xu and Ramin Mehran and Zhuo Xu and Luisa Zintgraf and Jiho Choi and Spurthi Amba Hombaiah and Romal Thoppilan and Sashank Reddi and Lukasz Lew and Li Li and Kellie Webster and KP Sawhney and Lampros Lamprou and Siamak Shakeri and Mayank Lunayach and Jianmin Chen and Sumit Bagri and Alex Salcianu and Ying Chen and Yani Donchev and Charlotte Magister and Signe Nørly and Vitor Rodrigues and Tomas Izo and Hila Noga and Joe Zou and Thomas Köppe and Wenxuan Zhou and Kenton Lee and Xiangzhu Long and Danielle Eisenbud and Anthony Chen and Connor Schenck and Chi Ming To and Peilin Zhong and Emanuel Taropa and Minh Truong and Omer Levy and Danilo Martins and Zhiyuan Zhang and Christopher Semturs and Kelvin Zhang and Alex Yakubovich and Pol Moreno and Lara McConnaughey and Di Lu and Sam Redmond and Lotte Weerts and Yonatan Bitton and Tiziana Refice and Nicolas Lacasse and Arthur Conmy and Corentin Tallec and Julian Odell and Hannah Forbes-Pollard and Arkadiusz Socala and Jonathan Hoech and Pushmeet Kohli and Alanna Walton and Rui Wang and Mikita Sazanovich and Kexin Zhu and Andrei Kapishnikov and Rich Galt and Matthew Denton and Ben Murdoch and Caitlin Sikora and Kareem Mohamed and Wei Wei and Uri First and Tim McConnell and Luis C. Cobo and James Qin and Thi Avrahami and Daniel Balle and Yu Watanabe and Annie Louis and Adam Kraft and Setareh Ariafar and Yiming Gu and Eugénie Rives and Charles Yoon and Andrei Rusu and James Cobon-Kerr and Chris Hahn and Jiaming Luo and Yuvein and Zhu and Niharika Ahuja and Rodrigo Benenson and Raphaël Lopez Kaufman and Honglin Yu and Lloyd Hightower and Junlin Zhang and Darren Ni and Lisa Anne Hendricks and Gabby Wang and Gal Yona and Lalit Jain and Pablo Barrio and Surya Bhupatiraju and Siva Velusamy and Allan Dafoe and Sebastian Riedel and Tara Thomas and Zhe Yuan and Mathias Bellaiche and Sheena Panthaplackel and Klemen Kloboves and Sarthak Jauhari and Canfer Akbulut and Todor Davchev and Evgeny Gladchenko and David Madras and Aleksandr Chuklin and Tyrone Hill and Quan Yuan and Mukundan Madhavan and Luke Leonhard and Dylan Scandinaro and Qihang Chen and Ning Niu and Arthur Douillard and Bogdan Damoc and Yasumasa Onoe and Fabian Pedregosa and Fred Bertsch and Chas Leichner and Joseph Pagadora and Jonathan Malmaud and Sameera Ponda and Andy Twigg and Oleksii Duzhyi and Jingwei Shen and Miaosen Wang and Roopal Garg and Jing Chen and Utku Evci and Jonathan Lee and Leon Liu and Koji Kojima and Masa Yamaguchi and Arunkumar Rajendran and AJ Piergiovanni and Vinodh Kumar Rajendran and Marco Fornoni and Gabriel Ibagon and Harry Ragan and Sadh MNM Khan and John Blitzer and Andrew Bunner and Guan Sun and Takahiro Kosakai and Scott Lundberg and Ndidi Elue and Kelvin Guu and SK Park and Jane Park and Arunachalam Narayanaswamy and Chengda Wu and Jayaram Mudigonda and Trevor Cohn and Hairong Mu and Ravi Kumar and Laura Graesser and Yichi Zhang and Richard Killam and Vincent Zhuang and Mai Giménez and Wael Al Jishi and Ruy Ley-Wild and Alex Zhai and Kazuki Osawa and Diego Cedillo and Jialu Liu and Mayank Upadhyay and Marcin Sieniek and Roshan Sharma and Tom Paine and Anelia Angelova and Sravanti Addepalli and Carolina Parada and Kingshuk Majumder and Avery Lamp and Sanjiv Kumar and Xiang Deng and Artiom Myaskovsky and Tea Sabolić and Jeffrey Dudek and Sarah York and Félix de Chaumont Quitry and Jiazhong Nie and Dee Cattle and Alok Gunjan and Bilal Piot and Waleed Khawaja and Seojin Bang and Simon Wang and Siavash Khodadadeh and Raghavender R and Praynaa Rawlani and Richard Powell and Kevin Lee and Johannes Griesser and GS Oh and Cesar Magalhaes and Yujia Li and Simon Tokumine and Hadas Natalie Vogel and Dennis Hsu and Arturo BC and Disha Jindal and Matan Cohen and Zi Yang and Junwei Yuan and Dario de Cesare and Tony Bruguier and Jun Xu and Monica Roy and Alon Jacovi and Dan Belov and Rahul Arya and Phoenix Meadowlark and Shlomi Cohen-Ganor and Wenting Ye and Patrick Morris-Suzuki and Praseem Banzal and Gan Song and Pranavaraj Ponnuramu and Fred Zhang and George Scrivener and Salah Zaiem and Alif Raditya Rochman and Kehang Han and Badih Ghazi and Kate Lee and Shahar Drath and Daniel Suo and Antonious Girgis and Pradeep Shenoy and Duy Nguyen and Douglas Eck and Somit Gupta and Le Yan and Joao Carreira and Anmol Gulati and Ruoxin Sang and Daniil Mirylenka and Emma Cooney and Edward Chou and Mingyang Ling and Cindy Fan and Ben Coleman and Guilherme Tubone and Ravin Kumar and Jason Baldridge and Felix Hernandez-Campos and Angeliki Lazaridou and James Besley and Itay Yona and Neslihan Bulut and Quentin Wellens and AJ Pierigiovanni and Jasmine George and Richard Green and Pu Han and Connie Tao and Geoff Clark and Chong You and Abbas Abdolmaleki and Justin Fu and Tongzhou Chen and Ashwin Chaugule and Angad Chandorkar and Altaf Rahman and Will Thompson and Penporn Koanantakool and Mike Bernico and Jie Ren and Andrey Vlasov and Sergei Vassilvitskii and Maciej Kula and Yizhong Liang and Dahun Kim and Yangsibo Huang and Chengxi Ye and Dmitry Lepikhin and Wesley Helmholz},
      year={2025},
      eprint={2507.06261},
      archivePrefix={arXiv},
      primaryClass={cs.CL},
      url={https://arxiv.org/abs/2507.06261}, 
}

@misc{Singh2026MINERVACulturalAB,
      title={{MINERVA-Cultural}: A Benchmark for Cultural and Multilingual Long Video Reasoning}, 
      author={Darshan Singh and Arsha Nagrani and Kawshik Manikantan and Harman Singh and Dinesh Tewari and Tobias Weyand and Cordelia Schmid and Anelia Angelova and Shachi Dave},
      year={2026},
      eprint={2601.10649},
      archivePrefix={arXiv},
      primaryClass={cs.CV},
      url={https://arxiv.org/abs/2601.10649}, 
}

@inproceedings{li-etal-2024-foodieqa,
    title = "{F}oodie{QA}: A Multimodal Dataset for Fine-Grained Understanding of {C}hinese Food Culture",
    author = "Li, Wenyan  and
      Zhang, Crystina  and
      Li, Jiaang  and
      Peng, Qiwei  and
      Tang, Raphael  and
      Zhou, Li  and
      Zhang, Weijia  and
      Hu, Guimin  and
      Yuan, Yifei  and
      S{\o}gaard, Anders  and
      Hershcovich, Daniel  and
      Elliott, Desmond",
    booktitle = "Proceedings of the 2024 Conference on Empirical Methods in Natural Language Processing",
    year = "2024",
    url = "https://aclanthology.org/2024.emnlp-main.1063",
    }

@misc{sealion2025,
      title={{SEA-LION}: {Southeast Asian} Languages in One Network},
      author={Raymond Ng and Thanh Ngan Nguyen and Yuli Huang and Ngee Chia Tai and Wai Yi Leong and Wei Qi Leong and Xianbin Yong and Jian Gang Ngui and Yosephine Susanto and Nicholas Cheng and Hamsawardhini Rengarajan and Peerat Limkonchotiwat and Adithya Venkatadri Hulagadri and Kok Wai Teng and Yeo Yeow Tong and Bryan Siow and Wei Yi Teo and Wayne Lau and Choon Meng Tan and Brandon Ong and Zhi Hao Ong and Jann Railey Montalan and Adwin Chan and Sajeban Antonyrex and Ren Lee and Esther Choa and David Ong Tat-Wee and Bing Jie Darius Liu and William Chandra Tjhi and Erik Cambria and Leslie Teo},
      year={2025},
      eprint={2504.05747},
      archivePrefix={arXiv},
      primaryClass={cs.CL},
      url={https://arxiv.org/abs/2504.05747},
}

@misc{wang2025internvl35,
      title={{InternVL3.5}: Advancing Open-Source Multimodal Models in Versatility, Reasoning, and Efficiency}, 
      author={Weiyun Wang and Zhangwei Gao and Lixin Gu and Hengjun Pu and Long Cui and Xingguang Wei and Zhaoyang Liu and Linglin Jing and Shenglong Ye and Jie Shao and Zhaokai Wang and Zhe Chen and Hongjie Zhang and Ganlin Yang and Haomin Wang and Qi Wei and Jinhui Yin and Wenhao Li and Erfei Cui and Guanzhou Chen and Zichen Ding and Changyao Tian and Zhenyu Wu and Jingjing Xie and Zehao Li and Bowen Yang and Yuchen Duan and Xuehui Wang and Zhi Hou and Haoran Hao and Tianyi Zhang and Songze Li and Xiangyu Zhao and Haodong Duan and Nianchen Deng and Bin Fu and Yinan He and Yi Wang and Conghui He and Botian Shi and Junjun He and Yingtong Xiong and Han Lv and Lijun Wu and Wenqi Shao and Kaipeng Zhang and Huipeng Deng and Biqing Qi and Jiaye Ge and Qipeng Guo and Wenwei Zhang and Songyang Zhang and Maosong Cao and Junyao Lin and Kexian Tang and Jianfei Gao and Haian Huang and Yuzhe Gu and Chengqi Lyu and Huanze Tang and Rui Wang and Haijun Lv and Wanli Ouyang and Limin Wang and Min Dou and Xizhou Zhu and Tong Lu and Dahua Lin and Jifeng Dai and Weijie Su and Bowen Zhou and Kai Chen and Yu Qiao and Wenhai Wang and Gen Luo},
      year={2025},
      eprint={2508.18265},
      archivePrefix={arXiv},
      primaryClass={cs.CV},
      url={https://arxiv.org/abs/2508.18265}, 
}

@inproceedings{
Guo2024TRACETG,
title={{TRACE}: Temporal Grounding Video {LLM}  via Causal Event Modeling},
author={Yongxin Guo and Jingyu Liu and Mingda Li and Qingbin Liu and Xi Chen and Xiaoying Tang},
booktitle={The Thirteenth International Conference on Learning Representations},
year={2025},
url={https://openreview.net/forum?id=14fFV0chUS}
}

@inproceedings{
zeng2025timesuite,
title={{TimeSuite}: Improving {MLLM}s for Long Video Understanding via Grounded Tuning},
author={Xiangyu Zeng and Kunchang Li and Chenting Wang and Xinhao Li and Tianxiang Jiang and Ziang Yan and Songze Li and Yansong Shi and Zhengrong Yue and Yi Wang and Yali Wang and Yu Qiao and Limin Wang},
booktitle={The Thirteenth International Conference on Learning Representations},
year={2025},
url={https://openreview.net/forum?id=nAVejJURqZ}
}

@INPROCEEDINGS {ALM-Bench_2025,
author = { Vayani, Ashmal and Dissanayake, Dinura and Watawana, Hasindri and Ahsan, Noor and Sasikumar, Nevasini and Thawakar, Omkar and Ademtew, Henok Biadglign and Hmaiti, Yahya and Kumar, Amandeep and Kuckreja, Kartik and Maslych, Mykola and Al Ghallabi, Wafa and Mihaylov, Mihail and Qin, Chao and Shaker, Abdelrahman M and Zhang, Mike and Ihsani, Mahardika Krisna and Esplana, Amiel and Gokani, Monil and Mirkin, Shachar and Singh, Harsh and Srivastava, Ashay and Hamerlik, Endre and Asma Izzati, Fathinah and Adamsyah Maani, Fadillah and Cavada, Sebastian and Chim, Jenny and Gupta, Rohit and Manjunath, Sanjay and Zhumakhanova, Kamila and Heriniaina Rabevohitra, Feno and Amirudin, Azril and Ridzuan, Muhammad and Kareem, Daniya and More, Ketan and Li, Kunyang and Shakya, Pramesh and Saad, Muhammad and Ghasemaghaei, Amirpouya and Djanibekov, Amirbek and Azizov, Dilshod and Jankovic, Branislava and Bhatia, Naman and Cabrera, Alvaro and Obando-Ceron, Johan and Otieno, Olympiah and Farestam, Fabian and Rabbani, Muztoba and Baliah, Sanoojan and Sanjeev, Santosh and Shtanchaev, Abduragim and Fatima, Maheen and Nguyen, Thao and Kareem, Amrin and Aremu, Toluwani and Xavier, Nathan and Bhatkal, Amit and Toyin, Hawau and Chadha, Aman and Cholakkal, Hisham and Anwer, Rao Muhammad and Felsberg, Michael and Laaksonen, Jorma and Solorio, Thamar and Choudhury, Monojit and Laptev, Ivan and Shah, Mubarak and Khan, Salman and Khan, Fahad Shahbaz },
booktitle = { 2025 IEEE/CVF Conference on Computer Vision and Pattern Recognition (CVPR) },
title = {{ All Languages Matter: Evaluating LMMs on Culturally Diverse 100 Languages }},
year = {2025},
volume = {},
ISSN = {},
pages = {19565-19575},
doi = {10.1109/CVPR52734.2025.01822},
url = {https://doi.ieeecomputersociety.org/10.1109/CVPR52734.2025.01822},
publisher = {IEEE Computer Society},
address = {Los Alamitos, CA, USA},
month =Jun}

@article{Christodoulou-2025,
 author = {Christodoulou, Anna-Maria and Glette, Kyrre and Lartillot, Olivier and Jensenius, Alexander Refsum},
 note = {10.5334/tismir.222},
 journal = {Transactions of the International Society for Music Information Retrieval},
 month = {Jul},
 title = {{MusiQAl}: A Dataset for Music Question–Answering through Audio–Video Fusion},
 year = {2025}
}

@misc{tonja2026afrimcq,
      title={{Afri-MCQA}: Multimodal Cultural Question Answering for {African} Languages}, 
      author={Atnafu Lambebo Tonja and Srija Anand and Emilio Villa-Cueva and Israel Abebe Azime and Jesujoba Oluwadara Alabi and Muhidin A. Mohamed and Debela Desalegn Yadeta and Negasi Haile Abadi and Abigail Oppong and Nnaemeka Casmir Obiefuna and Idris Abdulmumin and Naome A Etori and Eric Peter Wairagala and Kanda Patrick Tshinu and Imanigirimbabazi Emmanuel and Gabofetswe Malema and Alham Fikri Aji and David Ifeoluwa Adelani and Thamar Solorio},
      year={2026},
      eprint={2601.05699},
      archivePrefix={arXiv},
      primaryClass={cs.CL},
      url={https://arxiv.org/abs/2601.05699}, 
}

@inproceedings{hsieh-etal-2025-taiwanvqa,
    title = "{T}aiwan{VQA}: A Benchmark for Visual Question Answering for {T}aiwanese Daily Life",
    author = "Hsieh, Hsin-Yi  and
      Liu, Shang Wei  and
      Meng, Chang Chih  and
      Lin, Shuo-Yueh  and
      Chien-Hua, Chen  and
      Lin, Hung-Ju  and
      Huang, Hen-Hsen  and
      Wu, I-Chen",
    editor = "Zhang, Wei Emma  and
      Dai, Xiang  and
      Elliot, Desmond  and
      Fang, Byron  and
      Sim, Mongyuan  and
      Zhuang, Haojie  and
      Chen, Weitong",
    booktitle = "Proceedings of the First Workshop of Evaluation of Multi-Modal Generation",
    month = jan,
    year = "2025",
    address = "Abu Dhabi, UAE",
    publisher = "Association for Computational Linguistics",
    url = "https://aclanthology.org/2025.evalmg-1.6/",
    pages = "57--75",
}

@misc{Pranav2025RiceVLEV,
      title={{Rice-VL}: Evaluating Vision-Language Models for Cultural Understanding Across {ASEAN} Countries}, 
      author={Tushar Pranav and Eshan Pandey and Austria Lyka Diane Bala and Aman Chadha and Indriyati Atmosukarto and Donny Soh Cheng Lock},
      year={2025},
      eprint={2512.01419},
      archivePrefix={arXiv},
      primaryClass={cs.CV},
      url={https://arxiv.org/abs/2512.01419}, 
}

@misc{zheng2025mmaasia,
      title={{MMA-ASIA}: A Multilingual and Multimodal Alignment Framework for Culturally-Grounded Evaluation}, 
      author={Weihua Zheng and Zhengyuan Liu and Tanmoy Chakraborty and Weiwen Xu and Xiaoxue Gao and Bryan Chen Zhengyu Tan and Bowei Zou and Chang Liu and Yujia Hu and Xing Xie and Xiaoyuan Yi and Jing Yao and Chaojun Wang and Long Li and Rui Liu and Huiyao Liu and Koji Inoue and Ryuichi Sumida and Tatsuya Kawahara and Fan Xu and Lingyu Ye and Wei Tian and Dongjun Kim and Jimin Jung and Jaehyung Seo and Nadya Yuki Wangsajaya and Pham Minh Duc and Ojasva Saxena and Palash Nandi and Xiyan Tao and Wiwik Karlina and Tuan Luong and Keertana Arun Vasan and Roy Ka-Wei Lee and Nancy F. Chen},
      year={2025},
      eprint={2510.08608},
      archivePrefix={arXiv},
      primaryClass={cs.CL},
      url={https://arxiv.org/abs/2510.08608}, 
}

@inproceedings{faraz2026indicvisionbench,
 author = {Faraz, Ali and Akash and Khan, Shaharukh and Kolla, Raja and Patidar, Akshat and Goswami, Suranjan and Ravi, Abhinav and Khatri, Chandra and Agarwal, Shubham},
 booktitle = {International Conference on Learning Representations},
 editor = {C. Vondrick and B. Hariharan and C. Raffel and L. Pinto and D. Yang and A. Faust},
 pages = {147604--147646},
 title = {{IndicVisionBench}: Benchmarking Cultural and Multilingual Understanding in VLMs},
 url = {https://proceedings.iclr.cc/paper_files/paper/2026/file/eecdbf1beb9a006f69811b0663b45552-Paper-Conference.pdf},
 volume = {2026},
 year = {2026}
}

@InProceedings{Fahim_2026_WACV,
    author    = {Fahim, Md and Rahman, Md Sakib Ul and Rahman, Akm Moshiur and Ishmam, Md Farhan and Rahman, Md Tasmim and Shifat, Fariha Tanjim and Haider, Fabiha and Alam Bhuiyan, Md Farhad},
    title     = {{BanglaProtha}: Evaluating Vision Language Models in Underrepresented Long-tail Cultural Contexts},
    booktitle = {Proceedings of the IEEE/CVF Winter Conference on Applications of Computer Vision (WACV)},
    month     = {March},
    year      = {2026},
    pages     = {1159-1169}
}

@misc{jiang2026avmeme,
      title={{AVMeme Exam}: A Multimodal Multilingual Multicultural Benchmark for {LLMs'} Contextual and Cultural Knowledge and Thinking}, 
      author={Xilin Jiang and Qiaolin Wang and Junkai Wu and Xiaomin He and Zhongweiyang Xu and Yinghao Ma and Minshuo Piao and Kaiyi Yang and Xiuwen Zheng and Riki Shimizu and Yicong Chen and Arsalan Firoozi and Gavin Mischler and Sukru Samet Dindar and Richard Antonello and Linyang He and Tsun-An Hsieh and Xulin Fan and Yulun Wu and Yuesheng Ma and Chaitanya Amballa and Weixiong Chen and Jiarui Hai and Ruisi Li and Vishal Choudhari and Cong Han and Yinghao Aaron Li and Adeen Flinker and Mounya Elhilali and Emmanouil Benetos and Mark Hasegawa-Johnson and Romit Roy Choudhury and Nima Mesgarani},
      year={2026},
      eprint={2601.17645},
      archivePrefix={arXiv},
      primaryClass={cs.SD},
      url={https://arxiv.org/abs/2601.17645}, 
}

@misc{li2026maven,
      title={{MAVEN} A Multi-Agent Framework for Multicultural Text-to-Video Generation}, 
      author={Shuowei Li and Yuming Zhao and Parth Bhalerao and Oana Ignat},
      year={2026},
      eprint={2605.16716},
      archivePrefix={arXiv},
      primaryClass={cs.CV},
      url={https://arxiv.org/abs/2605.16716}, 
}

@inproceedings{luo2025museg,
    title = "{MUSEG}: Reinforcing Video Temporal Understanding via Timestamp-Aware Multi-Segment Grounding",
    author = "Luo, Fuwen  and
      Lou, Shengfeng  and
      Chen, Chi  and
      Wang, Ziyue  and
      Li, Chenliang  and
      Shen, Weizhou  and
      Guo, Jiyue  and
      Li, Peng  and
      Yan, Ming  and
      Zhang, Ji  and
      Huang, Fei  and
      Liu, Yang",
    editor = "Liakata, Maria  and
      Moreira, Viviane P.  and
      Zhang, Jiajun  and
      Jurgens, David",
    booktitle = "Proceedings of the 64th Annual Meeting of the {A}ssociation for {C}omputational {L}inguistics (Volume 1: Long Papers)",
    month = jul,
    year = "2026",
    address = "San Diego, California, United States",
    publisher = "Association for Computational Linguistics",
    url = "https://aclanthology.org/2026.acl-long.1644/",
    doi = "10.18653/v1/2026.acl-long.1644",
    pages = "35549--35561",
    ISBN = "979-8-89176-390-6"
}

@INPROCEEDINGS{satar2021semantic,
  author={Satar, Burak and Hongyuan, Zhu and Bresson, Xavier and Lim, Joo Hwee},
  booktitle={2021 IEEE International Conference on Image Processing (ICIP)}, 
  title={Semantic Role Aware Correlation Transformer For Text To Video Retrieval}, 
  year={2021},
  volume={},
  number={},
  pages={1334-1338},
  doi={10.1109/ICIP42928.2021.9506267}}

@inproceedings{satar2023debiasing,
  author       = {Burak Satar and
                  Hongyuan Zhu and
                  Hanwang Zhang and
                  Joo{-}Hwee Lim},
  title        = {Towards Debiasing Frame Length Bias in Text-Video Retrieval via Causal
                  Intervention},
  booktitle    = {34th British Machine Vision Conference 2023, {BMVC} 2023, Aberdeen,
                  UK, November 20-24, 2023},
  pages        = {650--658},
  publisher    = {{BMVA} Press},
  year         = {2023},
  url          = {https://papers.bmvc2023.org/0650.pdf},
  bibsource    = {dblp computer science bibliography, https://dblp.org}
}

@misc{ilaslan2025vgtvp,
      title={{VG-TVP}: Multimodal Procedural Planning via Visually Grounded Text-Video Prompting}, 
      author={Muhammet Furkan Ilaslan and Ali Koksal and Kevin Qinhong Lin and Burak Satar and Mike Zheng Shou and Qianli Xu},
      year={2024},
      eprint={2412.11621},
      archivePrefix={arXiv},
      primaryClass={cs.CV},
      url={https://arxiv.org/abs/2412.11621}, 
}

@misc{satar2022exploiting,
      title={Exploiting Semantic Role Contextualized Video Features for Multi-Instance Text-Video Retrieval {EPIC-KITCHENS-100} Multi-Instance Retrieval Challenge 2022}, 
      author={Burak Satar and Hongyuan Zhu and Hanwang Zhang and Joo Hwee Lim},
      year={2022},
      eprint={2206.14381},
      archivePrefix={arXiv},
      primaryClass={cs.CV},
      url={https://arxiv.org/abs/2206.14381}, 
}

@inproceedings{ponwitayarat2026seabed,
    title = "{SEA}-{BED}: How Do Embedding Models Represent {S}outheast {A}sian Languages?",
    author = "Ponwitayarat, Wuttikorn  and
      Limkonchotiwat, Peerat  and
      Ng, Raymond  and
      Montalan, Jann Railey  and
      Aung, Thura  and
      Ngui, Jian Gang  and
      Susanto, Yosephine  and
      Tjhi, William Chandra  and
      Tasawong, Panuthep  and
      Cambria, Erik  and
      Chuangsuwanich, Ekapol  and
      Nutanong, Sarana",
    editor = "Liakata, Maria  and
      Moreira, Viviane P.  and
      Zhang, Jiajun  and
      Jurgens, David",
    booktitle = "Proceedings of the 64th Annual Meeting of the {A}ssociation for {C}omputational {L}inguistics (Volume 1: Long Papers)",
    month = jul,
    year = "2026",
    address = "San Diego, California, United States",
    publisher = "Association for Computational Linguistics",
    url = "https://aclanthology.org/2026.acl-long.397/",
    doi = "10.18653/v1/2026.acl-long.397",
    pages = "8788--8822",
    ISBN = "979-8-89176-390-6"
}

@misc{han2026culturevidbench,
      title={{CultureVidBench}: Benchmarking Cultural Understanding in Text-to-Video Generation}, 
      author={Xianjing Han and Yuhan Su and Yang Deng and Dong Ma and Wee Peng Tay and Bin Zhu},
      year={2026},
      eprint={2608.01942},
      archivePrefix={arXiv},
      primaryClass={cs.CV},
      url={https://arxiv.org/abs/2608.01942}, 
}

@misc{lee2026juryduty,
      title={Jury Duty: Calibration and Orientation Failures in {MLLM}-as-a-Judge Under Cultural Ambiguity}, 
      author={Daniel Lee and Harsh Sharma and Eunkyu Park and Pranav Narayanan Venkit and Jeonghwan Kim and Kah Mun Chia and Andreas Vlachos and Shafiq Joty},
      year={2026},
      eprint={2606.20676},
      archivePrefix={arXiv},
      primaryClass={cs.CV},
      url={https://arxiv.org/abs/2606.20676}, 
}

@misc{irawan2025confused,
      title={Vision Language Models are Confused Tourists}, 
      author={Patrick Amadeus Irawan and Ikhlasul Akmal Hanif and Muhammad Dehan Al Kautsar and Genta Indra Winata and Fajri Koto and Alham Fikri Aji},
      year={2025},
      eprint={2511.17004},
      archivePrefix={arXiv},
      primaryClass={cs.CV},
      url={https://arxiv.org/abs/2511.17004}, 
}

@misc{wang2026valueground,
      title={{ValueGround}: Evaluating Culture-Conditioned Visual Value Grounding in {MLLMs}}, 
      author={Zhipin Wang and Christoph Leiter and Christian Frey and Mohamed Hesham Ibrahim Abdalla and Josif Grabocka and Steffen Eger},
      year={2026},
      eprint={2604.06484},
      archivePrefix={arXiv},
      primaryClass={cs.CL},
      url={https://arxiv.org/abs/2604.06484}, 
}

@misc{cahyawijaya2026anthropogenic,
      title={Anthropogenic Regional Adaptation in Multimodal Vision-Language Model}, 
      author={Samuel Cahyawijaya and Peerat Limkonchotiwat and Tack Hwa Wong and Hitesh Laxmichand Patel and Amit Agarwal and Manuel Antonio Rufino and Carlos Rafael Catalan and Muhammad Reza Qorib and Vicky Feliren and Holy Lovenia and Aye Hninn Khine and Frederikus Hudi and David Anugraha and Alham Fikri Aji and Romrawin Chumpu and Viet-Thanh Pham and Minghan Wang and Mohamed Fazli Imam and Ruochen Zhang and Joseph Marvin Imperial and Khumaisa Nur'aini and Do Xuan Long and Musa Izzanardi Wijanarko and Joel Ruben Antony Moniz and Patrick Amadeus Irawan and Hanif Muhammad Zhafran and Isaiah Flores and Salsabila Zahirah Pranida and Jun Kevin and Jostin Jerico Rosal and Patricia Nicole Monderin and Kun Kerdthaisong and Ahmad Mustafid and My Chiffon Nguyen and Natchapon Jongwiriyanurak and Siva Worajitwannakul and Haochen Li and Adrian Xuan Wei Lim and Bin Wang and Muhammad Ravi Shulthan Habibi and Lynnette Hui Xian Ng and Mithil Bangera and Yeshil Bangera and Priyaranjan Pattnayak and Dun Li Chan and Sherissa Caren Djuniwar and Cho Chan Myei Oo and Hee Ming Shan},
      year={2026},
      eprint={2604.11490},
      archivePrefix={arXiv},
      primaryClass={cs.AI},
      url={https://arxiv.org/abs/2604.11490}, 
}

@inproceedings{tasawong2026seaguard,
    title = "{SEA-Guard}: Culturally Grounded Multilingual Safeguard for {S}outheast {A}sia",
    author = "Tasawong, Panuthep  and
      Ngui, Jian Gang  and
      Aji, Alham Fikri  and
      Cohn, Trevor  and
      Limkonchotiwat, Peerat",
    editor = "Liakata, Maria  and
      Moreira, Viviane P.  and
      Zhang, Jiajun  and
      Jurgens, David",
    booktitle = "Findings of the {A}ssociation for {C}omputational {L}inguistics: {ACL} 2026",
    month = jul,
    year = "2026",
    address = "San Diego, California, United States",
    publisher = "Association for Computational Linguistics",
    url = "https://aclanthology.org/2026.findings-acl.141/",
    doi = "10.18653/v1/2026.findings-acl.141",
    pages = "2917--2941",
    ISBN = "979-8-89176-395-1"
}

@misc{kim2025worldinaframe,
      title={World in a Frame: Understanding Culture Mixing as a New Challenge for Vision-Language Models}, 
      author={Eunsu Kim and Junyeong Park and Na Min An and Junseong Kim and Hitesh Laxmichand Patel and Jiho Jin and Julia Kruk and Amit Agarwal and Srikant Panda and Fenal Ashokbhai Ilasariya and Hyunjung Shim and Alice Oh},
      year={2025},
      eprint={2511.22787},
      archivePrefix={arXiv},
      primaryClass={cs.CV},
      url={https://arxiv.org/abs/2511.22787}, 
}

@inproceedings{tan2025blendvis,
    title = "{BLEnD-Vis}: Benchmarking Multimodal Cultural Understanding in Vision Language Models",
    author = "Tan, Bryan Chen Zhengyu  and
      Zheng, Weihua  and
      Liu, Zhengyuan  and
      Chen, Nancy F.  and
      Lee, Hwaran  and
      Choo, Kenny Tsu Wei  and
      Lee, Roy Ka-Wei",
    editor = "Demberg, Vera  and
      Inui, Kentaro  and
      Marquez, Llu{\'i}s",
    booktitle = "Proceedings of the 19th Conference of the {E}uropean Chapter of the {A}ssociation for {C}omputational {L}inguistics (Volume 1: Long Papers)",
    month = mar,
    year = "2026",
    address = "Rabat, Morocco",
    publisher = "Association for Computational Linguistics",
    url = "https://aclanthology.org/2026.eacl-long.215/",
    doi = "10.18653/v1/2026.eacl-long.215",
    pages = "4647--4669",
    ISBN = "979-8-89176-380-7"
}

@inproceedings{koksal2025benchmarking,
author = {Ali Koksal and Loke Mei Hwan and Hui Li Tan and Nancy F. Chen},
title = {Benchmarking Visual Generative Models through Cultural Lens: A Case Study with {S}ingapore-Centric Multi-Cultural Context},
year = {2025},
isbn = {9798400720765},
publisher = {Association for Computing Machinery},
address = {New York, NY, USA},
url = {https://doi.org/10.1145/3747327.3764896},
doi = {10.1145/3747327.3764896},
booktitle = {Companion Proceedings of the 27th International Conference on Multimodal Interaction},
pages = {190–198},
numpages = {9},
location = {
},
series = {ICMI Companion '25}
}

@inproceedings{liu2024tempcompass,
    title = "{T}emp{C}ompass: Do Video {LLM}s Really Understand Videos?",
    author = "Liu, Yuanxin  and
      Li, Shicheng  and
      Liu, Yi  and
      Wang, Yuxiang  and
      Ren, Shuhuai  and
      Li, Lei  and
      Chen, Sishuo  and
      Sun, Xu  and
      Hou, Lu",
    editor = "Ku, Lun-Wei  and
      Martins, Andre  and
      Srikumar, Vivek",
    booktitle = "Findings of the Association for Computational Linguistics: ACL 2024",
    month = aug,
    year = "2024",
    address = "Bangkok, Thailand",
    publisher = "Association for Computational Linguistics",
    url = "https://aclanthology.org/2024.findings-acl.517/",
    doi = "10.18653/v1/2024.findings-acl.517",
    pages = "8731--8772"
}

\appendix

\section{Benchmark Construction}
\label{sec:app_construction}

\subsection{S1/S2 distractor sampling algorithm}
\label{sec:distractor-sampling}

Each S1 and S2 question has four options: one correct concept plus
three distractors. The same three distractors are reused across S1
(text) and S2 (video moment), so the difficulty profile is identical
at both stages. Distractors are drawn from a per-category candidate
pool using a two-phase constrained greedy algorithm
as shown in Algorithm~\ref{alg:distractor}.

\paragraph{Inputs and constraints.}

For each category $c$ with $N_c$ concepts (Celebration: 91, Dance:
60, Game: 76, Music: 68, Wedding: 11), the algorithm consumes (i) a
Sentence-BERT cosine-similarity matrix over the symbolic concept
descriptions, (ii) a concept-to-country mapping, and (iii) a
manually curated set of \emph{ambiguity groups} $\mathcal{A}$: sets
of concept names that refer to the same underlying ritual, dance
form, game, or instrument under different country naming
conventions. Across CMB we curated 17 ambiguity groups in total:
\begin{itemize}
\item \textbf{Celebration} (4): \{Visak Bochea, Waisak, Wesak Day\}
  (Vesak); \{Lebaran, Hari Raya Aidilfitri\} (Eid al-Fitr);
  \{Makha Bucha, Meak Bochea\}; \{Idul Adha, Hari Raya Haji\}
  (Eid al-Adha).
\item \textbf{Dance} (2): \{Tinikling, Mua Sap\} (bamboo-pole dance);
  \{Khon, Yama Zatdaw\} (Ramayana dance drama).
\item \textbf{Game} (8): \{Congklak, Congkak\} (mancala);
  \{Engklek, Piko\} (hopscotch); \{Lompat Tali, Nhay Day\}
  (jump rope); \{Gasing, Danh Cu, Mark Khang\} (spinning top);
  \{Wau, Tha Dieu\} (kite flying); \{Guli, Kelereng\} (marbles);
  \{Capteh, Da Cau, Sipa\} (shuttlecock);
  \{Sepak Raga, Chinlone\} (rattan ball).
\item \textbf{Music} (3): \{Kulintang (ID), Kulintang (MY),
  Kulintang (PH)\}; \{Rebab (ID), Rebab (MY)\};
  \{Suling (ID), Suling (MY)\}.
\item \textbf{Wedding} (0): no ambiguity groups required.
\end{itemize}
Two concepts in the same group are never placed in the same MCQ.

Each key concept has three distractor slots. Slot~0 holds a
same-country distractor (forcing fine-grained intra-country
discrimination); slots 1 and 2 hold distractors from two
\emph{different} other countries, so no two distractors of the same
item come from the same country. Each concept may be used as a
distractor at most $N_{\max} = \lfloor 0.1 \cdot N_c \rfloor$ times
across the category, capping per-concept re-use at $\sim$10\%. When a category's pool cannot fill every slot under this cap (Wedding, with 11 concepts, is the only case), the cap is relaxed until every slot fills; in Wedding the most-used concept appears four times. When a country holds a single concept in a category (several countries do for Wedding), the same-country slot is filled with the most similar different-country candidate instead.

\paragraph{Algorithm.}

The assignment is computed in two phases.
\textbf{Phase 1 (coverage guarantee):} concepts are sorted by
\emph{placability} (the number of valid open slots they fit
into), and each unused concept is placed in the highest-similarity
key with a compatible open slot, guaranteeing every concept appears
as a distractor at least once. \textbf{Phase 2 (similarity-first
greedy fill):} for each remaining open slot, the most-constrained
key is processed first, and the slot is filled with the
highest-similarity candidate that satisfies the validity predicate
(self, repeat, $N_{\max}$, ambiguity, country). The fill terminates
when all slots are populated or no further valid placement exists.

\begin{algorithm*}[t]
\small
\caption{Per-category distractor sampling (BSWA). \\$\bar{s}$ denotes the other foreign slot ($\bar{s}{=}2$ if $s{=}1$, else $1$); $\mathrm{ctry}(\bot)$ matches no country.}
\label{alg:distractor}
\begin{algorithmic}[1]
\Require Concepts $C$ ($|C|{=}N$); similarity $\mathrm{sim}(\cdot,\cdot)$;
country map $\mathrm{ctry}(\cdot)$; ambiguity groups $\mathcal{A}$
\Ensure Distractor triple $\mathrm{opt}[k] = (s_0, s_1, s_2)$ for each $k \in C$
\State $N_{\max} \gets \lfloor 0.1\cdot N \rfloor$
\State $\mathrm{opt}[k] \gets (\bot, \bot, \bot),\ \mathrm{usage}[c] \gets 0$ for all $k,c \in C$
\Statex
\State \textbf{Phase 1 (coverage guarantee).}
\State Sort $C$ ascending by placability $p(c) = |\{(k,s) : \textsc{Valid}(k, s, c)\}|$
\ForAll{$c \in C$ in sorted order \textbf{with} $\mathrm{usage}[c] = 0$}
  \State $(k^{*}, s^{*}) \gets \arg\max_{k,s} \mathrm{sim}(k, c)$
    \quad s.t.\ $\textsc{Valid}(k, s, c)$
  \State $\mathrm{opt}[k^{*}][s^{*}] \gets c$;\ $\mathrm{usage}[c] \gets \mathrm{usage}[c] + 1$
\EndFor
\Statex
\State \textbf{Phase 2 (similarity-first greedy fill).}
\While{any slot of any key is empty}
  \State Sort keys with empty slots ascending by remaining valid-candidate capacity
  \ForAll{such key $k$ (most-constrained first)}
    \ForAll{empty slot $s$ of $k$}
      \State $c^{*} \gets \arg\max_{c} \mathrm{sim}(k, c)$ \quad s.t.\ $\textsc{Valid}(k, s, c)$
      \If{$c^{*}$ exists}
        \State $\mathrm{opt}[k][s] \gets c^{*}$;\ $\mathrm{usage}[c^{*}] \gets \mathrm{usage}[c^{*}] + 1$;\ \textbf{break}
      \EndIf
    \EndFor
  \EndFor
\EndWhile
\Statex
\Function{Valid}{$k, s, c$}
  \If{$c = k$ \textbf{or} $c \in \mathrm{opt}[k]$ \textbf{or} $\mathrm{usage}[c] \ge N_{\max}$}
    \State \Return \textsc{false}
  \EndIf
  \If{$\exists\, a \in \{k\} \cup \mathrm{opt}[k] \setminus \{\bot\}$ s.t.\ $\{a, c\} \subseteq G$ for some $G \in \mathcal{A}$}
    \State \Return \textsc{false} \Comment{ambiguity block}
  \EndIf
  \If{$s = 0$} \State \Return $\mathrm{ctry}(c) = \mathrm{ctry}(k)$
  \Else \ \ \State \Return $\mathrm{ctry}(c) \neq \mathrm{ctry}(k)$ \textbf{and} $\mathrm{ctry}(c) \neq \mathrm{ctry}(\mathrm{opt}[k][\bar s])$
  \EndIf
\EndFunction
\end{algorithmic}
\end{algorithm*}

\paragraph{Validation.}

Across all five categories the algorithm produced no unfilled slots, no unused concepts, and, beyond the Wedding exceptions documented above, no over-cap usage, no country-constraint violations, and no ambiguity violations. The median Sentence-BERT cosine similarity between a correct concept and its three sampled distractors is 0.70, versus 0.60 for non-selected candidates in the same per-category pool, confirming the chosen distractors are systematically harder than a random baseline by construction. Semantic-role structure has similarly been used to align text and video \citep{satar2022exploiting}.

\subsection{Released record schema}
\label{sec:record-schema}

Each released CMB record corresponds to one of the 306 concepts and
carries the following fields:

\begin{description}
  \item[\texttt{concept\_id}] Stable identifier, format
    \texttt{<country>-<category>-<NN>}.
  \item[\texttt{country}] ISO-3166 alpha-2 code (one of KH, ID, MM,
    MY, PH, TH, VN).
  \item[\texttt{category}] One of \{Celebration, Dance, Game, Music,
    Wedding\}.
  \item[\texttt{concept\_name}] Concept name in native script with a
    Latin transliteration where applicable.
  \item[\texttt{description}] English-language symbolic description
    used as the stem of the S1 and S2 questions.
  \item[\texttt{s1\_options}] List of four candidate name strings.
  \item[\texttt{s1\_correct\_index}] Integer in $\{0,1,2,3\}$.
  \item[\texttt{s1\_distractor\_origin}] Per-distractor flag in
    \{same-country, different-country\}.
  \item[\texttt{video\_A\_url}, \texttt{video\_A\_s2\_moment}]
    YouTube URL and \texttt{(start\_s, end\_s)} timestamps for the
    S2 source clip.
  \item[\texttt{s2\_options}] Four \texttt{(video\_url, start\_s,
    end\_s)} triples, one correct and three semantic-similarity
    distractors.
  \item[\texttt{s2\_correct\_index}] Integer in $\{0,1,2,3\}$.
  \item[\texttt{video\_B\_url}, \texttt{video\_B\_s3\_span}]
    YouTube URL and \texttt{(start\_s, end\_s)} ground-truth span
    for S3, with $\texttt{video\_B} \neq \texttt{video\_A}$.
  \item[\texttt{s3\_cue}] Textual description of the target
    sub-event used as the S3 prompt.
  \item[\texttt{writing\_cluster}] Derived flag in \{Latin, Local\}.
  \item[\texttt{audio\_role}] Derived label in \{Complementary, Redundant, Distracting\} computed from the mean $\Delta$\,mIoU of the three modality-instrumented models (Gemini~3.1~Pro, Qwen3-VL-32B, Qwen3.5-27B); see \S\ref{sec:modality}.
  \item[\texttt{annotator\_role\_id}] Anonymised Cultural-Annotator
    identifier (one of seven).
  \item[\texttt{review\_outcome}] One of \{passed, rewritten,
    discarded-and-replaced\} from the Outsider Filter.
\end{description}

\subsection{Annotator recruitment and screening}
\label{sec:recruitment}

\paragraph{Procedural rules.} 

Every rater completes their Non-local batch before their Local batch to avoid own-country priming. Question text is identical to the model prompt. Raters cannot skip items or revisit previous answers after submission. No external aids: raters answered each S1 item from memory and regional priors only; web search, LLM assistants, and consultation with peers were prohibited. The rating interface required raters to confirm this rule at the start of each session.

\paragraph{Local Annotators.} 

Recruited via the authors' academic contacts at universities in each of the seven countries in Southeast Asia. Inclusion criteria: i) born and raised in the country, ii) native fluency in the country's primary language, iii) general cultural exposure documented through a short intake survey on festivals, family traditions, and regional dialects, iv) willingness to reach a unanimous consensus with one or two peers from the same country. Two Local Annotators per country, three for the Philippines and Myanmar, to cover their wider linguistic diversity.

\paragraph{Cultural Annotators.} 

A Cultural Annotator is a Local Annotator who additionally meets one of the following criteria: i) authored or co-authored a publication in cultural studies, anthropology, ethnomusicology, or adjacent fields covering the country's traditions; ii) sustained co-curricular involvement (cultural-society leadership, festival organization, or museum work) verified by two reference letters. Two Cultural Annotators are co-authors (Vietnam, Malaysia); the remaining five were recruited through the same academic networks and screened on a three-concept calibration task before formal annotation began.

\paragraph{Reviewers.} 

Non-local researchers, all co-authors, were selected based on documented cultural exposure to Southeast Asia, either through prior peer-reviewed publications on SEA culture or at least 6 months of fieldwork in the region. Reviewers are explicitly not permitted to be a Local Annotator for the country they are reviewing.

\paragraph{14-rater human-study pool.} 

Recruited via the authors' extended network and a public call distributed through partner cultural-studies departments. Each rater was screened on i) native fluency, ii) self-reported expert vs non-expert status (verified by follow-up), and iii) confirmation that they had not previously seen CMB material.

\subsection{Conceptual Knowledge vs.\ Procedural Traditions}
\label{sec:conceptual-procedural}

CMB stages probe two complementary kinds of cultural knowledge. \textbf{Conceptual Knowledge} (S1, S2) is knowing \textit{what a concept symbolizes}: the meaning behind the name and its visual signature. \textbf{Procedural Traditions} (S3) is knowing \textit{how the tradition unfolds in time}: which sub-events occur, in what order, and how they bind a recognizable moment. Procedural knowledge in video is likewise central to multimodal procedural planning \citep{ilaslan2025vgtvp}. We use this distinction to motivate the textual-only LLM audits: R1's S1 pass guards Conceptual Knowledge items against text-level shortcuts, and its S3 pass guards Procedural Traditions items against under-specified sub-event descriptions. Visual judgment for both kinds remains exclusively human. Reported calibration and orientation failures of MLLM judges under cultural ambiguity support this choice \citep{lee2026juryduty}.


\begin{table*}[!htb]
\centering
\small
\resizebox{0.9\textwidth}{!}{
\begin{tabular}{llp{11cm}}
\toprule
\textbf{Country} & \textbf{Category} & \textbf{Concepts} \\
\midrule
Cambodia & Celebration (5) & Bon Chrat Preah Neangkoal, Choul Chnam Thmey, Meak Bochea, Pchum Ben, Visak Bochea \\
 & Dance (7) & Robam Bes Kravanh, Robam Chuon Por, Robam Kngaok, Robam Phlet, Robam Phloy Suoy, Robam Tep Apsara, Robam Yike \\
 & Game (6) & Bos Angkunh, Chol Chhoung, Krob Moin, Leak Kanseng, Ok Chaktrong, Sey \\
 & Music (9) & Chhing, Khaen, Khloy, Kong Thom, Kong Touch, Roneat Ek, Skor Thom, Sneng, Sralai \\
 & Wedding (1) & Kar Khmer \\
\addlinespace
Indonesia & Celebration (14) & Bulan Puasa, Festival Gamelan Yogyakarta, Festival Rawa Pening, Festival Tabot, Galungan, Idul Adha, Kuningan, Lebaran, Nyepi, Pasola, Perang Topat, Pesta Kesenian Bali, Sekaten, Waisak \\
 & Dance (8) & Tari Barong, Tari Cendrawasih, Tari Gambyong, Tari Kecak, Tari Legong, Tari Merak, Tari Reog Ponorogo, Tari Saman \\
 & Game (8) & Bola Gebok, Congklak, Egrang, Engklek, Galah Asin, Gasing, Kelereng, Lompat Tali \\
 & Music (13) & Angklung, Gambang, Gong, Kecapi, Kempul, Keroncong, Kolintang, Kulintang, Rebab, Sape', Suling, Talempong, Tanjidor \\
 & Wedding (5) & Pernikahan Adat Bali, Pernikahan Adat Batak, Pernikahan Adat Betawi, Pernikahan Adat Jawa, Pernikahan Adat Sunda \\
\addlinespace
Malaysia & Celebration (16) & Bon Odori, Chinese New Year, Deepavali, Hari Gawai, Hari Merdeka, Hari Raya Aidilfitri, Hari Raya Haji, Hungry Ghost Festival, King's Birthday, Mid-Autumn Festival, Pesta Kaamatan, Ponggal, Qing Ming, Thaipusam, Thimithi, Wesak Day \\
 & Dance (11) & Mak Yong, Tarian Asli, Tarian Bhangra, Tarian Inang, Tarian Joget, Tarian Kuda Kepang, Tarian Ngajat, Tarian Silat, Tarian Sumazau, Tarian Ulek Mayang, Tarian Zapin \\
 & Game (10) & Baling Selipar, Batu Seremban, Capteh, Congkak, Galah Panjang, Guli, Lari Dalam Guni, Sepak Raga, Silat, Wau \\
 & Music (6) & Gambus, Kompang, Kulintang, Rebab, Sape, Suling \\
 & Wedding (1) & Perkahwinan Melayu \\
\addlinespace
Myanmar & Celebration (13) & Hle Pwe, Kason Festival, Naga Hnit Thit Kuu Pwe Daw, Nat Pwe, Sanda Muni Pwe, Sein Taw Ya Paya Pwe, Shwedagon Pagoda Pwe, Taungbyon Nat Pwe, Tazaungdaing Mee Pone Pyan Pwe, Thadingyut Pwe, Thei Pone Cedi Pwe, Thingyan Festival, Tipitaka Chanting Ceremony \\
 & Dance (10) & Kinnari Kinnara A Ka, Kyaukse Cin A Ka, Shan Daung A Ka, Si Mi Gwet A Ka, Ta Bin Daing A Ka, Thangyat A Ka, Yama Zatdaw A Ka, Yein A Ka, Zat Pwe A Ka, Zawgyi A Ka \\
 & Game (7) & Chaw Taing Tet, Chinlone, Hlei Pyaing Pwe, Htoke Si Toe, Lethwei, Lun Swe Pwe, Paik Kyaw Chin \\
 & Music (6) & Hne, Maung, Palwei, Pattala, Saung, Si Neh Wa \\
 & Wedding (1) & Mingala Saung \\
\addlinespace
Philippines & Celebration (25) & Ati-Atihan, Banga Festival, Buwan ng Wika, Carabao Festival, Dinagyang Festival, Kaamulan, Kadayawan, Kalilangan, Kasanggayahan, Kawayan Festival, MassKara, Obando Fertility Rites, Pahiyas Festival, Panagat Festival, Panagbenga Festival, Parada ng Lechon, Pintados-Kasadyaan, Pista ng San Juan, Sanduguan, Santacruzan, Sillag Festival, Sinakulo, Sinulog Festival, Suman Festival, Ylang-Ylang Festival \\
 & Dance (7) & Binasuan, La Jota Moncadena, Pandanggo sa Ilaw, Sagayan, Singkil, Subli, Tinikling \\
 & Game (12) & Agawang Buko, Bati-Cobra, Dama, Jolen, Langit-Lupa, Luksong Baka, Luksong Tinik, Patintero, Piko, Pukpok Palayok, Sipa, Tumbang Preso \\
 & Music (7) & Banduria, Gabbang, Kudyapi, Kulintang, Luntang, Tamburin, Tumpong \\
 & Wedding (1) & Kasal \\
\addlinespace
Thailand & Celebration (4) & Constitution Day, End of Buddhist Lent, Makha Bucha, Songkran \\
 & Dance (9) & Fon, Khon, Lakhon Nai, Manorah, Rabam Chao Na, Ram Krabi Krabong, Ram Muay, Ram Si Nuan, Ram Thoet Thoeng \\
 & Game (17) & Chon Kho Khon, Dueng Nang, Ka Dot Chueak, Ka Fak Kai, Ka Toeng Ka Toi, Kaeng Kwien, Kaeng Ruea Bok, Kee Ma Song Muang, Kee Tu Klang Na, Ling Ching Lak, Mark Kep, Mark Khang, Morn Son Pa, Tee Gai, Tee Jub, Ti Luk Lo, Wing Piao \\
 & Music (11) & Angkalung, Jakay, Khlui, Kong Chai, Kong Wong Lek, Pee, Pin Pia, Ranat, Sor Duang, Ta-pone, Tone \\
 & Wedding (1) & Ngarn Taeng Ngarn \\
\addlinespace
Vietnam & Celebration (14) & Gio To Hung Vuong, Hoi Dua Bo Bay Nui, Hoi Dua Voi Buon Don, Hoi Lim, Le Hoi Dan Toc Co Tu, Le Hoi Den Long Hoi An, Le Hoi Long Tong, Le Hoi Thap Ba Ponagar, Le Hoi Tien Cong, Le Phat Dan, Le Tich Dien, Tet Doan Ngo, Tet Nguyen Dan, Tet Trung Thu \\
 & Dance (8) & Mua Cong Chieng, Mua Gay Senh Tien, Mua Khen, Mua Lan, Mua Non, Mua Phach Senh Tien, Mua Quat, Mua Sap \\
 & Game (16) & Banh Chuyen Banh Dua, Bau Cua, Bit Mat Bat De, Bit Mat Dap Nieu, Da Cau, Danh Cu, Di Ca Kheo, Keo Co, Nem Con, Nhay Bao Bo, Nhay Day, Nhay Lo Co, O An Quan, Oan Tu Xi, Rong Ran Len May, Tha Dieu \\
 & Music (16) & Cong Chieng, Dan Bau, Dan Da, Dan Day, Dan Nhi, Dan T'rung, Dan Tranh, Dan Ty Ba, Ken La, Khen, Phach Ca Tru, Phach Hat Van, Sao, Senh Tien, Song Loan, Trong \\
 & Wedding (1) & Dam Cuoi \\
\bottomrule
\end{tabular}
}
\caption{The 306 CMB cultural concepts (Common Latin script), grouped by country and category. Counts in parentheses. Official Latin and native-script (Local) variants are released in the JSON dataset; see Appendix~A.11.}\label{tab:concept-inventory}
\end{table*}


\subsection{Dataset Analysis}
\label{sec:dataset-analysis}

This subsection summarizes CMB's concept and question distributions, video and span durations, and the difficulty of the semantic-similarity distractors. Table~\ref{tab:concept-inventory} lists the full concept inventory by country and category. All statistics are derived from the released CMB dataset: the per-concept analyses use $n{=}306$ items, and the S3 analyses use $n{=}631$ pairs across 318 unique source videos (some Set~B source videos contribute more than one sub-event span).

\paragraph{Concept and question distribution.}

\begin{figure*}[t]
\centering
\includegraphics[width=\linewidth]{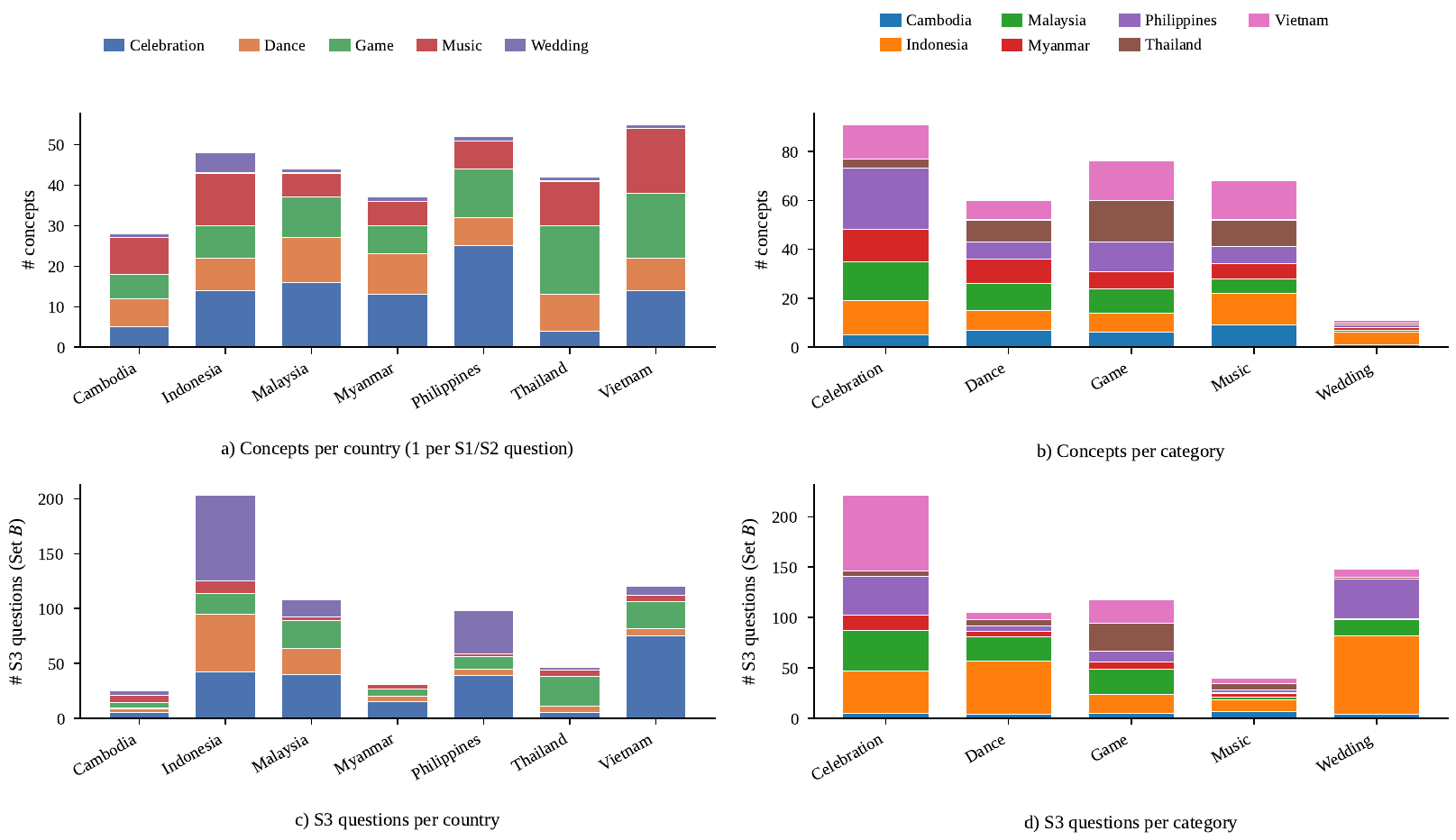}
\caption{CMB concept and S3 question distribution. (a, b) Concepts per country and per category ($n{=}306$). Each concept corresponds to one S1 and one S2 question on Set~$A$. (c, d) S3 questions per country and per category on Set~$B$ ($n{=}631$ across 318 source videos).}
\label{fig:dist}
\end{figure*}

Figure~\ref{fig:dist} shows the per-country and per-category breakdowns. Each concept yields one S1 and one S2 question on Set~$A$, so subplots (a) and (b) describe both the concept inventory and the Set~$A$ question pool. Subplots (c) and (d) show the Set~$B$ S3 question pool, which is larger because some source videos host more than one sub-event span. The distribution is dominated by Celebration (91) and Music (68); Wedding (11) is the smallest category, reflecting the relatively narrow space of distinct wedding traditions per country (typically one canonical ceremony name).

\paragraph{Question length and video duration.}

\begin{figure*}[t]
\centering
\includegraphics[width=0.8\linewidth]{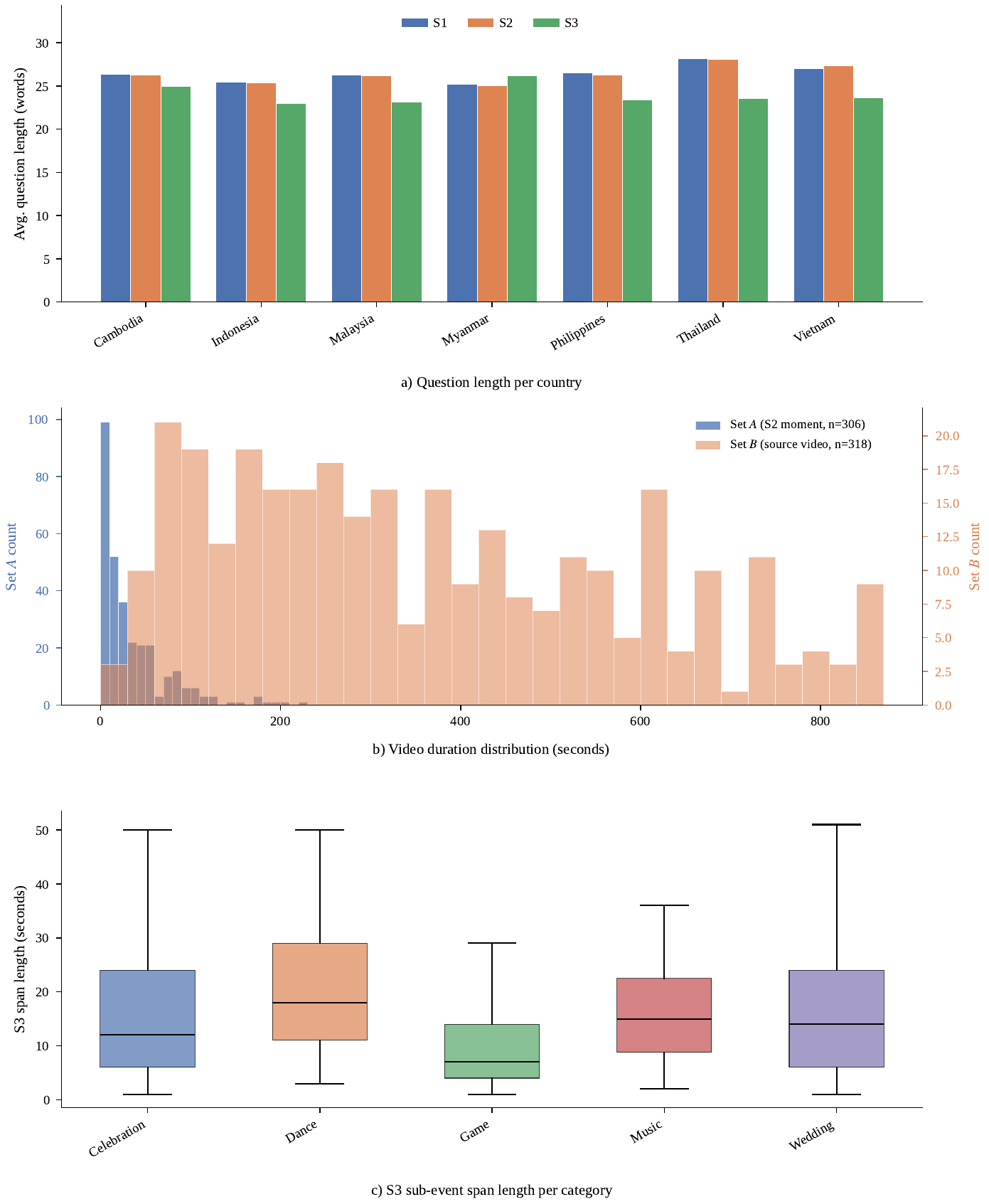}
\caption{Question length and video duration distributions. (a) Average S1, S2, and S3 question length (words) per country. (b) Video-duration histogram: Set~$A$ S2 moments are short (median 20\,s), Set~$B$ S3 source videos are long (median 316\,s). (c) S3 sub-event span lengths per category, median 13\,s overall.}
\label{fig:lengths-durations}
\end{figure*}

Figure~\ref{fig:lengths-durations} reports word-count statistics for the S1, S2, and S3 questions, video-duration histograms for Set~$A$ (S2 moments) and Set~$B$ (S3 source videos), and a per-category box plot of S3 sub-event span lengths. Questions are tightly clustered in length (S1: $26.4$ words on average, S2: $26.4$, S3: $23.5$), reflecting the constraint that descriptions stay focused on a single symbolic concept. Set~$A$ moments are short (median $20$\,s) by design, while Set~$B$ source videos are substantially longer (median $316$\,s; mean $371$\,s, matching the $\sim$6.2~min figure), giving S3 a non-trivial localization span to search. Sub-event spans are short across all categories (median $13$\,s), with Game spans being the shortest because game rounds are typically self-contained and ritualized.

\paragraph{Distractor similarity.}

\begin{figure*}[!htb]
\centering
\includegraphics[width=0.9\linewidth]{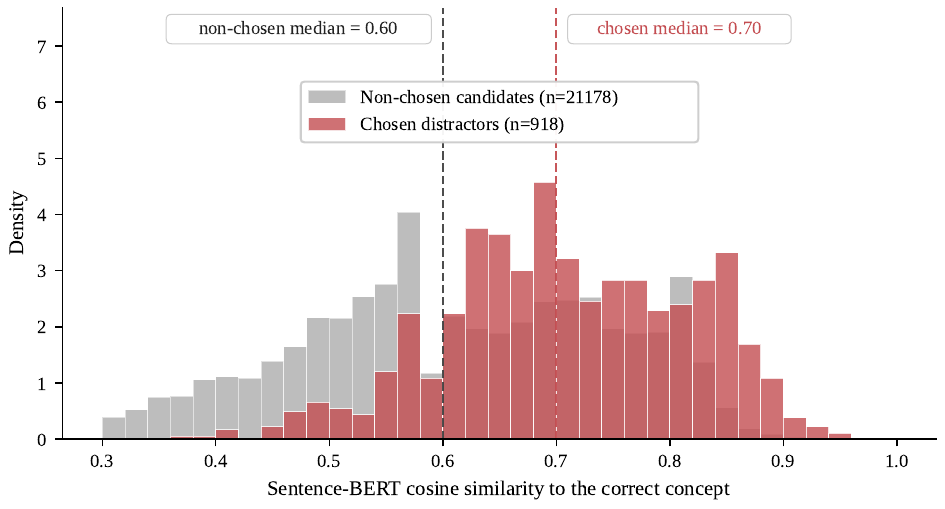}
\caption{Sentence-BERT cosine similarity between the correct concept and (red) the three chosen S1/S2 distractors versus (gray) the remaining non-chosen candidates in the same per-category pool. Chosen distractors are systematically harder, with a median similarity of $0.70$ vs $0.60$ for non-chosen candidates.}
\label{fig:distractor-sim}
\end{figure*}

Figure~\ref{fig:distractor-sim} confirms that the three S1/S2 distractors selected per concept are systematically harder than a random sample from the same candidate pool. The selected distractors have a Sentence-BERT cosine similarity to the correct concept of median $0.70$ (vs $0.60$ for the non-selected candidate pool), so the model cannot fall back on coarse keyword matching to discriminate the four options.

\paragraph{Summary statistics.}

Table~\ref{tab:dataset-stats} consolidates the headline numbers used elsewhere in the paper.

\begin{table*}[!htb]
\centering
\small
\resizebox{0.75\textwidth}{!}{%
\begin{tabular}{lr}
\toprule
\textbf{Statistic} & \textbf{Value} \\
\midrule
Concepts                                 & 306 \\
S1 questions ($=$ concepts)              & 306 \\
S2 questions ($=$ concepts)              & 306 \\
S3 questions (Set~$B$, with repeats)     & 631 \\
Unique Set~$A$ S2-moment videos          & 306 \\
Unique Set~$B$ S3-source videos          & 318 \\
\midrule
S2 moment duration (Set~$A$, $n{=}306$): median (IQR), seconds  & 20 (5--48) \\
S3 source duration (Set~$B$, $n{=}318$): median (IQR), seconds  & 316 (173--540) \\
S3 sub-event span ($n{=}631$ pairs): median (IQR), seconds      & 13 (6--24) \\
\midrule
S1 question length: mean words             & 26.4 \\
S2 question length: mean words             & 26.4 \\
S3 question length: mean words             & 23.5 \\
\midrule
Chosen distractor similarity: median       & 0.70 \\
Non-chosen candidate similarity: median    & 0.60 \\
\bottomrule
\end{tabular}
}
\caption{CMB dataset summary statistics. Set A counts are 1-to-1 with concepts (one S2-moment video per concept); Set B counts are larger because multi-event traditions contribute more than one S3 sub-event span.}
\label{tab:dataset-stats}
\end{table*}

\section{Human Evaluation}
\label{sec:app_humaneval}

\subsection{Human Evaluation Design}
\label{sec:human_eval}

To quantify the cultural gap between current video-language models and SEA native speakers, we conduct a human evaluation on Stages 1 and 2 under two complementary roles: a within-country \textbf{Native} ceiling and a within-region but out-of-country \textbf{Non-native} floor. For every assigned concept, each rater answers the S1 item and its matched S2 item, which keeps the same description as its stem and replaces the name options with the four candidate video moments. Raters see exactly the same question text and options as the models did, so the human and model numbers are directly comparable on the same concepts.

\paragraph{Human Rater pool.}

We recruit \textbf{14 qualified raters}, two per SEA country (Cambodia, Indonesia, Malaysia, Myanmar, Philippines, Thailand, Vietnam). Every rater serves in two roles: they are one of two \textit{Local} raters for their own country, and one of two \textit{Non-local} raters for exactly one other country. Consequently, each of the seven countries is evaluated by four distinct raters: two Native raters and two Non-native raters.

\paragraph{Non-local rotation.}

Non-local assignments follow a fixed $+1/+2$ rotation. Ordering the countries $\mathrm{KH}\!=\!0$, $\mathrm{ID}\!=\!1$, $\mathrm{MY}\!=\!2$, $\mathrm{MM}\!=\!3$, $\mathrm{PH}\!=\!4$, $\mathrm{TH}\!=\!5$, $\mathrm{VN}\!=\!6$, person $a$ of country $i$ rates country $(i\!+\!1)\bmod 7$ as Non-local, and person $b$ of country $i$ rates country $(i\!+\!2)\bmod 7$. This guarantees that each country receives Non-local ratings from two \textit{different} home countries; the resulting pairings are shown in Table~\ref{tab:human_rotation}.

\paragraph{Split-and-overlap rating structure.}

Within each country's 306-concept universe, both Local raters split the concepts: 20\% of the concepts form an \textbf{overlap subset} rated by \textit{both} Locals, while the remaining 80\% is partitioned equally, with each Local rating 40\% solo. The two Non-local raters for that country follow the same split on the same 20\% overlap subset. Each non-overlapping concept therefore receives two ratings (one Native and one Non-native); each overlapping concept receives four ratings (two Native and two Non-native).

\paragraph{Overlap-subset selection.}

The 20\% overlap subset is drawn by stratified random sampling within each country--category cell using $\max(1,\operatorname{round}(0.2\,n_{\mathrm{cat}}))$, with a fixed random seed. The selection is frozen before annotation begins and released as part of the benchmark for reproducibility; raters are not told which items are overlap versus solo.

\paragraph{Aggregation.}

For each concept, the \textit{Native score} is the mean of available Local ratings, and the \textit{Non-native score} is the mean of available Non-local ratings (binary $\{0,1\}$ on solo items; continuous $\{0, 0.5, 1\}$ on overlap items). Per-country Native and Non-native floors are the mean of per-concept scores over that country's concepts. Model numbers reported alongside use the same per-country concept subsets, so all three quantities (Native / Non-native / Model) are computed on the same items.

\paragraph{Workload.}

Per-rater load ranges from 42 to 64 S1 MCQs (approximately 45\,min per rater), producing approximately 732 ratings total ($\sim$9.3 person-hours aggregate). Each assigned concept also carried its matched S2 item, answered in the same session; the workload figures above count S1 MCQs, and the S2 pass added approximately the same number of judgments plus video-viewing time. Table~\ref{tab:human_workload} gives the per-kit breakdown.

\paragraph{Acknowledged structural asymmetry.}

Under any design with two Non-local raters per country, the Latin-script vs.\ non-Latin-script country imbalance (4 vs.\ 3) forces exactly one country to deviate from the ``1 Latin~+~1 Local Non-local'' pattern. We use the \emph{+1/+2 rotation}, as shown in Table~\ref{tab:human_rotation}, in which each country's two Local raters are also paired with the next country (+1) and the country after that (+2) in a cyclic order.
Under this rotation, the deviation lands on \textbf{Myanmar}, whose two Non-locals are both from the Latin-script cluster (MY and ID). Because Myanmar is itself a non-Latin-script country, both are still cross-cluster raters from its perspective; six of seven countries receive mixed (Latin + Local) Non-local perspectives, and Myanmar's Outsider floor is computed entirely from Latin-script raters. Script-cluster sub-analyses (same-cluster vs.\ different-cluster Non-native floor) are available on the six mixed countries.

\begin{table*}[!htb]
  \centering
  \small
  \setlength{\tabcolsep}{4pt}
  \renewcommand{\arraystretch}{1.08}
  \resizebox{0.75\textwidth}{!}{%
  \begin{tabular}{l l l l l}
    \toprule
    Country (target) & Cluster & Local raters & Non-local raters & Non-local clusters \\
    \midrule
    Cambodia    & Local & KH-a, KH-b & VN-a, TH-b & Latin + Local \\
    Indonesia   & Latin & ID-a, ID-b & KH-a, VN-b & Local + Latin \\
    Malaysia    & Latin & MY-a, MY-b & ID-a, KH-b & Latin + Local \\
    Myanmar     & Local & MM-a, MM-b & MY-a, ID-b & \textbf{Latin + Latin} \\
    Philippines & Latin & PH-a, PH-b & MM-a, MY-b & Local + Latin \\
    Thailand    & Local & TH-a, TH-b & PH-a, MM-b & Latin + Local \\
    Vietnam     & Latin & VN-a, VN-b & TH-a, PH-b & Local + Latin \\
    \bottomrule
  \end{tabular}
  }
  \caption{Annotator assignment under the $+1/+2$ rotation. Each of the seven countries is evaluated by two Local raters (own-country) and two Non-local raters drawn from two different other countries. Six of seven countries receive Non-local ratings from both script clusters (Latin-script: Indonesia(ID)/Malaysia(MY)/Philippines(PH)/Vietnam(VN); non-Latin-script: Cambodia(KH)/Myanmar(MM)/Thailand(TH)); Myanmar is the structural exception discussed in the text.}
  \label{tab:human_rotation}
\end{table*}

\begin{table}[!htb]
  \centering
  \small
  \setlength{\tabcolsep}{3.5pt}
  \renewcommand{\arraystretch}{1.05}
  \resizebox{1\columnwidth}{!}{%
  \begin{tabular}{l l r l r r}
    \toprule
    \# & Annotator & \shortstack{Local\\(60\% of own)} & Non-local country & \shortstack{Non-local\\(60\%)} & Total \\
    \midrule
    1  & KH-a & 17 (of 28) & Indonesia   & 29 & 46 \\
    2  & KH-b & 17         & Malaysia    & 26 & 43 \\
    3  & ID-a & 29 (of 48) & Malaysia    & 26 & 55 \\
    4  & ID-b & 29         & Myanmar     & 22 & 51 \\
    5  & MY-a & 26 (of 44) & Myanmar     & 22 & 48 \\
    6  & MY-b & 26         & Philippines & 31 & 57 \\
    7  & MM-a & 22 (of 37) & Philippines & 31 & 53 \\
    8  & MM-b & 22         & Thailand    & 25 & 47 \\
    9  & PH-a & 31 (of 52) & Thailand    & 25 & 56 \\
    10 & PH-b & 31         & Vietnam     & 33 & 64 \\
    11 & TH-a & 25 (of 42) & Vietnam     & 33 & 58 \\
    12 & TH-b & 25         & Cambodia    & 17 & 42 \\
    13 & VN-a & 33 (of 55) & Cambodia    & 17 & 50 \\
    14 & VN-b & 33         & Indonesia   & 29 & 62 \\
    \bottomrule
  \end{tabular}
  }
  \caption{Per-annotator S1 workload under the split-and-overlap
  scheme. Each annotator rates 60\% of their own country's concepts
  (40\% solo + 20\% overlap) plus 60\% of one Non-local country under
  the same split. Per-annotator range: 42--64 MCQs
  ($\sim$32--48\,min); total ratings collected: $\sim$732.}
  \label{tab:human_workload}
\end{table}

\subsection{Human Raters}

Expert raters comprise 2 Ph.D. holders, 3 Ph.D. candidates, 1 B.A. graduate, and 1 B.A. student; 5 have published cultural-research papers, and 2 have documented hands-on cultural fieldwork experience. Non-expert raters are native speakers without cultural-research publications or formal cultural-studies training; they were screened only on native fluency and confirmation that they had not previously seen CMB material.

\subsection{Per-country inter-rater agreement}
\label{app:irr_per_country}

Table~\ref{tab:irr_per_country} reports 4-way Fleiss' $\kappa$ on S1 and S2 for each of the seven countries' overlap items ($n=6$--$12$ per country). Per-country values are noisy at this sample size but follow the pattern observed in the pooled analysis: S2 $\kappa$ values are positive across all seven countries, while S1 $\kappa$ values fluctuate around zero. This is consistent with S2's moment-grounded design reducing ambiguity relative to S1's abstract symbolic prompts. The near-chance pooled S1 $\kappa$ is consistent with, rather than evidence against, the design: cultural concepts of a neighboring country are hard even for trained native raters, which is exactly the gap the Outsider Filter (\S3.3) is intended to harden.

\begin{table}[!htb]
\centering
\small
\resizebox{1\columnwidth}{!}{
\begin{tabular}{lrrrr}
\toprule
Country & $S_1$ $\kappa$ & $S_1$ n & $S_2$ $\kappa$ & $S_2$ n \\
\midrule
Cambodia    & $+0.005$ &  6 & $+0.120$ &  6 \\
Indonesia   & $+0.079$ & 11 & $+0.175$ & 11 \\
Malaysia    & $+0.056$ &  9 & $+0.193$ &  9 \\
Myanmar     & $-0.172$ &  8 & $+0.099$ &  8 \\
Philippines & $-0.202$ & 10 & $+0.225$ & 10 \\
Thailand    & $-0.052$ &  9 & $+0.152$ &  9 \\
Vietnam     & $+0.106$ & 12 & $+0.024$ & 12 \\
\midrule
\shortstack[l]{Pooled\\(Fleiss, all countries)} & $+0.011$ & 65 & $+0.166$ & 65 \\
\bottomrule
\end{tabular}
}
\caption{Per-country 4-way Fleiss' $\kappa$ on S1 and S2 across the 20\% overlap items. Pooled row uses all 65 overlap items together. S2 $\kappa$ is positive across all seven countries; S1 $\kappa$ fluctuates around zero, consistent with S2's visual grounding reducing rater disagreement relative to S1's abstract symbolic prompts.}
\label{tab:irr_per_country}
\end{table}

\subsection{Human-evaluation results by rater tier}
\label{app:human_results}

\begin{figure*}[!htb]
    \centering
    \includegraphics[width=0.8\linewidth]{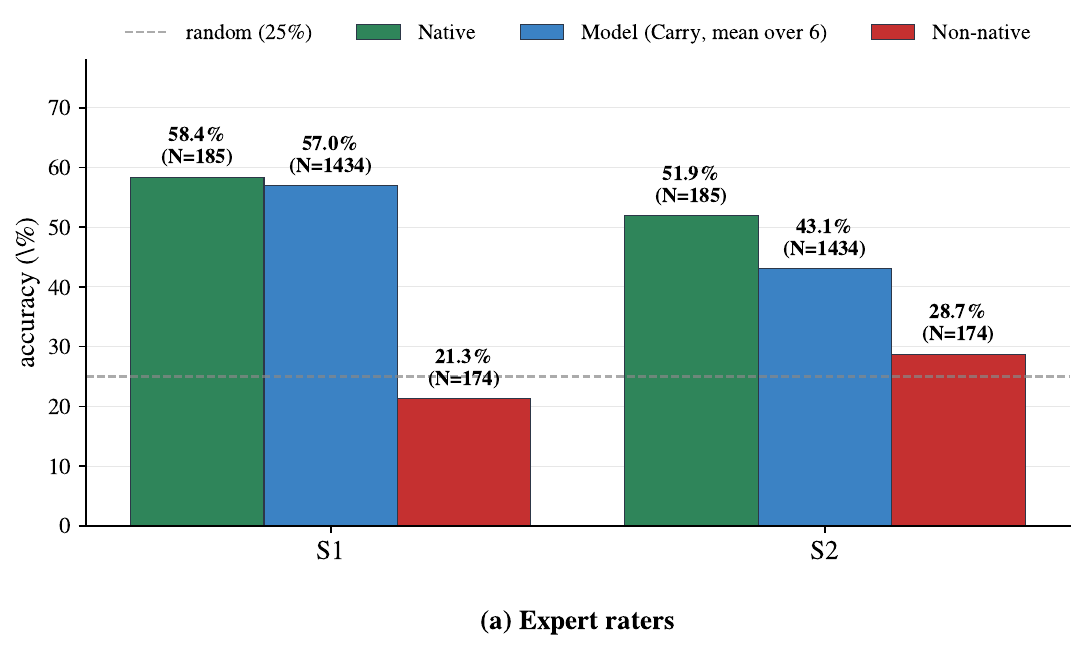}
    \includegraphics[width=0.8\linewidth]{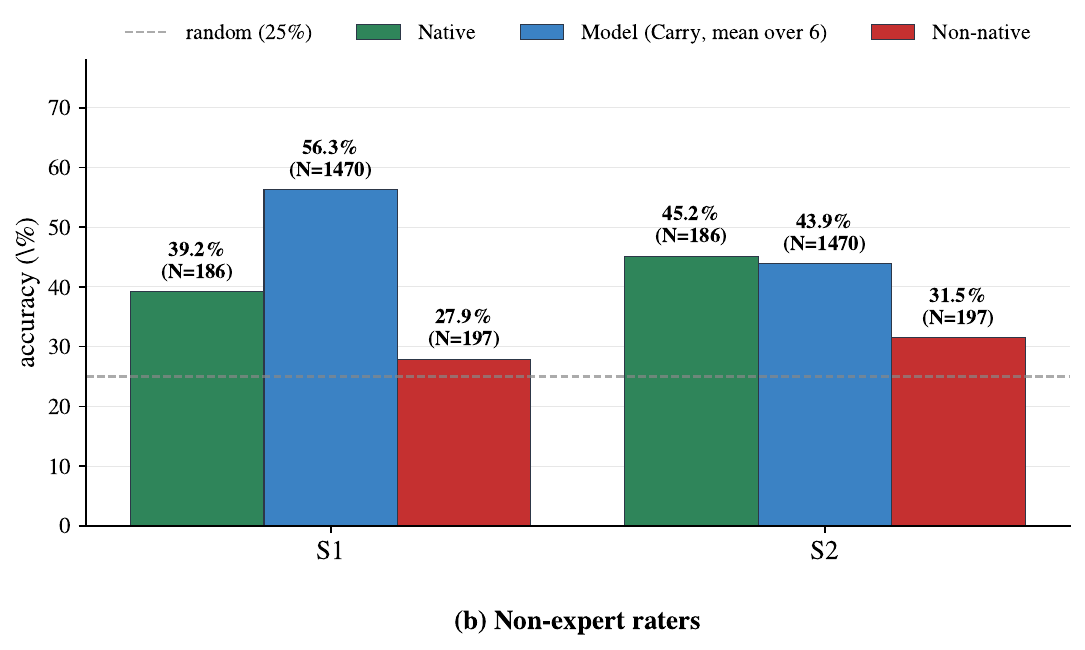}
    \caption{Human evaluation accuracy on a stratified 20\% CMB subset. \textbf{(a) Expert raters} = 7 kits (ID-a, KH-b, MM-a, MY-b, PH-a, TH-b, VN-a) with prior cultural-research experience. \textbf{(b) Non-expert raters} = 7 kits (ID-b, KH-a, MM-b, MY-a, PH-b, TH-a, VN-b) from the general population. ``Native'' = answering one's own country; ``Non-native'' = answering one other country via a fixed $+1/+2$ rotation. The Model bar averages over the six Carry-mode models on the same concept set the humans saw (N = unique concepts $\times$ 6 models).}
    \label{fig:app_human_eval}
\end{figure*}

Figure~\ref{fig:app_human_eval} breaks the human study down by rater tier and by whether the rater answered their own country or a neighboring one, with the six-model Carry-mode mean on the same concepts as a reference line. Table~\ref{tab:human_eval} in Section~\ref{sec:modality} summarizes the same data.

\section{Implementation}
\label{sec:app_impl}

\subsection{More on Implementation Details}

\paragraph{Joint score: zero-imputation and worked example.}

In the Joint formula $\text{Joint} = \frac{1}{N}\sum_{i=1}^{N} \mathbf{1}\{\text{S1}_{i}\}\cdot \mathbf{1}\{\text{S2}_{i}\}\cdot \text{IoU}_{i}$, $\text{IoU}_{i}$ is the per-concept S3 IoU, set to zero if the model produces no S3 prediction for concept $i$. An equivalent form makes the components explicit: $\text{Joint} = \frac{n_{\text{S1}\cap\text{S2}}}{N}\cdot \overline{\text{IoU}}_{\text{S1}\cap\text{S2}}$, where $n_{\text{S1}\cap\text{S2}}$ is the number of concepts with both S1 and S2 correct, and $\overline{\text{IoU}}_{\text{S1}\cap\text{S2}}$ is the mean S3 IoU over those concepts. For example, with $n_{\text{S1}\cap\text{S2}}=138$ and $\overline{\text{IoU}}_{\text{S1}\cap\text{S2}}=0.32$, $\text{Joint} = (138/306)\cdot 0.32 = 14.4\%$.

The main S1/S2/S3 experiments all use a single query form (Latin-common), so the split is a post hoc property of the countries, not an experimentally manipulated condition.

\paragraph{Implementation Details.}

In all evaluated models, each frame is resized to $224\!\times\!224$ before being passed to the model. S1/S2 prompts use the four-way MCQ template shown in Figure~\ref{fig:teaser}; S3 prompts ask the model to return the predicted $[\hat{t}_{\text{start}}, \hat{t}_{\text{end}}]$ span in seconds.

\paragraph{API and compute budget.}

Closed-source inference was conducted through the Gemini and OpenAI APIs. The total API cost for the full experiment grid (3 stages $\times$ 3 modes on S1/S2 plus three modality ablations on S3) was approximately USD~700, split between Gemini and GPT. Open-source models were run on a single multi-GPU node.

\section{Additional Results}
\label{sec:app_results}

\subsection{Statistical Reliability}
\label{sec:app_stats}

All confidence intervals in this paper come from a cluster bootstrap over concepts: the $306$
concepts are resampled with replacement $10{,}000$ times, and each interval is the $2.5$ to
$97.5$ percentile range of the resampled statistic. The concept is the resampling unit because
questions within a concept share source video and annotator, so treating questions as
independent would understate uncertainty. All six models are evaluated on the same $306$
concepts and the same $631$ Stage-3 pairs, so every cell below is directly comparable across models. Tables~\ref{tab:app_ci_s1}, \ref{tab:app_ci_s2} and~\ref{tab:app_ci_miou} give the per-cell intervals for S1, S2 and mIoU; Table~\ref{tab:app_macro_micro} compares micro and macro averages, and Table~\ref{tab:app_paired_bootstrap} tests the pairwise model differences. Significance markers are per-comparison; we do not correct for multiple comparisons. The script that produces every number in this subsection is released with the benchmark.



\begin{table*}[t]
\centering
\small
\setlength{\tabcolsep}{4pt}
\begin{tabular}{lrcccccc}
\toprule
Cell & $n$ & Gemini & GPT-5.4 & Qwen3-VL & Qwen3.5 & IVL (n) & IVL (t) \\
\midrule
Indonesia & 48 & \ci{85.4}{74.5}{94.6} & \ci{70.8}{57.1}{83.3} & \ci{56.2}{42.0}{70.3} & \ci{58.3}{44.0}{72.4} & \ci{47.9}{33.3}{62.0} & \ci{54.2}{40.0}{68.1} \\
Malaysia & 44 & \ci{86.4}{75.6}{95.7} & \ci{86.4}{75.6}{95.8} & \ci{52.3}{37.5}{66.7} & \ci{61.4}{46.3}{75.7} & \ci{54.5}{39.4}{69.6} & \ci{43.2}{28.6}{58.3} \\
Philippines & 52 & \ci{84.6}{74.4}{93.8} & \ci{76.9}{64.6}{87.7} & \ci{55.8}{42.5}{69.0} & \ci{57.7}{44.2}{71.1} & \ci{48.1}{34.5}{62.2} & \ci{42.3}{28.8}{55.9} \\
Vietnam & 55 & \ci{89.1}{80.0}{96.5} & \ci{74.5}{62.7}{85.7} & \ci{52.7}{39.3}{66.1} & \ci{58.2}{44.8}{71.2} & \ci{45.5}{32.6}{58.8} & \ci{32.7}{20.8}{45.5} \\
Cambodia & 28 & \ci{67.9}{50.0}{84.6} & \ci{57.1}{38.5}{75.7} & \ci{50.0}{31.0}{69.2} & \ci{32.1}{15.4}{50.0} & \ci{53.6}{34.5}{71.4} & \ci{39.3}{21.4}{58.3} \\
Myanmar & 37 & \ci{56.8}{40.0}{72.7} & \ci{62.2}{46.2}{77.5} & \ci{56.8}{40.0}{72.7} & \ci{45.9}{29.6}{62.9} & \ci{45.9}{30.0}{62.5} & \ci{45.9}{30.0}{62.5} \\
Thailand & 42 & \ci{61.9}{46.8}{76.5} & \ci{59.5}{44.1}{74.2} & \ci{31.0}{17.1}{45.7} & \ci{50.0}{34.2}{65.2} & \ci{33.3}{19.2}{48.6} & \ci{31.0}{17.1}{45.2} \\
\midrule
Celebration & 91 & \ci{81.3}{73.1}{89.1} & \ci{78.0}{69.2}{86.2} & \ci{53.8}{43.6}{64.2} & \ci{53.8}{43.5}{64.1} & \ci{50.5}{40.4}{60.8} & \ci{38.5}{28.4}{48.4} \\
Dance & 60 & \ci{83.3}{73.2}{92.4} & \ci{71.7}{60.0}{82.8} & \ci{50.0}{37.3}{62.5} & \ci{55.0}{42.6}{67.7} & \ci{53.3}{40.7}{66.1} & \ci{40.0}{27.4}{52.6} \\
Game & 76 & \ci{72.4}{62.0}{82.2} & \ci{64.5}{53.2}{75.0} & \ci{47.4}{36.2}{58.8} & \ci{55.3}{44.2}{66.3} & \ci{32.9}{22.6}{43.7} & \ci{44.7}{33.7}{56.0} \\
Music & 68 & \ci{73.5}{62.9}{83.6} & \ci{66.2}{54.5}{77.3} & \ci{51.5}{39.7}{63.1} & \ci{51.5}{39.4}{63.3} & \ci{51.5}{39.7}{63.2} & \ci{41.2}{29.4}{53.0} \\
Wedding & 11 & \ci{81.8}{55.6}{100.0} & \ci{81.8}{55.6}{100.0} & \ci{54.5}{22.2}{85.7} & \ci{45.5}{15.4}{77.8} & \ci{45.5}{15.4}{77.0} & \ci{45.5}{14.3}{77.8} \\
\midrule
All (micro) & 306 & \ci{77.8}{73.2}{82.4} & \ci{70.9}{65.7}{75.8} & \ci{51.0}{45.4}{56.5} & \ci{53.6}{48.0}{59.2} & \ci{46.7}{41.2}{52.3} & \ci{41.2}{35.6}{46.7} \\
\bottomrule
\end{tabular}
\caption{Cluster-bootstrap 95\% confidence intervals for \textsc{S1} (\%) under Carry with the base prompt, resampling the $306$ concepts with replacement ($10{,}000$ resamples, seed $42$, percentile method); $n$ is concepts per cell. The statistic is concept-level: each resampled concept contributes one binary outcome. All six models are evaluated on the same $306$ concepts and $631$ Stage-3 pairs, so cells are directly comparable. Model abbreviations: Gemini~=~Gemini 3.1 Pro, GPT-5.4, Qwen3-VL~=~Qwen3-VL-32B, Qwen3.5~=~Qwen3.5-27B, IVL~=~InternVL3.5-14B with (t) and without (n) inference-time reasoning.}
\label{tab:app_ci_s1}
\end{table*}

\begin{table*}[t]
\centering
\small
\setlength{\tabcolsep}{4pt}
\begin{tabular}{lrcccccc}
\toprule
Cell & $n$ & Gemini & GPT-5.4 & Qwen3-VL & Qwen3.5 & IVL (n) & IVL (t) \\
\midrule
Indonesia & 48 & \ci{81.2}{69.4}{91.8} & \ci{68.8}{55.3}{81.5} & \ci{29.2}{16.7}{42.6} & \ci{50.0}{35.7}{64.3} & \ci{22.9}{11.5}{35.6} & \ci{25.0}{13.0}{37.8} \\
Malaysia & 44 & \ci{86.4}{75.0}{95.6} & \ci{77.3}{63.8}{89.2} & \ci{31.8}{18.2}{46.2} & \ci{40.9}{26.2}{56.0} & \ci{22.7}{11.1}{36.0} & \ci{27.3}{14.3}{40.8} \\
Philippines & 52 & \ci{82.7}{71.8}{92.3} & \ci{75.0}{62.7}{86.3} & \ci{36.5}{23.7}{50.0} & \ci{40.4}{27.3}{54.2} & \ci{25.0}{13.6}{37.0} & \ci{19.2}{8.9}{30.8} \\
Vietnam & 55 & \ci{81.8}{70.9}{91.5} & \ci{80.0}{69.0}{90.0} & \ci{30.9}{18.9}{43.5} & \ci{43.6}{30.3}{57.1} & \ci{25.5}{14.1}{37.1} & \ci{36.4}{24.0}{50.0} \\
Cambodia & 28 & \ci{53.6}{34.5}{72.0} & \ci{67.9}{50.0}{85.2} & \ci{25.0}{9.5}{42.3} & \ci{28.6}{12.1}{46.7} & \ci{35.7}{18.2}{54.2} & \ci{25.0}{10.0}{42.3} \\
Myanmar & 37 & \ci{56.8}{40.0}{72.4} & \ci{59.5}{43.2}{75.0} & \ci{40.5}{24.3}{56.8} & \ci{40.5}{25.0}{57.1} & \ci{21.6}{9.1}{35.5} & \ci{32.4}{17.6}{48.3} \\
Thailand & 42 & \ci{52.4}{37.2}{67.6} & \ci{42.9}{28.1}{58.0} & \ci{23.8}{11.4}{37.0} & \ci{33.3}{19.4}{47.6} & \ci{21.4}{9.8}{34.3} & \ci{21.4}{9.5}{34.8} \\
\midrule
Celebration & 91 & \ci{76.9}{68.0}{85.3} & \ci{78.0}{69.3}{86.3} & \ci{31.9}{22.5}{41.5} & \ci{44.0}{33.7}{54.1} & \ci{25.3}{16.5}{34.4} & \ci{25.3}{16.3}{34.4} \\
Dance & 60 & \ci{73.3}{61.8}{84.2} & \ci{66.7}{54.4}{78.6} & \ci{31.7}{20.0}{43.8} & \ci{35.0}{23.1}{47.1} & \ci{23.3}{13.1}{34.5} & \ci{28.3}{17.2}{40.0} \\
Game & 76 & \ci{71.1}{60.5}{81.2} & \ci{57.9}{46.7}{68.8} & \ci{32.9}{22.4}{43.5} & \ci{38.2}{27.1}{49.3} & \ci{23.7}{14.5}{33.8} & \ci{28.9}{18.8}{39.7} \\
Music & 68 & \ci{67.6}{56.2}{78.3} & \ci{67.6}{56.1}{78.5} & \ci{30.9}{20.0}{42.2} & \ci{45.6}{33.8}{58.0} & \ci{26.5}{16.4}{37.1} & \ci{25.0}{15.1}{35.9} \\
Wedding & 11 & \ci{81.8}{55.6}{100.0} & \ci{72.7}{42.9}{100.0} & \ci{18.2}{0.0}{44.4} & \ci{27.3}{0.0}{57.1} & \ci{18.2}{0.0}{44.4} & \ci{27.3}{0.0}{57.1} \\
\midrule
All (micro) & 306 & \ci{72.9}{67.6}{77.8} & \ci{68.3}{63.1}{73.5} & \ci{31.4}{26.1}{36.6} & \ci{40.5}{35.0}{46.1} & \ci{24.5}{19.9}{29.4} & \ci{26.8}{21.9}{32.0} \\
\bottomrule
\end{tabular}
\caption{Cluster-bootstrap 95\% confidence intervals for \textsc{S2} (\%) under Carry with the base prompt, resampling the $306$ concepts with replacement ($10{,}000$ resamples, seed $42$, percentile method); $n$ is concepts per cell. The statistic is concept-level: each resampled concept contributes one binary outcome. All six models are evaluated on the same $306$ concepts and $631$ Stage-3 pairs, so cells are directly comparable. Model abbreviations: Gemini~=~Gemini 3.1 Pro, GPT-5.4, Qwen3-VL~=~Qwen3-VL-32B, Qwen3.5~=~Qwen3.5-27B, IVL~=~InternVL3.5-14B with (t) and without (n) inference-time reasoning.}
\label{tab:app_ci_s2}
\end{table*}

\begin{table*}[t]
\centering
\small
\setlength{\tabcolsep}{4pt}
\begin{tabular}{lrcccccc}
\toprule
Cell & $n$ & Gemini & GPT-5.4 & Qwen3-VL & Qwen3.5 & IVL (n) & IVL (t) \\
\midrule
Indonesia & 48 & \ci{33.4}{24.8}{44.0} & \ci{27.2}{19.4}{38.2} & \ci{23.3}{16.9}{30.8} & \ci{25.1}{18.5}{33.3} & \ci{2.5}{1.2}{4.4} & \ci{4.7}{2.1}{8.4} \\
Malaysia & 44 & \ci{33.4}{23.0}{47.1} & \ci{37.0}{26.8}{45.6} & \ci{23.9}{13.2}{34.5} & \ci{26.8}{16.6}{39.6} & \ci{3.2}{1.2}{5.9} & \ci{2.2}{1.0}{3.7} \\
Philippines & 52 & \ci{25.3}{14.3}{47.1} & \ci{24.8}{11.9}{46.3} & \ci{15.1}{9.5}{28.9} & \ci{17.8}{11.1}{31.2} & \ci{2.7}{1.1}{5.8} & \ci{2.0}{0.6}{5.3} \\
Vietnam & 55 & \ci{30.3}{23.5}{38.1} & \ci{27.8}{20.3}{37.1} & \ci{20.4}{13.2}{29.7} & \ci{20.5}{14.1}{29.1} & \ci{2.2}{0.6}{4.0} & \ci{2.3}{0.4}{4.7} \\
Cambodia & 28 & \ci{53.3}{38.1}{71.0} & \ci{47.4}{31.1}{67.0} & \ci{36.8}{22.6}{53.9} & \ci{28.9}{16.7}{44.1} & \ci{3.4}{0.0}{8.7} & \ci{4.0}{0.3}{9.0} \\
Myanmar & 37 & \ci{43.6}{31.4}{56.7} & \ci{35.2}{23.8}{46.7} & \ci{32.6}{20.0}{45.9} & \ci{32.3}{20.0}{45.1} & \ci{1.9}{0.4}{3.7} & \ci{2.2}{0.0}{6.3} \\
Thailand & 42 & \ci{41.3}{28.6}{59.3} & \ci{42.7}{30.2}{60.6} & \ci{19.0}{9.4}{27.5} & \ci{27.6}{17.1}{42.5} & \ci{0.7}{0.0}{2.0} & \ci{1.6}{0.0}{4.7} \\
\midrule
Celebration & 91 & \ci{39.8}{33.8}{47.1} & \ci{36.9}{29.7}{44.6} & \ci{29.5}{22.9}{36.7} & \ci{26.5}{19.8}{34.4} & \ci{3.0}{1.7}{4.7} & \ci{1.5}{0.7}{2.6} \\
Dance & 60 & \ci{30.5}{23.6}{39.6} & \ci{31.7}{24.8}{40.9} & \ci{16.9}{9.3}{23.1} & \ci{27.5}{21.5}{35.3} & \ci{3.3}{1.4}{5.7} & \ci{4.3}{1.3}{8.4} \\
Game & 76 & \ci{37.7}{30.1}{46.4} & \ci{29.9}{23.8}{37.8} & \ci{19.7}{14.0}{25.7} & \ci{24.3}{18.5}{31.2} & \ci{2.0}{0.6}{4.0} & \ci{5.0}{2.0}{8.4} \\
Music & 68 & \ci{56.4}{44.6}{67.8} & \ci{53.6}{41.1}{65.6} & \ci{36.1}{24.1}{48.6} & \ci{38.5}{26.7}{50.7} & \ci{3.7}{1.2}{7.0} & \ci{4.0}{0.7}{8.1} \\
Wedding & 11 & \ci{16.4}{8.9}{30.5} & \ci{16.3}{7.2}{32.7} & \ci{13.5}{8.2}{22.9} & \ci{13.9}{7.8}{23.9} & \ci{1.1}{0.2}{1.8} & \ci{2.4}{0.6}{6.1} \\
\midrule
All (micro) & 306 & \ci{33.4}{28.3}{39.1} & \ci{31.0}{25.8}{36.7} & \ci{22.3}{18.5}{26.4} & \ci{24.1}{20.1}{28.6} & \ci{2.5}{1.8}{3.3} & \ci{3.0}{1.9}{4.3} \\
\bottomrule
\end{tabular}
\caption{Cluster-bootstrap 95\% confidence intervals for \textsc{mIoU} (\%) under Carry with the base prompt, resampling the $306$ concepts with replacement ($10{,}000$ resamples, seed $42$, percentile method); $n$ is concepts per cell. The statistic is pair-weighted: each resampled concept contributes all of its Stage-3 pairs, so concepts drawn twice count twice. All six models are evaluated on the same $306$ concepts and $631$ Stage-3 pairs, so cells are directly comparable. Model abbreviations: Gemini~=~Gemini 3.1 Pro, GPT-5.4, Qwen3-VL~=~Qwen3-VL-32B, Qwen3.5~=~Qwen3.5-27B, IVL~=~InternVL3.5-14B with (t) and without (n) inference-time reasoning.}
\label{tab:app_ci_miou}
\end{table*}

\begin{table}[t]
\centering
\small
\setlength{\tabcolsep}{3pt}
\begin{tabular}{llccc}
\toprule
Model & Scope & \textsc{S1} & \textsc{S2} & mIoU \\
\multicolumn{5}{c}{\footnotesize all values in \%, matching Table~\ref{tab:overall}} \\
\midrule
\multirow{3}{*}{Gemini}
 & micro & \ci{77.8}{73.2}{82.4} & \ci{72.9}{67.6}{77.8} & \ci{33.4}{28.3}{39.1} \\
 & macro-ctry & \ci{76.0}{71.1}{80.8} & \ci{70.7}{65.5}{75.7} & \ci{37.2}{33.1}{43.5} \\
 & macro-cat & \ci{78.5}{72.1}{84.2} & \ci{74.2}{67.5}{80.1} & \ci{36.2}{32.5}{40.7} \\
\midrule
\multirow{3}{*}{GPT-5.4}
 & micro & \ci{70.9}{65.7}{75.8} & \ci{68.3}{63.1}{73.5} & \ci{31.0}{25.8}{36.7} \\
 & macro-ctry & \ci{69.6}{64.2}{74.7} & \ci{67.3}{61.9}{72.5} & \ci{34.6}{30.4}{40.7} \\
 & macro-cat & \ci{72.4}{65.7}{78.4} & \ci{68.6}{61.3}{75.3} & \ci{33.7}{29.8}{38.6} \\
\midrule
\multirow{3}{*}{Qwen3-VL}
 & micro & \ci{51.0}{45.4}{56.5} & \ci{31.4}{26.1}{36.6} & \ci{22.3}{18.5}{26.4} \\
 & macro-ctry & \ci{50.7}{45.0}{56.4} & \ci{31.1}{25.9}{36.5} & \ci{24.4}{20.8}{28.8} \\
 & macro-cat & \ci{51.4}{43.6}{59.2} & \ci{29.1}{23.0}{36.0} & \ci{23.1}{19.7}{27.0} \\
\midrule
\multirow{3}{*}{Qwen3.5}
 & micro & \ci{53.6}{48.0}{59.2} & \ci{40.5}{35.0}{46.1} & \ci{24.1}{20.1}{28.6} \\
 & macro-ctry & \ci{52.0}{46.4}{57.6} & \ci{39.6}{34.0}{45.2} & \ci{25.6}{22.0}{30.4} \\
 & macro-cat & \ci{52.2}{44.7}{60.0} & \ci{38.0}{31.1}{45.5} & \ci{26.2}{22.7}{30.2} \\
\midrule
\multirow{3}{*}{IVL (n)}
 & micro & \ci{46.7}{41.2}{52.3} & \ci{24.5}{19.9}{29.4} & \ci{2.5}{1.8}{3.3} \\
 & macro-ctry & \ci{47.0}{41.3}{52.8} & \ci{25.0}{20.2}{30.1} & \ci{2.4}{1.6}{3.4} \\
 & macro-cat & \ci{46.7}{39.2}{54.5} & \ci{23.4}{17.6}{29.9} & \ci{2.6}{1.8}{3.6} \\
\midrule
\multirow{3}{*}{IVL (t)}
 & micro & \ci{41.2}{35.6}{46.7} & \ci{26.8}{21.9}{32.0} & \ci{3.0}{1.9}{4.3} \\
 & macro-ctry & \ci{41.2}{35.6}{46.9} & \ci{26.7}{21.6}{31.8} & \ci{2.7}{1.8}{4.0} \\
 & macro-cat & \ci{42.0}{34.3}{49.8} & \ci{27.0}{20.3}{34.2} & \ci{3.5}{2.2}{5.0} \\
\bottomrule
\end{tabular}
\caption{Micro and macro averages with cluster-bootstrap 95\% confidence intervals (Carry, base prompt). \emph{micro} pools all $306$ concepts and, for mIoU, all $631$ Stage-3 pairs. \emph{macro-ctry} and \emph{macro-cat} give each of the $7$ countries and $5$ categories equal weight. Macro intervals are bootstrapped through the same concept resamples as the cells, not assembled from their bounds. \textsc{S1} and \textsc{S2} agree within $2.5$ points across every model and scope. All values are percentages, matching Table~\ref{tab:overall}. mIoU runs $1$ to $4$ points higher under macro because Stage-3 pair mass is concentrated in Wedding, which supplies $3.6\%$ of concepts but $23.5\%$ of pairs and is the hardest cell for every model. Model abbreviations: Gemini~=~Gemini 3.1 Pro, GPT-5.4, Qwen3-VL~=~Qwen3-VL-32B, Qwen3.5~=~Qwen3.5-27B, IVL~=~InternVL3.5-14B with (t) and without (n) inference-time reasoning.}
\label{tab:app_macro_micro}
\end{table}

\begin{table}[t]
\centering
\small
\setlength{\tabcolsep}{1.5pt}
\begin{tabular}{@{}lccc@{}}
\toprule
Comparison & $\Delta$\textsc{S1} & $\Delta$\textsc{S2} & $\Delta$mIoU \\
\midrule
Gemini vs.\ GPT-5.4 & \ci{\textbf{+6.9}}{+2.3}{+11.4} & \ci{+4.6}{-0.7}{+9.8} & \ci{+2.5}{-1.6}{+6.4} \\
Gemini vs.\ Qwen3-VL & \ci{\textbf{+26.8}}{+20.6}{+33.0} & \ci{\textbf{+41.5}}{+34.6}{+48.4} & \ci{\textbf{+11.2}}{+7.8}{+15.2} \\
Gemini vs.\ Qwen3.5 & \ci{\textbf{+24.2}}{+17.6}{+30.7} & \ci{\textbf{+32.4}}{+25.5}{+39.2} & \ci{\textbf{+9.4}}{+5.7}{+13.5} \\
Gemini vs.\ IVL (n) & \ci{\textbf{+31.0}}{+24.5}{+37.6} & \ci{\textbf{+48.4}}{+41.2}{+55.2} & \ci{\textbf{+31.0}}{+26.0}{+36.4} \\
Gemini vs.\ IVL (t) & \ci{\textbf{+36.6}}{+29.7}{+43.5} & \ci{\textbf{+46.1}}{+38.9}{+52.9} & \ci{\textbf{+30.4}}{+25.5}{+35.9} \\
GPT-5.4 vs.\ Qwen3-VL & \ci{\textbf{+19.9}}{+13.4}{+26.1} & \ci{\textbf{+36.9}}{+30.1}{+43.5} & \ci{\textbf{+8.7}}{+4.7}{+13.2} \\
GPT-5.4 vs.\ Qwen3.5 & \ci{\textbf{+17.3}}{+10.5}{+23.9} & \ci{\textbf{+27.8}}{+21.2}{+34.3} & \ci{\textbf{+6.9}}{+3.0}{+11.3} \\
GPT-5.4 vs.\ IVL (n) & \ci{\textbf{+24.2}}{+17.6}{+31.0} & \ci{\textbf{+43.8}}{+36.9}{+50.7} & \ci{\textbf{+28.5}}{+23.4}{+34.1} \\
GPT-5.4 vs.\ IVL (t) & \ci{\textbf{+29.7}}{+22.9}{+36.6} & \ci{\textbf{+41.5}}{+34.3}{+48.4} & \ci{\textbf{+28.0}}{+23.0}{+33.6} \\
Qwen3-VL vs.\ Qwen3.5 & \ci{-2.6}{-8.8}{+3.6} & \ci{\textbf{-9.2}}{-14.7}{-3.6} & \ci{-1.8}{-4.8}{+1.3} \\
Qwen3-VL vs.\ IVL (n) & \ci{+4.2}{-2.3}{+10.8} & \ci{\textbf{+6.9}}{+0.3}{+13.7} & \ci{\textbf{+19.8}}{+16.0}{+23.9} \\
Qwen3-VL vs.\ IVL (t) & \ci{\textbf{+9.8}}{+2.9}{+16.3} & \ci{+4.6}{-2.6}{+11.4} & \ci{\textbf{+19.3}}{+15.5}{+23.5} \\
Qwen3.5 vs.\ IVL (n) & \ci{\textbf{+6.9}}{+0.0}{+13.4} & \ci{\textbf{+16.0}}{+9.2}{+23.2} & \ci{\textbf{+21.6}}{+17.6}{+26.1} \\
Qwen3.5 vs.\ IVL (t) & \ci{\textbf{+12.4}}{+5.6}{+19.0} & \ci{\textbf{+13.7}}{+6.2}{+21.2} & \ci{\textbf{+21.1}}{+17.3}{+25.4} \\
IVL (n) vs.\ IVL (t) & \ci{+5.6}{-0.7}{+11.8} & \ci{-2.3}{-8.2}{+3.9} & \ci{-0.5}{-1.7}{+0.5} \\
\bottomrule
\end{tabular}
\caption{Paired cluster bootstrap on pairwise model differences (Carry, base prompt, micro). Because all six models are evaluated on the same $306$ concepts, the difference is taken \emph{within} each of the $10{,}000$ concept resamples; this is materially more powerful than checking whether two marginal intervals overlap. $\Delta$mIoU is in mIoU points $\times 100$. \textbf{Bold} marks differences whose $95\%$ interval excludes zero. Every closed-weight against open-weight comparison is significant on all three metrics; the Gemini lead over GPT-5.4 is significant on \textsc{S1} only; and inference-time reasoning yields no measurable difference for InternVL3.5-14B. Model abbreviations: Gemini~=~Gemini 3.1 Pro, GPT-5.4, Qwen3-VL~=~Qwen3-VL-32B, Qwen3.5~=~Qwen3.5-27B, IVL~=~InternVL3.5-14B with (t) and without (n) inference-time reasoning. Significance markers are per-comparison; we do not correct for multiple comparisons.}
\label{tab:app_paired_bootstrap}
\end{table}

\subsection{Modality Ablation Details}

We re-ran the three lowest-cost models (Gemini 3.1 Pro, Qwen3-VL-32B, Qwen3.5-27B) on the modality ablations; the remaining three (GPT-5.4, InternVL3.5-14B in both configurations) were excluded due to API and compute budget. Table~\ref{tab:modality_aggregate} reports the aggregate mIoU shift for each ablation.

\begin{table}[h]
\centering
\small
\begin{tabular}{lrrr}
\toprule
Ablation     & $n$ pairs & $\overline{\Delta}$ mIoU points & $p$ \\
\midrule
No-audio     & 631 & $-0.02$ & .55  \\
No-subtitles & 631 & $-0.65$ & .10  \\
Visual-only  & 631 & $-0.81$ & .045 \\
\bottomrule
\end{tabular}
\caption{Aggregate mIoU shift per modality ablation, pooled across the three ablation-instrumented models (Gemini~3.1~Pro, Qwen3-VL-32B, Qwen3.5-27B). $p$-values from paired tests against the baseline. Per-pair role classifications derived from these shifts appear in Figure~\ref{fig:modality_overview}.}
\label{tab:modality_aggregate}
\end{table}

\subsection{Modality Role by Country Cluster}
\label{app:cluster_stats}

The \textit{Distracting} role shares split sharply along the writing-system clusters, as Table~\ref{tab:cluster_stats} shows. For audio, the cluster gap is $+15$~points ($z{=}3.84$, $p{<}.001$ from a two-proportion $z$-test); for subtitles, the gap is smaller ($+8.5$~points) but points in the same direction.

\begin{table}[h]
\centering
\small
\resizebox{1\columnwidth}{!}{
\begin{tabular}{lrr}
\toprule
Cluster & \textit{Distr.} audio (\%) & \textit{Distr.} subs (\%) \\
\midrule
\shortstack[l]{Latin-script\\(ID/MY/PH/VN)}     & 14.0 & 11.9 \\ \midrule
\shortstack[l]{non-Latin-script\\(KH/MM/TH)}    & 29.1 & 20.4 \\
\bottomrule
\end{tabular}
}
\caption{Distracting role share (\%) by writing-system cluster, pooled across Gemini~3.1~Pro, Qwen3-VL-32B, and Qwen3.5-27B over the same 631 S3 pairs as Figure~\ref{fig:modality_overview}. Non-Latin-script cluster shows a $+15$\,pp audio gap ($p<.001$) and a $+8.5$\,pp subtitle gap.}
\label{tab:cluster_stats}
\end{table}

The Distracting concepts in the non-Latin-script cluster are dominated by traditional music concepts: of 26 such concepts, 8 are music (47\% of non-Latin music concepts with Stage-3 pairs vs 4\% in Latin), and 10 are traditional games. The music subset features solo instruments (Saung, Sralai, Chhing, Sor Duang, Ta-pone, Maung, Kong Chai, Tone) whose audio plays uniformly across the clip, providing no temporal signal for the specific sub-event the question asks about. The pattern is also consistent with audio being anchored less reliably for underrepresented languages in pretraining, though we cannot verify the training-side mechanism because the models' training mixtures are not disclosed.

\subsection{Sensitivity of the modality-role threshold}
\label{sec:app_threshold}

Section~\ref{sec:modality} bins each video by $\Delta$, the change in mIoU in percentage points when a modality is removed, with the cut at $5$ points. Table~\ref{tab:app_threshold} repeats the binning at $3$ and $7$ points over the same $631$ Stage-3 pairs. Moving the cut moves mass between \textit{Redundant} and the two active roles, as any threshold must, but neither claim in Section~\ref{sec:modality} depends on the choice. \textit{Redundant} remains the majority role for all three ablations at all three cuts, ranging from $56.3\%$ to $74.8\%$. The script-cluster gap also holds: for audio, the \textit{Distracting} share is $17\%$ for Latin-script countries ($n{=}528$ pairs) against $33\%$ for non-Latin-script countries ($n{=}103$) at $\pm 3$, $14\%$ against $29\%$ at $\pm 5$, and $11\%$ against $21\%$ at $\pm 7$.

\begin{table}[t]
\centering
\small
\setlength{\tabcolsep}{5pt}
\begin{tabular}{lccc}
\toprule
Ablation & Comp. & Redun. & Distr. \\
\midrule
\multicolumn{4}{l}{\emph{Cut at $\pm 3$ points}} \\
No-audio      & 20.6 & 59.6 & 19.8 \\
No-subtitles  & 20.3 & 63.5 & 16.2 \\
Visual-only   & 24.2 & 56.3 & 19.5 \\
\midrule
\multicolumn{4}{l}{\emph{Cut at $\pm 5$ points, used in the paper}} \\
No-audio      & 17.0 & 66.4 & 16.6 \\
No-subtitles  & 16.8 & 69.7 & 13.5 \\
Visual-only   & 20.8 & 63.1 & 16.2 \\
\midrule
\multicolumn{4}{l}{\emph{Cut at $\pm 7$ points}} \\
No-audio      & 15.2 & 71.8 & 13.0 \\
No-subtitles  & 14.1 & 74.8 & 11.1 \\
Visual-only   & 17.9 & 69.3 & 12.8 \\
\bottomrule
\end{tabular}
\caption{Modality-role shares as a percentage of the $631$ Stage-3 pairs, at three values of the classification cut. $\Delta$ is the mean change in mIoU across the three ablation-instrumented models. \textit{Redundant} stays the majority role in every row.}
\label{tab:app_threshold}
\end{table}

\subsection{Cascade deltas by model and mode}
\label{app:cascade_deltas}

\begin{figure*}[!htb]
    \centering
    \includegraphics[width=0.85\textwidth]{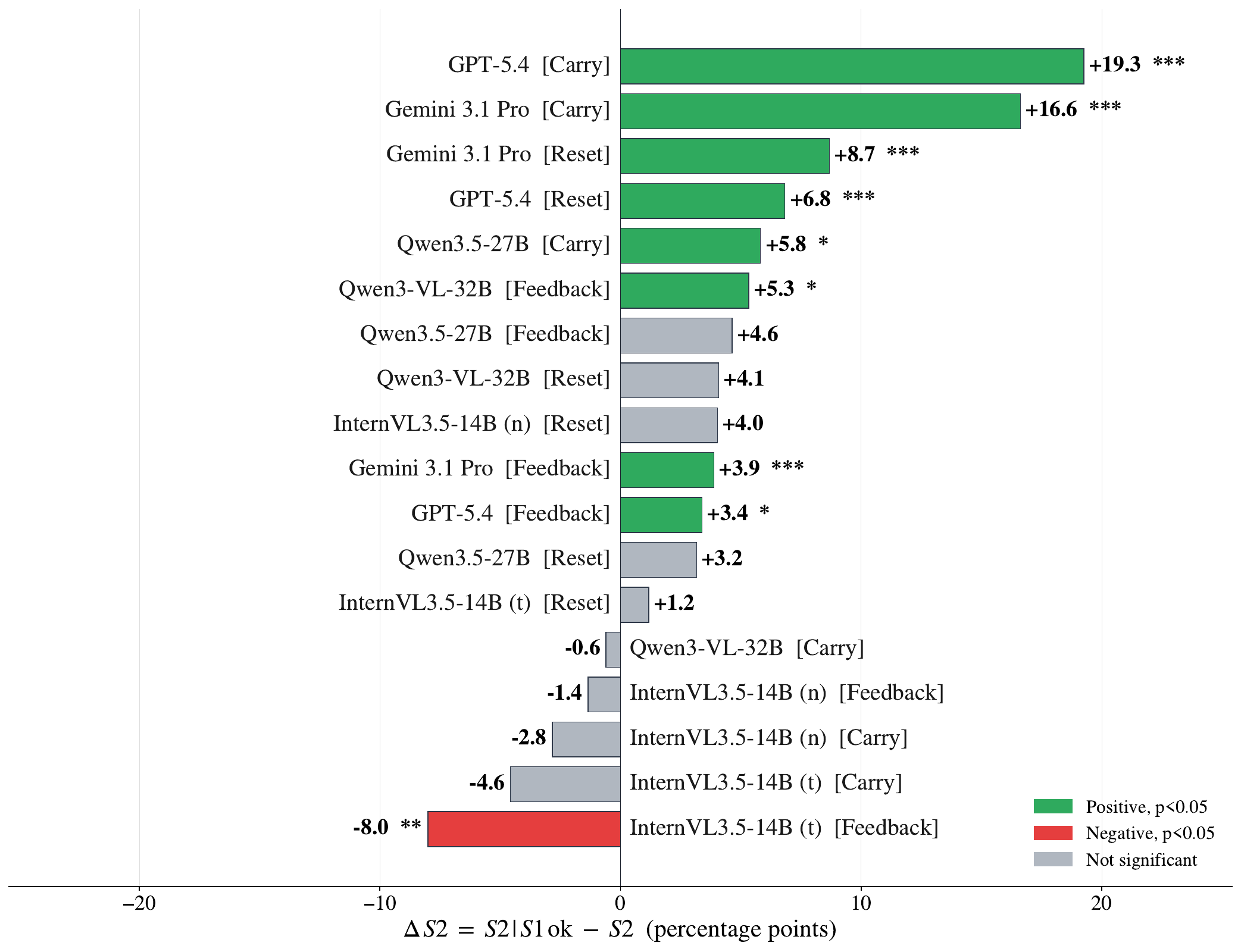}
    \includegraphics[width=0.85\textwidth]{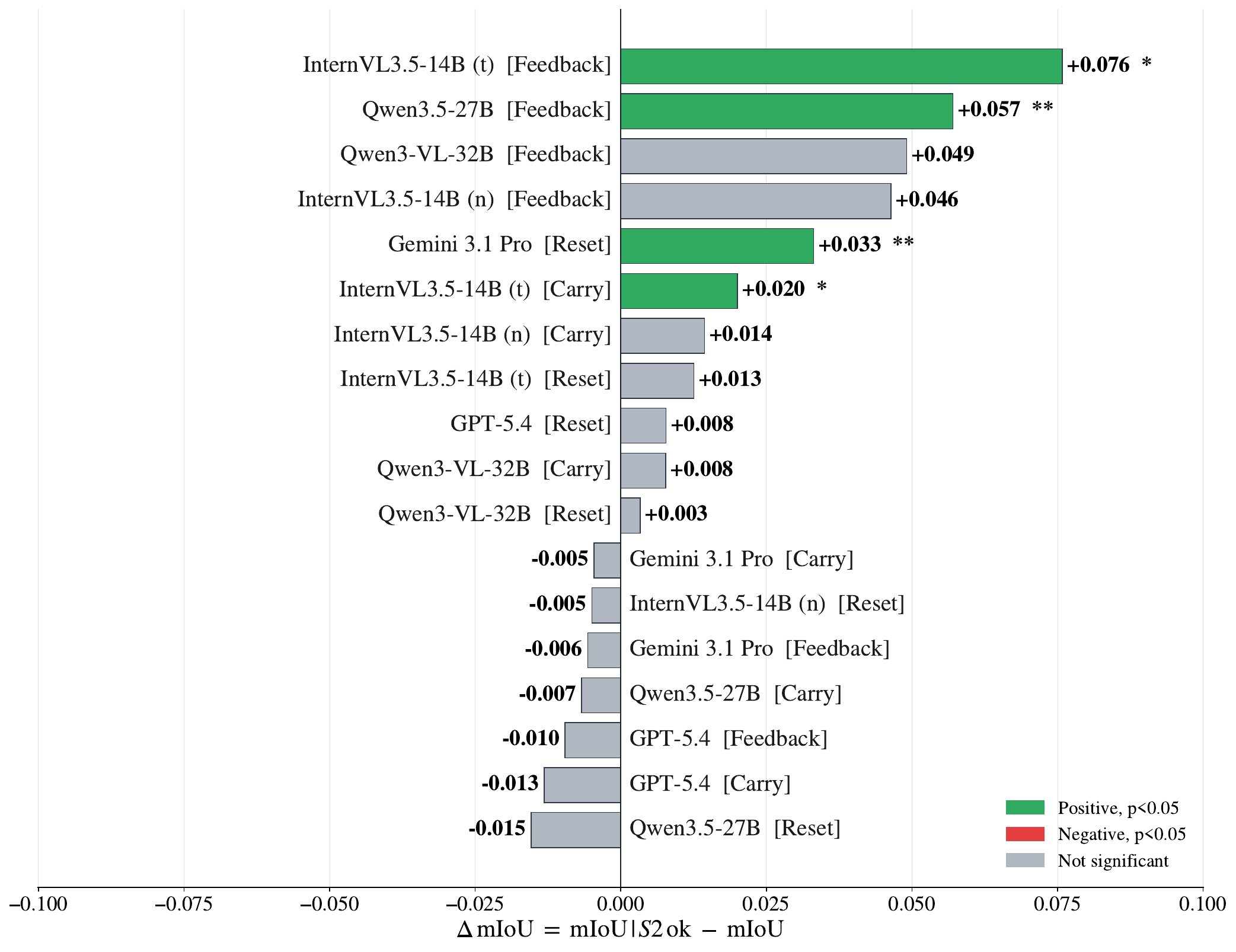}
    \caption{Per-(model, mode) cascade deltas. Top: $\Delta\text{S2} = \text{c-S2} - \text{S2}$ (percentage points), measuring how much correct naming lifts visual recognition. Bottom: $\Delta\text{mIoU} = \text{c-mIoU} - \text{mIoU}$, measuring how much correct visual recognition lifts temporal localization. Bars are sorted by effect size within each panel. Stars: $*\,p<0.05$, $**\,p<0.01$, $***\,p<0.001$ against the unconditional baseline. $\Delta\text{mIoU}$ is in IoU units on [0,1]; multiplied by 100, it gives the points used in the text. }
    \label{fig:app_cascade_delta}
\end{figure*}

Figure~\ref{fig:app_cascade_delta} reports the two cascade deltas behind Section~\ref{sec:cascade} for every model and mode: $\Delta$S2 = c-S2 $-$ S2, how much correct naming lifts visual recognition, and $\Delta$mIoU = c-mIoU $-$ mIoU, how much correct visual recognition lifts temporal localization. Significance is against the unconditional baseline for the same model and mode.

\subsection{Stage-3 result breakdowns}
\label{app:s3_breakdowns}

\begin{figure*}[!htb]
  \includegraphics[width=1\textwidth]{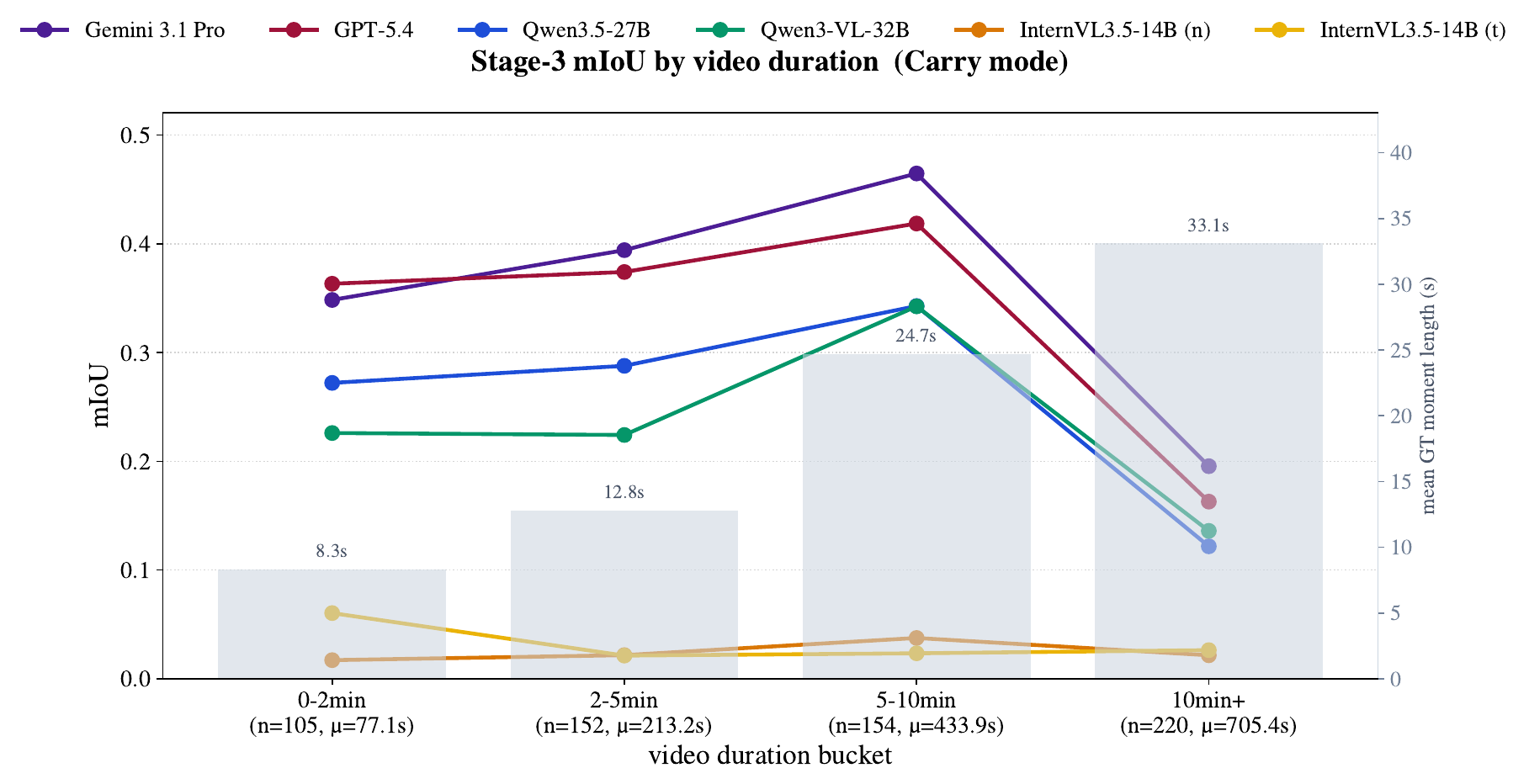}
  \caption{S3 mIoU under Carry mode, bucketed by Set~B source-video duration. Below each bucket: $n=$ number of S3 pairs in the bucket, and $\overline{\ell}=$ mean source-video duration. Markers annotate the mean ground-truth sub-event span in the bucket (right axis). Every model falls sharply once videos exceed about ten minutes, while mIoU rises across the shorter buckets as the mean sub-event span grows. Closed-source models degrade more gracefully. mIoU on [0,1].}
  \label{fig:app_video_length}
\end{figure*}

\begin{figure*}[!htb]
    \centering
    \includegraphics[width=0.9\textwidth]{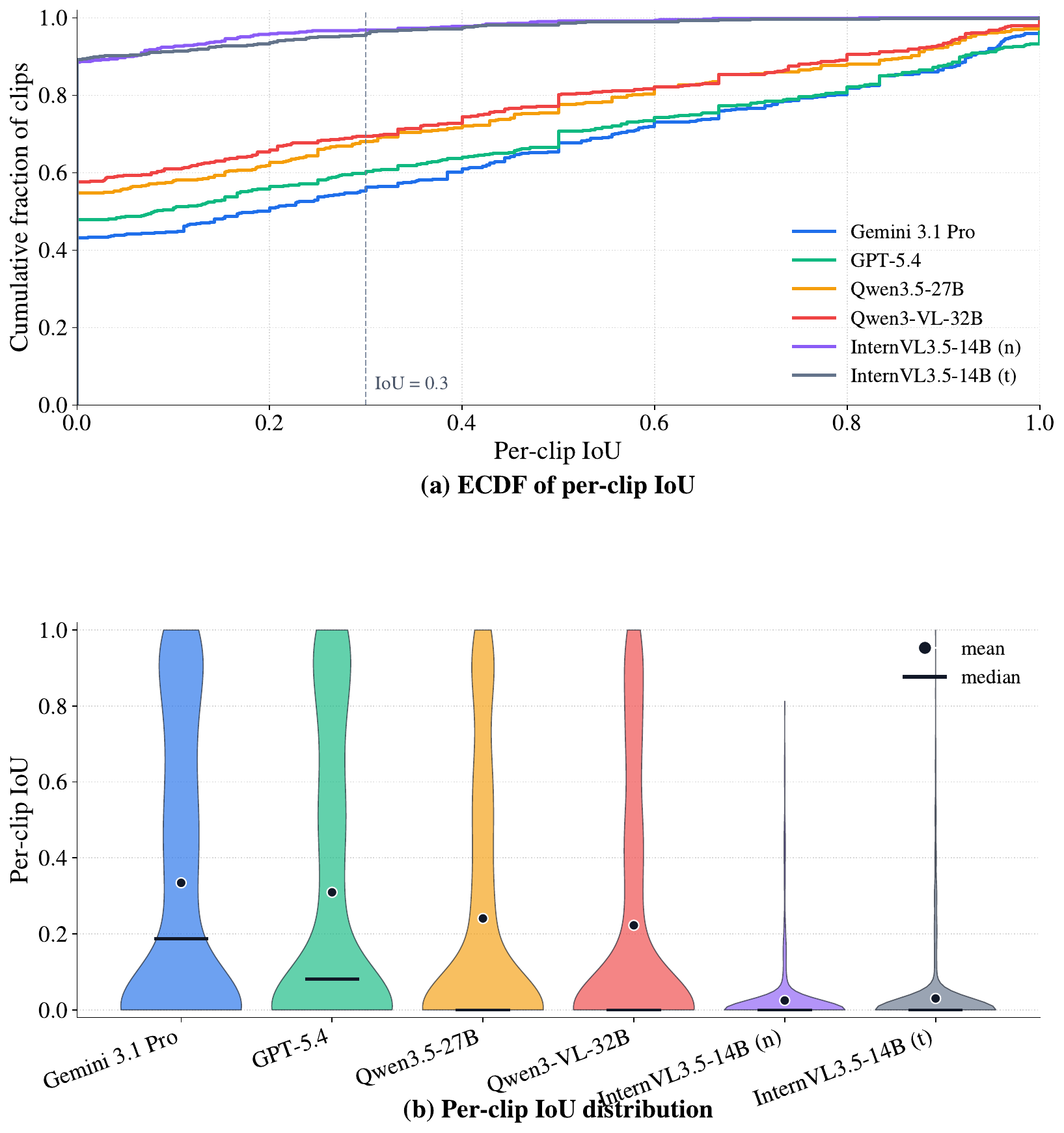}
    \caption{Per-clip S3 IoU distribution under Carry mode, across all 306 concepts. (a) ECDF of per-clip IoU; the dashed vertical line marks IoU$=$0.3; one minus the ECDF value at the line gives the IoU@0.3 hit rate per model. (b) Violin plot of the same distributions, with a strip of mean (filled dot) and median (horizontal bar) markers. A single mIoU number hides a long-tailed distribution with a heavy mass near zero for every model.}
    \label{fig:app_iou_dist}
\end{figure*}

\begin{figure*}[!htb]
    \includegraphics[width=0.95\textwidth]{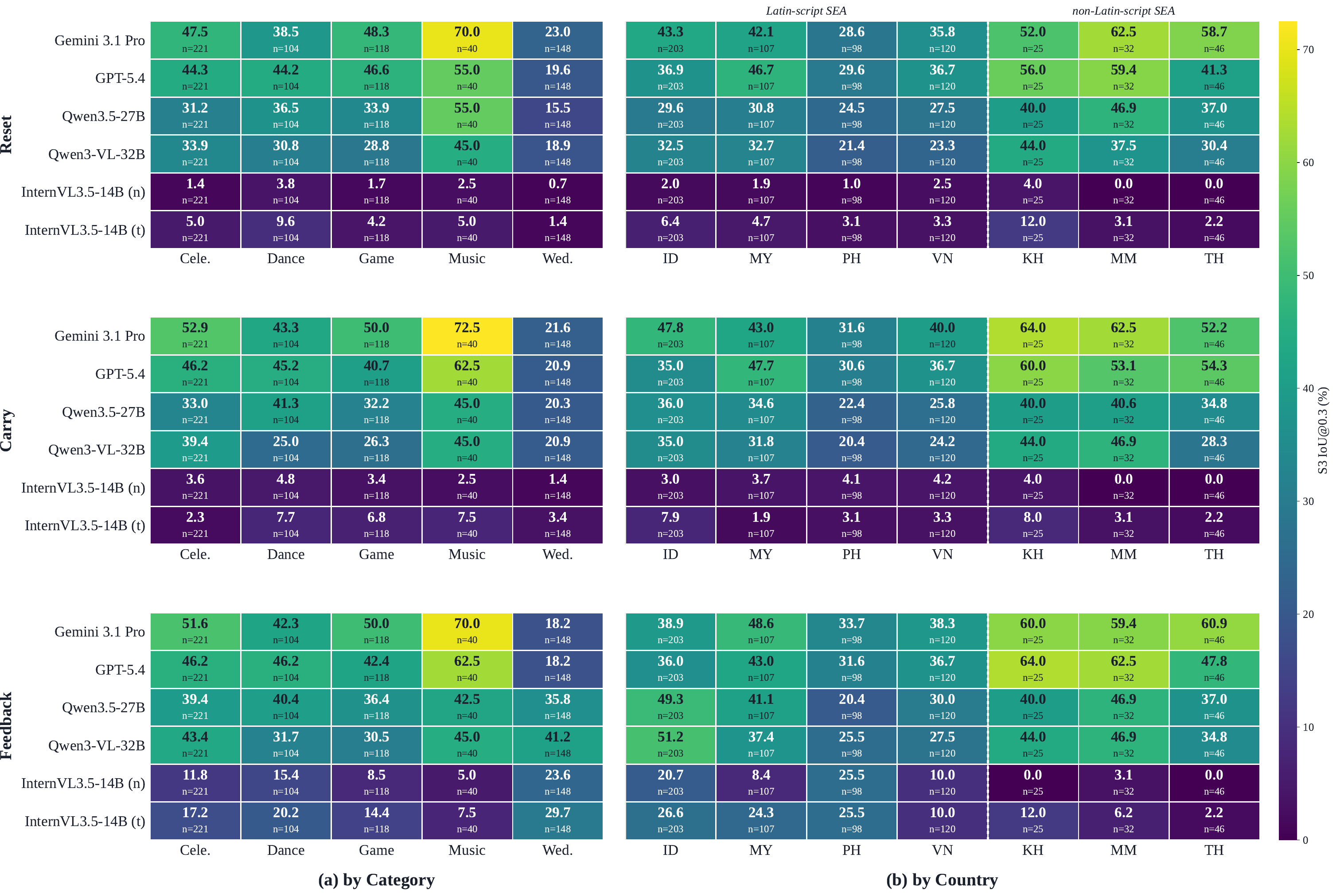}
    \caption{S3 IoU@0.3 (percentage of pairs with IoU $\ge 0.3$) for all six models under all three modes, split by category (left) and by country writing-system cluster (right). Cells report the percentage; row $n$ is the number of pairs aggregated.}
    \label{fig:app_s3_iou03}
\end{figure*}

\begin{figure*}[!htb]
    \includegraphics[width=0.95\textwidth]{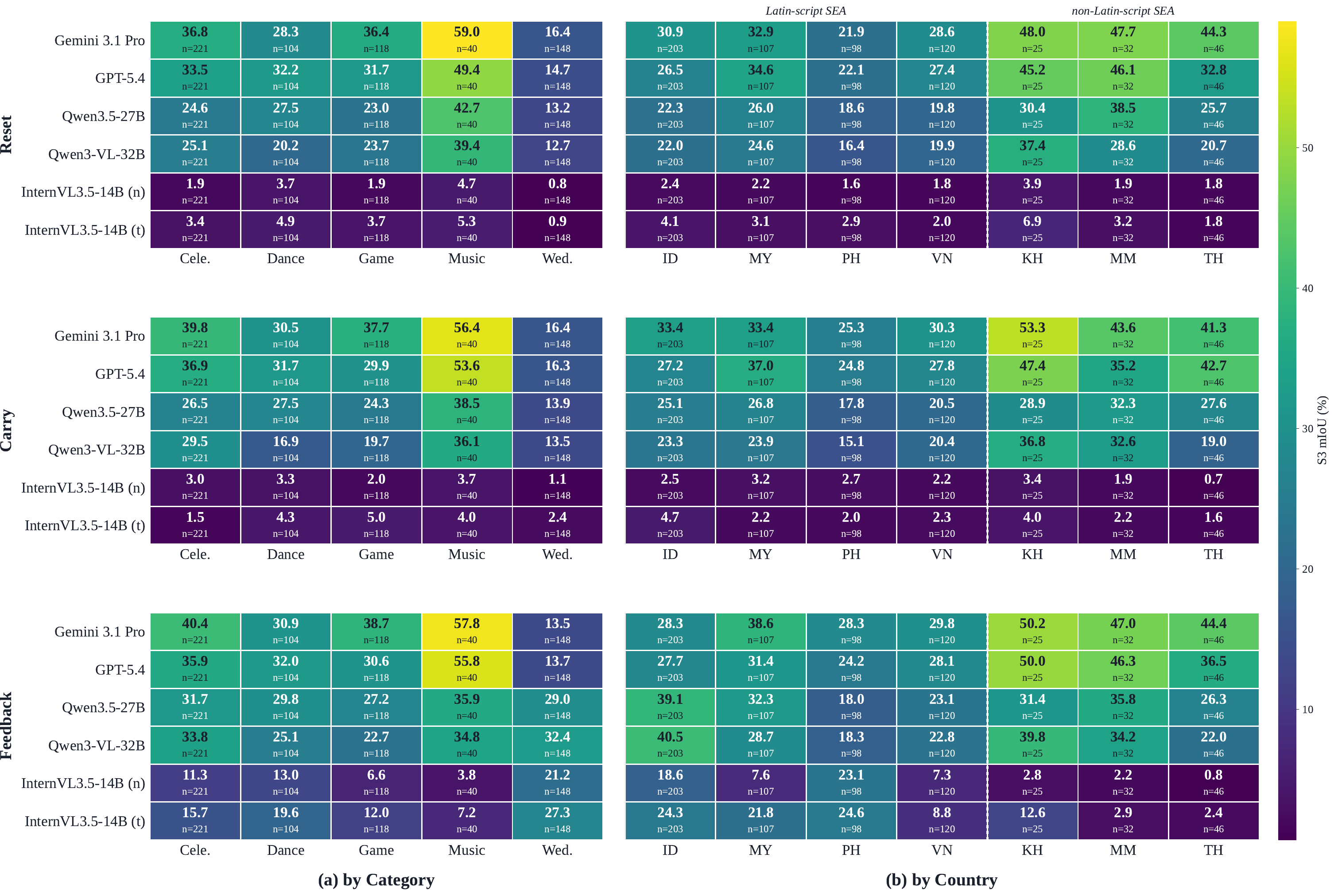}
    \caption{S3 mIoU (mean Intersection-over-Union) for all six models under all three modes, split by category (left) and by country writing-system cluster (right). Cells report mIoU in percent; row $n$ is the number of pairs aggregated.}
    \label{fig:app_s3_miou}
\end{figure*}

Four views of the Stage-3 results support Section~\ref{sec:overall}. Figure~\ref{fig:app_video_length} buckets mIoU by source-video duration: every model falls sharply once videos exceed about ten minutes, while the rise across shorter buckets tracks the growing mean sub-event span; closed-source models degrade more gracefully. A related bias is documented in text-video retrieval, where models favor particular clip lengths unless the effect is removed by causal intervention \cite{satar2023debiasing}. Figure~\ref{fig:app_iou_dist} gives the per-clip IoU distribution behind the mIoU averages, which is long-tailed with heavy mass near zero for every model. Figures~\ref{fig:app_s3_iou03} and~\ref{fig:app_s3_miou} report IoU@0.3 and mIoU for all six models under all three modes, split by category and by writing-system cluster.


\subsection{Specialized temporal-localization models on S3}
\label{sec:s3-specialized}

TempCompass \citep{liu2024tempcompass} decomposes temporal understanding for general video; CMB asks the same diagnostic question for culture-specific sub-events. To check whether models designed specifically for video temporal grounding outperform the general-purpose VLMs on S3, we evaluate TimeSuite~\citep{zeng2025timesuite}, TRACE~\citep{Guo2024TRACETG}, and MUSEG~\cite{luo2025museg} on the same 306 S3 items. All three models read only the S3 textual cue plus the corresponding source video $B$; no S1 or S2 information is provided. Table~\ref{tab:s3-specialized-models} reports mIoU per category and overall.

\begin{table}[h]
  \centering
  \small
  \begin{tabular}{lccc}
    \toprule
                  & \textbf{TimeSuite} & \textbf{TRACE} & \textbf{MUSEG} \\
                  & mIoU (\%)          & mIoU (\%)      & mIoU (\%)      \\
    \midrule
    Celebration   &  9.5              & 8.8            & 22.4            \\
    Dance         &  3.9              & 2.0            & 15.5            \\
    Game          & 11.2              & 6.0            & 21.6            \\
    Music         & 10.6              & 18.6           & 26.4            \\
    Wedding       & 10.5              & 13.5           & 13.8            \\
    \midrule
    \textbf{Overall} & \textbf{9.2}   & \textbf{5.9}   & \textbf{19.3}   \\
    \bottomrule
  \end{tabular}
  \caption{S3 temporal-localization mIoU of specialized temporal-grounding models on CMB, broken down by category. All three models receive the concept name and sub-event cue as text input and produce a free-form $[\hat{t}_{\text{start}}, \hat{t}_{\text{end}}]$ span on video~$B$. They all evaluate at 1 fps with 224$\times$224 resolution.}
  \label{tab:s3-specialized-models}
\end{table}

\subsection{Stage-1 result breakdowns}
\label{app:s1_breakdowns}

\begin{figure*}[!htb]
    \centering
    \includegraphics[width=0.9\textwidth]{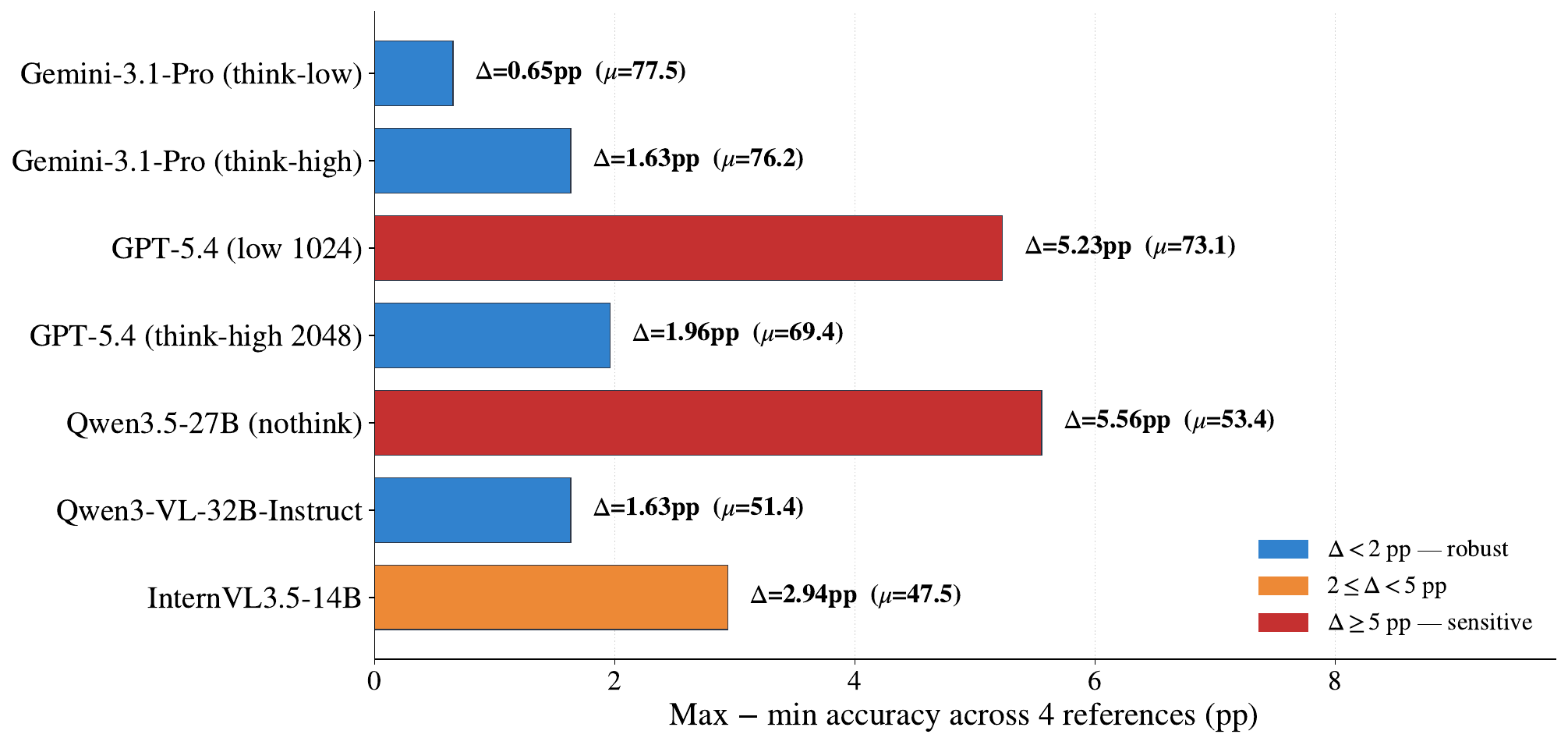}
    \caption{S1 reference variance across model configurations. For each model, S1 is rerun four times using the four name-reference variants per concept (common vs. official $\times$ Latin transliteration vs. local script); the bar reports the maximum minus the minimum S1 accuracy across the four runs (in percentage points), and the value in parentheses is the mean S1 accuracy ($\mu$) across them. Two configurations are reference-sensitive at the $\ge 5$\,pp threshold (GPT-5.4 low-1024, Qwen3.5-27B nothink); the remaining five stay within 3\,pp, indicating that S1 scores are robust to the choice of name reference.}
    \label{fig:app_s1_variance}
\end{figure*}

\begin{figure*}[!htb]
    \centering
    \includegraphics[width=1\textwidth]{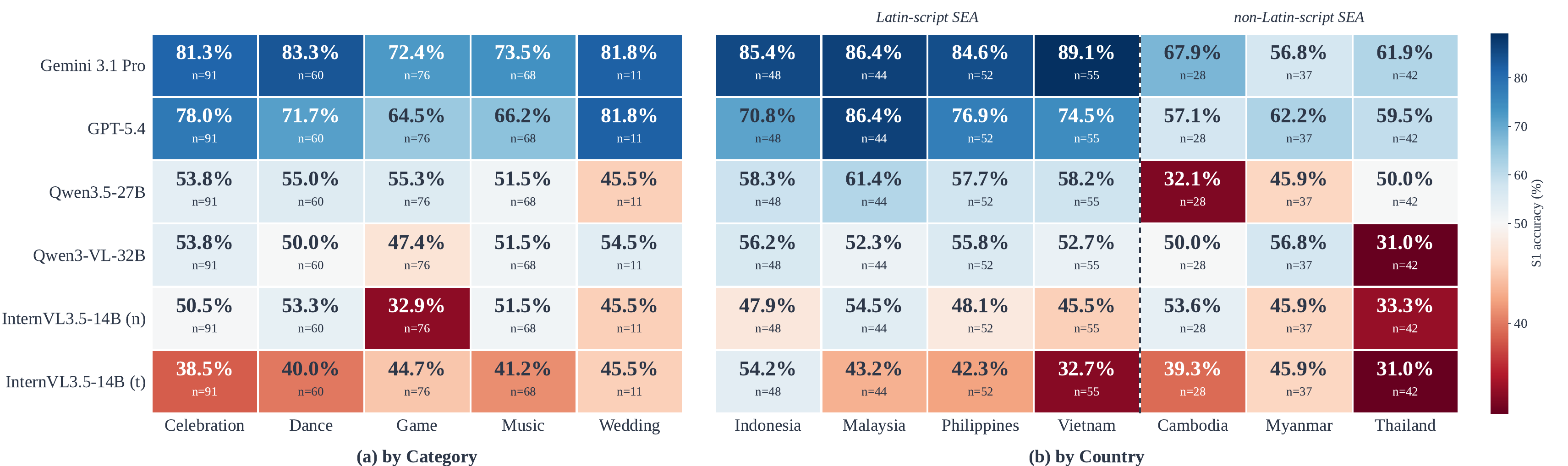}
    \caption{S1 (Naming) accuracy for all six models, split by category (left) and by country writing-system cluster (right). Cells report accuracy (\%); row $n$ is the number of concepts aggregated.}
    \label{fig:app_s1_combined}
\end{figure*}

Two views of the Stage-1 results support Section~\ref{sec:overall}. Figure~\ref{fig:app_s1_variance} reruns S1 with each of the four name-reference variants per concept and reports the spread; five of the seven configurations stay within 3 points, so S1 scores are largely robust to which name reference is shown. Figure~\ref{fig:app_s1_combined} splits S1 accuracy by category and by writing-system cluster.


\subsection{Model-specific intervention priorities}
\label{sec:cmb-priorities}

\paragraph{Beyond diagnosis, CMB suggests model-specific intervention priorities.}

Adding cultural knowledge is the priority for the smallest models, where naming itself is the limit. Improving visual recognition is the priority for the two Qwen models, which name well but do not carry the name to a video moment. Temporal localization is a persistent bottleneck across all the models. Regional adaptation of general-purpose VLMs is a complementary route to closing these gaps \citep{cahyawijaya2026anthropogenic}.

\section{Benchmark Categorization}
\label{sec:app_categorization}


This appendix explains how each entry in Table~\ref{tab:related_comparison} was categorized, with a focus on properties that required interpretation of the original paper.

\paragraph{SCB \citep{satar-etal-2025-seeing}.}

SCB is image-based and covers 138 cultural concepts from seven countries in Southeast Asia. We mark it as multi-stage with cascade because it has two distinct ability stages: Stage 1 is a visual MCQ for cultural reasoning, and Stage 2 evaluates segmentation of the cultural artifact only when Stage 1 is correct. This is a sequential cascade with conditional dependency. Stage 2 is spatial grounding, not temporal localization, so we mark it as \xmark on free-form temporal localization. SCB does not report human accuracy on the same task as models (described as future work in the paper).

\paragraph{VideoNorms \citep{VideoNorms_2025}.}

VideoNorms covers American and Chinese cultural norms through TV show clips. Each clip is fixed at 15 seconds (0.25 min). The benchmark has three parallel tasks (binary adherence/violation classification, evidence extraction, and norm generation), but these are scored independently rather than sequentially. Evidence quality (Task 2) is computed only on correctly-labeled samples, which is a partial form of conditional analysis; we mark Conditional metrics as \xmark because the headline metric is not a stage cascade. Humans validated only a subsample of LLM-judged outputs; no human accuracy on the full task is reported.

\paragraph{GIMMICK \citep{schneider2025gimmickgloballyinclusive}.}

GIMMICK is a multitask cultural benchmark spanning UNESCO ICH. The benchmark includes both image (CIVQA) and video (CVVQA) subsets; we compare against the CVVQA video subset, which has 1{,}809 samples with 10-second clips (0.17 min) drawn from 139 countries. The six tasks are evaluated independently, so we mark Multi-stage cascade as \xmark. GIMMICK explicitly evaluates only in English; SEA's focus is partial because the Asia-Pacific region is one of six macro-regions, but it is not the focus.

\paragraph{ViMUL-Bench \citep{shafique-etal-2025-culturally}.}

ViMUL-Bench spans 14 languages across 14 countries. The 14 languages include Arabic, Bengali, Chinese, English, French, German, Hindi, Japanese, Russian, Sinhala, Spanish, Swedish, Tamil, and Urdu; none is a language of Southeast Asia, so we mark SEA focus as \xmark. Question formats (MCQ, short open-ended, long open-ended) are parallel, not cascade stages, so Multi-stage cascade is \xmark. Average video duration was computed from the released dataset as 175 seconds (2.92 min). Humans only validate the LLM-judge on a sub-sample, so independent evaluation is marked partial.

\paragraph{MINERVA-Cultural \citep{Singh2026MINERVACulturalAB}.}

MINERVA-Cultural is a long-form (12.62 min average) multilingual benchmark across 18 locales. Of these 18 locales, 2 are in Southeast Asia (id-ID for Indonesia, th-TH for Thailand), so the SEA focus is partial. The benchmark format is single-stage open-ended QA judged by an LLM on a 0/1/2 scale; we therefore mark Multi-stage cascade as \xmark. MINERVA introduces Iterative Error Isolation (IEI), which traverses an evidence DAG with corrective hints across three iterations. We mark Conditional metrics as \cmark because IEI reports cross-step success-given-success. MINERVA's independent human evaluators do not see ground-truth answers but are permitted to use web search for grounding unfamiliar cultural entities; we therefore mark Human eval as \cmark (without the \textit{closed} annotation), reflecting independent evaluation under an open-book protocol.

\paragraph{AVMeme Exam \citep{jiang2026avmeme}.}

AVMeme Exam evaluates audio-visual cultural memes. The paper describes its coverage as ``East and South Asia, Middle East, Europe, and North America'' with contributors from the U.S., China, Japan, India, and the Middle East; no country in Southeast Asia is named in the coverage list, contributor list, or reported results languages, so SEA focus is \xmark. Twenty independent human participants evaluated on the same task with an explicit closed-book protocol: ``without web search, collaboration, or external assistance,'' so we mark Human eval as \cmark (\textit{closed}). Video duration is capped at 30 seconds (0.5 min); no average is reported.

\paragraph{ChineseVideoBench \citep{nie2025chinesevideobench}.}

ChineseVideoBench is monocultural (Chinese-only), with 1{,}625 videos averaging about 1 minute each. It has eight main task categories, including a ``Temporal Localization'' category, but inspection of the question format shows it is an MCQ over orderings rather than free-form span prediction; we therefore mark Free-form temporal localization as \xmark. The benchmark explicitly removes audio tracks to force visual reasoning. The reported human baseline of 94.8\% comes from the third subgroup of the same nine-annotator pool that built the benchmark, so we mark Human eval as \xmark: the evaluators are not independent of the annotators.

\paragraph{VideoVista-CulturalLingo \citep{chen-etal-2025-videovista}.} 

VideoVista-CulturalLingo covers three cultural regions (Chinese, North American, European) across 1{,}389 videos, with no focus on Southeast Asia, so we mark SEA focus as \ding{55}. The benchmark is a single-stage MCQ design without conditional or cascade scoring, so Multi-stage cascade and Conditional metrics are both \ding{55}. Average video duration reported as 4.05~min. The released paper does not describe an independent human-eval study on the same task, so Human eval is \ding{55}.

\end{document}